\documentclass[runningheads]{llncs}
\usepackage{graphicx}
\usepackage{comment}
\usepackage{amsmath,amssymb} % define this before the line numbering.
\usepackage{color}
\usepackage{array}

\usepackage{math}

\usepackage{graphicx}
\usepackage{amsmath}
\usepackage{amssymb}
\usepackage{caption}
\usepackage{subcaption}
\usepackage{makecell}
\usepackage{booktabs}
\usepackage{multirow}

\usepackage[pagebackref=true,breaklinks=true,letterpaper=true,colorlinks,bookmarks=false,hypertexnames=false]{hyperref}
\newcommand\blfootnote[1]{%
  \begingroup
  \renewcommand\thefootnote{}\footnote{#1}%
  \addtocounter{footnote}{-1}%
  \endgroup
}
\def\ie{\textit{i.e.~}}

\def\etal{\textit{et al.~}}

\begin{document}
% \linenumbers
\pagestyle{headings}
\mainmatter
\def\ECCVSubNumber{3745}  % Insert your submission number here

\title{Motion-supervised Co-Part Segmentation} % Replace with your title

\begin{comment}
\titlerunning{ECCV-20 submission ID \ECCVSubNumber} 
\authorrunning{ECCV-20 submission ID \ECCVSubNumber} 
\author{Anonymous ECCV submission}
\institute{Paper ID \ECCVSubNumber}
\end{comment}
%******************

%\begin{comment}
\titlerunning{Motion-supervised Co-Part Segmentation}
%\author{First Author\inst{1}\orcidID{0000-1111-2222-3333} \and
%Second Author\inst{2,3}\orcidID{1111-2222-3333-4444} \and
%Third Author\inst{3}\orcidID{2222--3333-4444-5555}}
%
\author{
Aliaksandr Siarohin\inst{1*} \and
Subhankar Roy\inst{1, 4*} \and
St{\'e}phane Lathuili{\`e}re\inst{2} \and
Sergey Tulyakov\inst{3} \and
Elisa Ricci\inst{1, 4} \and
Nicu Sebe\inst{1, 5}
}
\authorrunning{A. Siarohin and S. Roy et al.}
\institute{\textsuperscript{1}DISI, University of Trento;
 \textsuperscript{2}LTCI, Télécom Paris, Institut polytechnique de Paris;
 \textsuperscript{3}Snap Inc.;
 \textsuperscript{4}Fondazione Bruno Kessler;
 \textsuperscript{5}Huawei Technologies Ireland }
%\end{comment}
%******************

\maketitle

\begin{abstract}
  \blfootnote{* denotes equal contribution}
  Recent co-part segmentation methods mostly operate in a supervised learning setting, %\alex{Can you rewrite this supervised setting?}, 
  which requires a large amount of annotated data for training.  To overcome this limitation, we propose a self-supervised deep learning method for co-part segmentation. Differently from previous works, our approach develops the idea that motion information inferred from videos can be leveraged to discover meaningful object parts. To this end, our method relies on pairs of frames sampled from the same video. The network learns to predict part segments  together with a representation of the motion between two frames, which permits reconstruction of the target image.
   Through extensive experimental evaluation on publicly available video sequences we demonstrate that our approach can produce improved segmentation maps with respect to previous self-supervised co-part segmentation approaches.
\end{abstract}

\section{Introduction}
Discovering objects and object parts in images is one of the fundamental steps towards semantic understanding of visual scenes. In computer vision this problem is referred to as semantic segmentation and is approached within a machine learning framework as a dense labeling task, where the goal is to assign a categorical label to each pixel of an image. In this paper we address a more challenging problem and a special case of semantic segmentation, referred to as \textit{co-part segmentation}. The task of the co-part segmentation problem is to identify segments corresponding to different parts within a single object. For instance, in a human body the relevant parts correspond to hands, legs, head and torso. %Discovery of constituent parts in an object are of utmost importance since ... that makes them robust to sensor changes and appearance variations.
Such parts are of special interest for the automatic analysis of visual scenes since they constitute intermediate representations, which are robust to sensor changes and appearance variations.
%This paper introduces a novel framework for unsupervised co-part segmentation.  Co-part segmentation refers to a special case of the more general problem of semantic segmentation. Semantic segmentation is a dense labeling task, where the goal is to assign a categorical label to each pixel of an image. In co-part segmentation the task is to identify segments corresponding to different parts of a single object. For instance, given a human body relevant parts would correspond to hands, legs, head and torso. 

%% \begin{figure}[t]\centering
%% \includegraphics[width=0.78\linewidth]{figures/teaser.pdf}
%% \caption{Co-part segmentation is learned following a self-supervised approach. We leverage motion information to learn a segmentation network without annotation. At training time, we consider frame pairs extracted from the same video (\emph{source} and \emph{target}).
%% The key idea consists in predicting segments from the target frame that can be combined with a sparse motion representation between the two frames in order to reconstruct the input target frame.}
%% \label{fig:teaser}
%% \end{figure}
%$\mathcal{T}_{\mathbf{S} \leftarrow \mathbf{T}}$
\begin{figure}[t]\centering
\begin{subfigure}[b]{0.5\textwidth}
         \centering
\includegraphics[height=3.2cm]{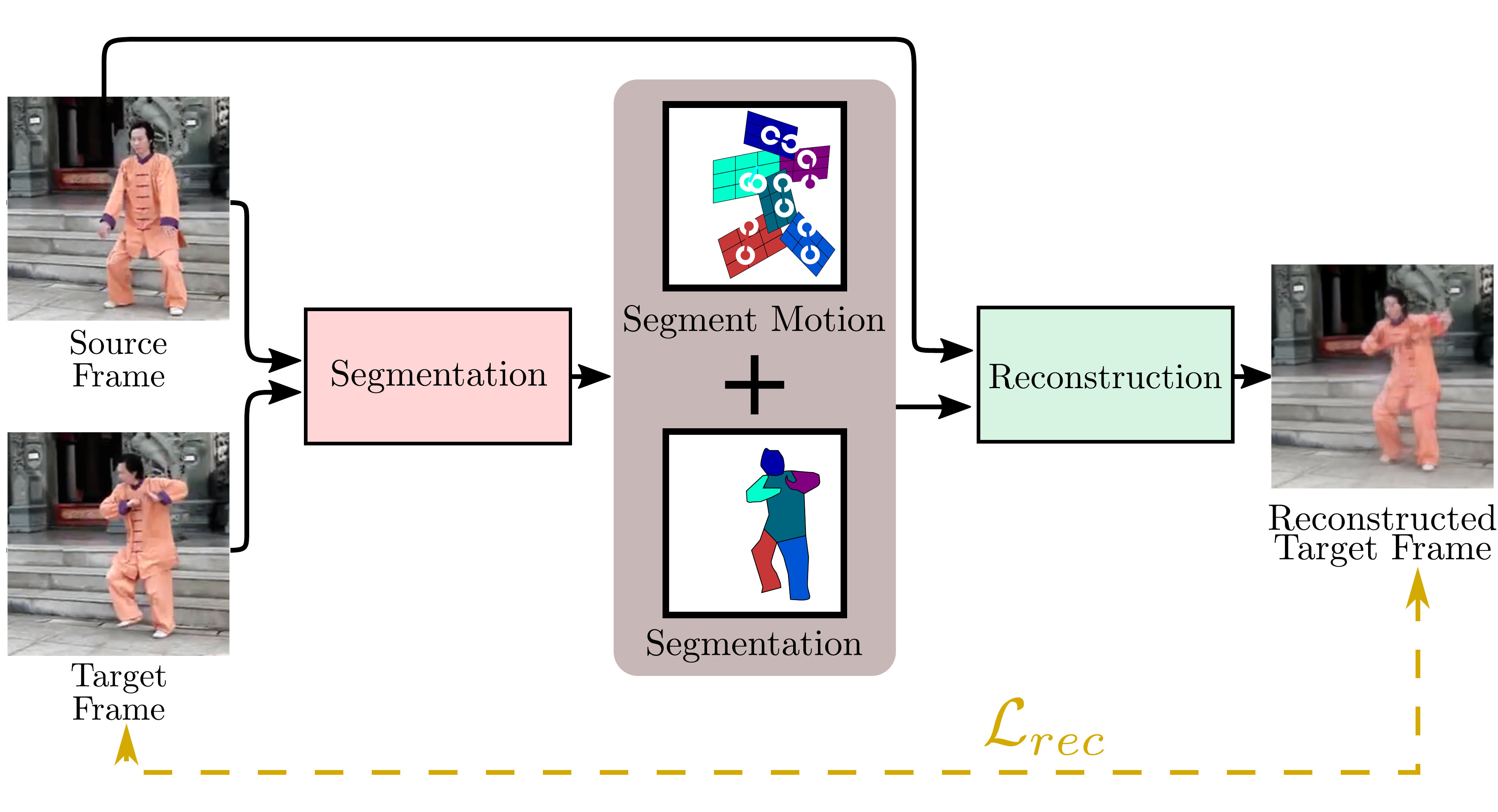} \label{Training on video frames}
\vspace{-0.2cm}
\caption{Self-supervised training on video frames}
     \end{subfigure}
\begin{subfigure}[b]{0.4\textwidth}
         \centering
\includegraphics[height=3.2cm]{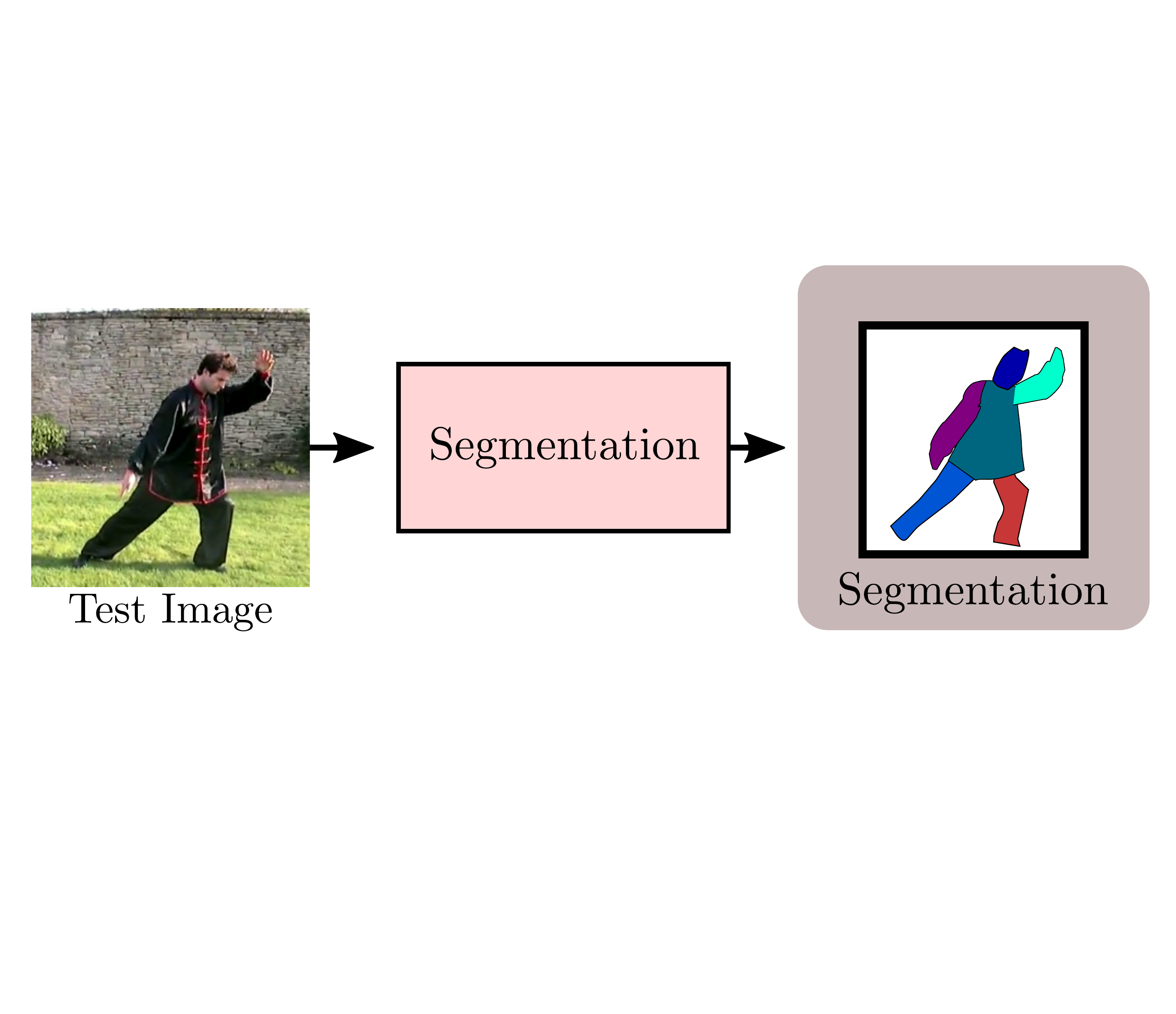} \label{Test on still images}
\vspace{-0.2cm}
\caption{Test on images}
\end{subfigure}
\vspace{-0.3cm}
\caption{{\textbf{Co-part segmentation via a self-supervised approach.} We leverage motion information to train a segmentation network without annotation. At training time (a), we use frame pairs (\emph{source} and \emph{target}) extracted from the same video.
The key idea consists in predicting segments from the target frame that can be combined with a motion representation between the two frames in order to reconstruct the target frame. At inference time (b) only the trained segmentation model is used to predict object parts segments on a test image.}}
\label{fig:teaser}
\vspace{-0.7cm}
\end{figure}

Recently, co-part segmentation algorithms have gained popularity as they are key-enabling components for image editing and animations tools.
Several works use automatically computed object parts for virtual try-on~\cite{yu2019vtnfp}, pose-guided generation~\cite{dong2018soft}, face-swap and re-enactment~\cite{nirkin2019fsgan}. The vast majority of co-part segmentation methods operates in a supervised setting~\cite{chen2014detect,liu2015deep}, thus requiring a large amount of human annotated data. To overcome this limitation some works have focused on the challenging problem of unsupervised co-part segmentation~\cite{collins2018,Hung_2019_CVPR}, demonstrating that object parts can be inferred without relying on labeled data. While effective, these approaches are inherently limited by the fact that only appearance-based features are exploited to compute segments. 
In contrast we argue and experimentally demonstrate that more precise information about object parts can be extracted by leveraging motion information from vast amounts of unlabeled videos. Thus, in this paper we propose a novel video-based self-supervised co-part segmentation method. Once trained, our proposed framework takes a single image as input and outputs the segmentation map indicating different object parts. Differently from previous methods~\cite{Hung_2019_CVPR,collins2018}, we assume that at training time, a large collection of videos depicting objects within the same category is available. 

Our approach is inspired by previous works on self-supervised keypoints estimation~\cite{jakabunsupervised,zhang2018unsupervised,lorenz2019unsupervised}.
The main idea behind these methods is to disentangle the semantic and appearance representation of an object by imposing a reconstruction objective. Specifically, these approaches operate by considering two images of the same object in different poses, typically referred to as source and target images, obtained through synthetic deformations or from video sequences. In summary, they compute the appearance representation from the source image and the semantic representation from the target and then attempt to reconstruct the target image combining these two representations. However, in order for these approaches to work and to successfully extract the semantic representation, the network should have a tight information bottleneck. If this condition is not met, the semantic representation will be contaminated by the appearance information and the disentanglement will be poor. We call this phenomenon \textit{leaking}. Because of \emph{leaking}, most self-supervised methods~\cite{jakabunsupervised,zhang2018unsupervised,lorenz2019unsupervised} are inherently limited to focus on low-dimensional object representation, \ie keypoints. Whereas, in our case this bottleneck semantic representation are segmentation maps.
%  In fact, the amount of information that can be encoded into several keypoints is very limited, and those no appearance information can leak into semantic representation. 
In fact, compared to the keypoints, the semantic segmentation maps lie in a much higher dimensional space. Consequently, naive usage of the above methods will lead to heavily \emph{leaking} models that use the segmentation map to encode the target appearance rather than predicting representative segmentation maps.

This paper tackles this problem and introduces a novel deep architecture that predicts meaningful semantic representations without \emph{leaking}. 
Our approach predicts segments associated with the target frame and uses them, combined with a per-segment motion representation,
%a representation of individual segment motion derived from the source and target frames, 
to reconstruct the target image (Fig.~\ref{fig:teaser}).
Specifically, we propose a part-based network that operates by deforming each part from the source image in order to match the corresponding part in the target image. This deformation is derived from the predicted motion representation for each part. In addition, we incorporate \textit{background visibility mask} in order to achieve better background-foreground separation. We tested our method on two datasets: Tai-Chi-HD~\cite{siarohin2019neurips} and VoxCeleb~\cite{Nagrani17}. Through an in-depth ablation study and an extensive comparison with previous methods on co-part segmentation~\cite{collins2018,Hung_2019_CVPR}, we demonstrate the effectiveness of our approach and the rationale behind our architectural choices.

To summarize, our contribution is twofold. We introduce the problem of video-based co-part segmentation, showing that motion information can and should be adopted for inferring meaningful object parts. We hope that our work will stimulate future work in this new research direction. Additionally, we propose a novel deep architecture for co-part segmentation. Our approach advances the state-of-the-art for self-supervised object parts discovery, demonstrating that not only object landmarks but also complex segmentation maps can be inferred through disentanglement within a reconstruction framework.  We make our source code publically avalaible\footnote{\href{https://github.com/AliaksandrSiarohin/motion-cosegmentation}{https://github.com/AliaksandrSiarohin/motion-cosegmentation}}.

%but the main problem of these approaches is that segmentation is based only on the appearance information, this assumption is quite limiting and, in some case, it prevents the discovery of meaningful parts. In this paper we propose to exploit motion information extracted from video streams.

\section{Related Works}

\begin{figure*}[t]\centering
\includegraphics[width=0.99\linewidth]{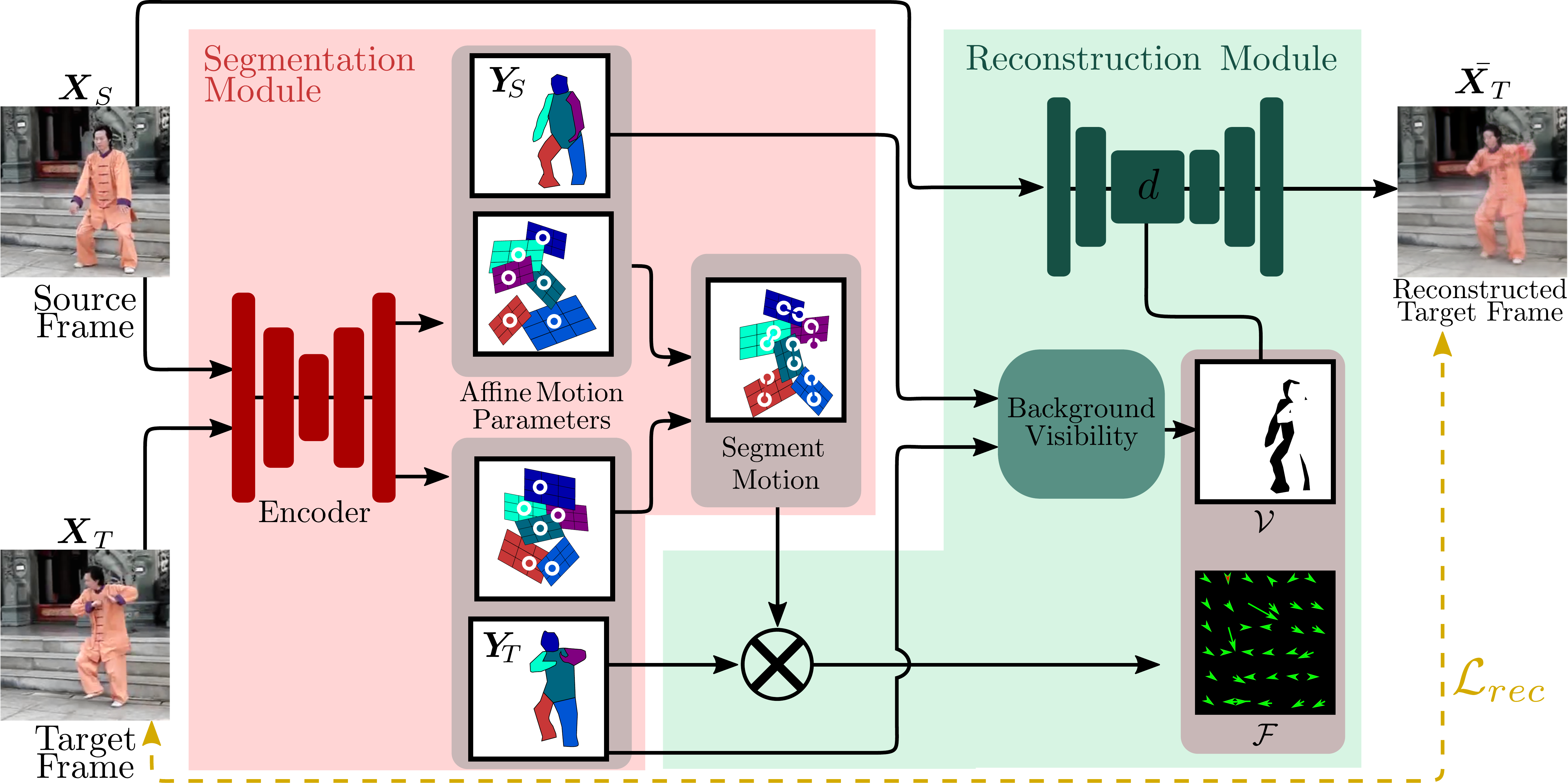}
\caption{\textbf{Overview of our method.} It consist of two modules: \textit{Segmentation Module} and \textit{Reconstruction Module}. On the one hand, the \textit{Segmentation Module} takes as input a source frame $\Xvect_S$ and a target frame $\Xvect_T$, sampled from the same video clip. The \textit{Segmentation Module} predicts the segmentation maps $\Yvect_S$ and $\Yvect_T$, along with the affine motion parameters. On the other hand, the \textit{Reconstruction Module}, is in charge of reconstructing the target images from the source image and the segmentation module outputs.  First, it computes a background visibility mask $\Vvect$ and an optical flow $\mathcal{F}$ from the segmentation maps and the affine motion parameters. Finally, it reconstructs the target frame $\Xvect_T$ by warping the features of the source frame $\Xvect_S$ and masking occluded features.
%\steph{O,dense motion->eq ,occlusion-> eq}
}
\label{fig:pipeline}
\vspace{-0.7cm}
\end{figure*}

\textbf{Self-supervision.}
Self-supervised learning is a form of unsupervised learning where supervision is attained by solving a pretext task. Recently, it has emerged as an efficient way to learn rich image or video representations. A popular strategy consists in modifying the input data and asking the network to automatically recover the applied modification. Depending on the application, a different family of transformations is applied such as rotations~\cite{agrawal2015learning,feng2019self,Gidaris2018UnsupervisedRL}, or patch shuffling~\cite{noroozi2016unsupervised}.  Another successful approach trains a network on a prediction task on partially observed data. The task then consists in either predicting an unobserved part~\cite{srivastava2015unsupervised,luo2017unsupervised} or the relationship between different data parts~\cite{doersch2015unsupervised}. Self-supervision can also be based on the principle of auto-encoding~\cite{hinton2006reducing,song2018self,Zhang_2017}. This auto-encoding approach is especially popular when it comes to learning disentangled representations~\cite{siarohin2018animating,jakabunsupervised,ma2018disentangled}. %\eli{In this work, we propose an auto-encoder approach trained with an equivariance loss that enforces equivariance to provided set of transformations}.
Our method also falls under this category.

\noindent\textbf{Landmark Discovery.}
Recently, several methods have been proposed to learn landmarks in a self-supervised manner. Thewlis \etal\cite{Thewlis_2017_ICCV} proposed to learn a keypoint detector relying on geometric priors, where their training loss enforces the network to be equivariant to affine and thin spline transformations. Zhang \etal\cite{zhang2018unsupervised} extended this approach using an auto-encoder architecture where the bottleneck information is the landmark locations. A similar auto-encoding strategy is used in~\cite{jakabunsupervised}. % by learning to detect landmarks in images by training on videos. 
In the context of video generation, Siarohin \etal\cite{siarohin2018animating,siarohin2019neurips} proposed animation methods that discover landmarks specifically designed to encode motion information. Similarly to Siarohin \etal\cite{siarohin2018animating,siarohin2019neurips}, our method is also based on the motion information. However, while animation methods assumes similar part based motion model, those methods require at least 2 frames to predict keypoint neighbourhoods, even during the inference. This makes predictions of Siarohin~\etal\cite{siarohin2018animating,siarohin2019neurips} highly dependent on the other frame in a pair. For example (i) if there is not enough motion between 2 frames then these methods will not predict any neighbourhoods; (ii) image animations methods may include a lot of background motion in the motion prediction. In contrast our approach encodes more semantically meaningful parts by making independent frame-based predictions.

% Assuming an image $\vect{X}$ and its anchor keypoints $p\!=\!\{p^k\}_{k=1..K}$, we sample a random spatial deformation $\mathcal{D}$ and estimate the keypoints $\tilde{p}\!=\!\{\tilde{p}^{k}\}_{k=1..K}$ of $\mathcal{D}(\Xvect)$. We define the equivariance loss as follows:
% \begin{equation}
% \mathcal{L}(p,\tilde{p})= \sum_{k=1}^K ||p^k-\tilde{p}^k||^2.
%   \end{equation}
% Besides, we also impose an equivariance loss on the $\Avect_k$ matrices used in our affine model. More specifically, we employ the equivariance loss of \cite{siarohin2019neurips}. The total of the two equivariance losses is referred to as $\mathcal{L}_{eq}$ \eli{formula???}.  Differently, in our case this motion information is used for obtaining segmentation maps. 

\noindent\textbf{Unsupervised Co-Part Segmentation.}
%Earlier works on part segmentation are based on energy maximization problems which must be solved at inference time \cite{rubio2012unsupervised,joulin2012multi,rubinstein2013unsupervised,tsai2016semantic}. %Furthermore, these methods are limited to foreground pixel segmentation.
Most co-part segmentation methods operate in a supervised setting~\cite{chen2014detect,liu2015deep}. However, here we only review previous works on unsupervised co-part segmentation. Recent methods propose to study the internal feature representations derived from pre-trained ConvNets to find object part information~\cite{bau2017network,gonzalez2018semantic}. %Following this line of research, class activation maps methods~\cite{zhou2016learning,selvaraju2017grad} are derived in order to localize image regions corresponding to high response from a pre-trained classifier.
Collins \etal\cite{collins2018} described an approach to estimate segments corresponding to object parts by applying Non-negative
Matrix Factorization~\cite{sra2006generalized} (NMF) on features computed from a pretrained ConvNet. This requires a costly optimization process at inference time and also do not incorporate geometric priors.
{Xu \etal\cite{xu2019unsupervised} proposed a deep model to discover object parts and the associated hierarchical structure and dynamical model from unlabeled videos. However, they assume that precomputed motion information is available. Differently, our method estimates the motion within the same architecture.
Recently, Hung \etal\cite{Hung_2019_CVPR} proposed a self-supervised deep learning approach for co-part segmentation from static images.
A network is trained using several losses in order to impose geometric, equivariance and semantic consistency constraints. Hung \etal\cite{Hung_2019_CVPR} rely on a large collection of unlabelled images of the same object category. In this work, we also adopt a self-supervised learning strategy for co-part segmentation. However, rather than considering static images, we propose to train our model on a collection of unlabelled videos. Our motivation is to leverage motion information with the purpose of obtaining segments that correspond to group of pixels associated to object parts moving together.}

As a final remark, we would like to point out that co-part segmentation should not be confused with co-image and co-video segmentation~\cite{li2018deep,chiu2013multi}. In co-image and co-video segmentation the task is to discover pixels corresponding to common foreground objects of the same class within images and videos, respectively. In both cases, the final goal is to segment a single foreground object and not to discover object parts.

\section{Method}

In this work, we are interested in training a deep neural network that at inference time takes a single image as input and outputs co-part segments. At training time, we assume a large collection of videos containing objects of the same category is available. The proposed model is trained using pairs of frames from the same video: $\Xvect_S\in\mathbb{R}^{3\times H\times W}$ and $\Xvect_T\in\mathbb{R}^{3\times H\times W}$ of spatial dimension $H\times W$ and are referred to as \emph{source} and \emph{target} frames, respectively. Note that the pair of frames are randomly sampled from the entire video sequence.
%More specifically, the model is trained by reconstructing the target frame $\Fvect$ from the source frame $\Svect$, the estimated segmentation of $\Fvect$, and a sparse motion representation between $\Fvect$ and $\Svect$. 

The overall pipeline is illustrated in Fig.~\ref{fig:pipeline}. Our framework is composed of two main modules. The first module, named \emph{Segmentation Module}, is in charge of extracting semantic segments and the motion corresponding to those segments from the input frames. We employ a neural network to independently segment the frames $\Xvect_S$ and $\Xvect_T$ into $K\!+\!1$ parts, \ie$K$ segments for the foreground and one for the background. The predicted segments are represented using $K\!+\!1$ channel tensors, \ie $\Yvect_S\in[0,1]^{K+1\times H'\times W'}$ and $\Yvect_T\in[0,1]^{K+1\times H'\times W'}$ associated to the source and the target frames, respectively. Note that, the segmentation source $\Yvect_S$ is predicted independently from the target segmentation and, consequently, the proposed model can be used for single image segmentation at inference time.
To reduce time for computing the segmentation masks we use a smaller resolution $H'\times W'$.%, which is smaller than the image resolution. %% Note that the last segmentation part is associated to the background.
We use a U-Net-like architecture with a channel-wise softmax layer at the output to obtain tensors that can be interpreted as part-assignment confidence maps.

In our self-supervised setting, we also aim at training a generator network that reconstructs the target frame from the source frame and the target segmentation. However, reconstructing the target frame directly from its segmentation maps tends to provide pathological solutions that use $\Yvect_T$ to encode $\Xvect_T$, rather than predicting desired semantic segmentation. In contrast to unsupervised landmark detection, this problem appears particularly severe in the case of segmentation because of the relatively high dimension of $\Yvect_T$ (typically $64\!\times\!64$). In this case, $\Yvect_T$ is too large to serve as the information bottleneck allowing the information about the target frame to \emph{leak} to the generator. Therefore, we use the estimated segmentation to reconstruct the target frame in an indirect way. More precisely, we propose to reconstruct the target frame by modeling the motion between the source and the target frames and we include the segmentation $\Yvect_T$ into our model of motion. To this end, we introduce the segment motion that describes the motion between the source and the target frames in each segment. The details are provided in Sec.~\ref{sec:sparse}.

Finally, the second module, named \textit{Reconstruction Module}, is in charge of reconstructing the target frame from the source frame and the segmentation module outputs.
With the help of segment motion we estimate optical flow between source and target frames. Importantly, we do not use an external optical flow estimator, and rely solely on segmentation maps and segment motions. Optical flow $\mathcal{F}\!:\mathbb{R}^2\!\rightarrow\!\mathbb{R}^2$ maps each pixel location in $\Xvect_T$ to its corresponding location in $\Xvect_{S}$. In other words, $\mathcal{F}$ is the estimated backward optical flow between the source and the target frames. Based on segmentation maps we also estimate a background visibility mask $\Vvect \in \mathbb{R}^{H\times W}$. Background visibility mask indicates parts of the background that were occluded in the source frame. Details concerning optical flow and background visibility mask are provided in Sec.~\ref{sec:dense}.
The target frame is reconstructed from the source frame using an encoder-decoder network. To this aim, we align the intermediate feature representation of the source frame with the pose of the object in the target frame using the estimated optical flow $\mathcal{F}$.
The overall architecture is trained by minimizing a reconstruction loss between the target and reconstructed target frames.
The reconstruction module and the training procedure are described in Sec.~\ref{sec:recTrain}.

%% Besides, this encoder network predicts $K$ keypoint locations in $\Xvect_S$ and $\Xvect_T$, along with local descriptors of the object pose (see details in Sec.\ref{sec:sparse}). Importantly, each keypoint is associated with one segmented part. 
%%  

\subsection{Segmentation Module}
\label{sec:sparse}

In this section, we detail how we model the motion of each segment. We design our model with the aim of obtaining segments that group together pixels corresponding to object parts that move together. The motion of the pixels within each segment is modeled by an affine transformation. In other words, the optical flow $\mathcal{F}$ can be approximated by an affine transformation on the support corresponding to each segmented part. The encoder network should provide segments such that our affine model is valid. Otherwise, it would lead to wrong optical flow $\mathcal{F}$ and, subsequently, to high reconstruction loss.

Formally, let $\Yvect_T^k\in[0,1]^{H'\times W'}$ be the $k^{th}$ channel of the segmentation $\Yvect_T$. Even if $\Yvect_T$ and $\Yvect_S$ are tensors with continuous values, for the sake of notation, here we assume that they are binary tensors (this point is further discussed in Sec.~\ref{sec:dense}). We define $\mathcal{Y}_T^k$ as the set of locations associated to the segment $k$, $\mathcal{Y}_T^k=\{z\in[1,H']\times[1,W']~| ~\Yvect_T^k[z]=1\}$. In a similar way, we introduce $\mathcal{Y}_S^k$ for the source segmentation. Assuming that the motion of each segment follows an affine model, this implies that there exists $\Avect\in\mathbb{R}^{2\times 2}$  and $\beta\in\mathbb{R}^{2}$ such that:
\vspace{-0.1cm}
\begin{equation}
\vspace{-0.1cm}
  \forall z \in\mathcal{Y}_T^k,~\mathcal{F}(z)
  =  \Avect z + \beta.  \label{approxaffine}
\end{equation}
In our context, we need to explicitly estimate the affine parameters $\Avect$ and $\beta$ in order to obtain $\mathcal{F}$. To this end, we rewrite Eq.~\eqref{approxaffine} as follows: 
\vspace{-0.2cm}
\begin{equation}
\vspace{-0.2cm}
%%   \medmuskip=0mu
%% \thinmuskip=0mu
  %% \thickmuskip=0mu
  \forall z\in\mathcal{Y}_T^k,~\mathcal{F}(z)
  =  p_S^k +  \Avect_S^k{\Avect_T^k}^{-1}\big(z - p_T^k\big)  \label{approxJ}
\end{equation}
where $\Avect_S^k\in\mathbb{R}^{2\times 2}$, ${\Avect_T^k}^{-1}\in\mathbb{R}^{2\times 2}$, $p_S^k\in\mathbb{R}^{2}$ and $p_T^k\in\mathbb{R}^{2}$ are estimated by the segmentation network, see Sup. Mat.~\ref{sec:imp-det} for details.
The motivation for this re-writing is to separate terms that are estimated only from the source (\ie $p_S^k$ and $\Avect_S^k$) from those estimated only from the target (\ie $p_T^k$ and $\Avect_T^k$), see \cite{siarohin2019neurips}.

%\alex{Intuitively this can be understood as translating target to some canonical coordinate system and translating from this canonical coordinate system to source. This choice further prevent the \emph{leaking} problem, since target $p_T^k$ and $\Avect_T^k$ should work equally well for all possible sources. }
%% The locations $p_S^k\in\mathbb{R}^{2}$ and $p_T^k\in\mathbb{R}^{2}$ are estimated from the source and from the target frame respectively.

\subsection{Flow Model}
\label{sec:dense}

We now detail how optical flow $\mathcal{F}$ is obtained by combining the segment motion representations and the predicted segmentation.
For each segment, we employ the model given by Eq.~\eqref{approxJ} to obtain $K$ optical flow fields $\Fvect^k\in \mathbb{R}^{2\times H'\times W'}$ of $\mathcal{F}$ in each respective segment  $\mathcal{Y}_F^k$. The tensor $\Fvect^k$ can be interpreted as the partial optical flow for the pixels of the segment $\mathcal{Y}_T^k$ between the source and target frames. Note that the last optical flow field $\Fvect^{K+1}$ is associated to the background. We assume a static background and consequently set $\Fvect^{K+1}=z$. We consider that each object part corresponds to a segment in $\Yvect_T$ and that its motion can be encoded by its corresponding segment motion descriptor. Consequently, we obtain the following discretized estimator $\hat{\Fvect}$ of the field $\mathcal{F}$:
\vspace{-0.2cm}
\begin{equation}
\vspace{-0.2cm}
\hat{\Fvect}=\sum_{k=1}^{K+1} \Yvect_F^{k}\otimes \Fvect^{k} \label{eq:estim}
\end{equation}
where $\otimes$ denotes the element-wise product. In Eq.~\eqref{eq:estim}, the segmentation mask $\Yvect_T^{k}$ is used to assign a partial optical flow field $\Fvect^{k}$ to each location. In other words, the encoder network must output segmentation masks that allows to correctly assign partial  optical flow field $\Fvect^{k}$ to each location $z$. Because of this, the segmentation mask must gather together locations that move in the same way.
In practice, $\Yvect_T^{k}$ are continuous tensors. Consequently, Eq.~\eqref{eq:estim} can be interpreted as a soft assignment model that selects motion vector for each pixel location. The use of continuous tensors facilitates gradient-based optimization.

\noindent\textbf{Background visibility mask.}
Additionally we estimate the background visibility mask $\Vvect \in [0,1]^{H'\times W'}$. It indicates which pixels of the background are occluded by the foreground object in the source frame. Later, information about the occluded parts of the background will be suppressed in the \emph{Reconstruction Module}. Usage of the background visibility mask enforces superior background/foreground separation. Poor segmentation maps will lead to poor background visibility maps and this will affect the final reconstruction quality. $\Vvect$ is computed from the source and target segmentations. 

We estimate the foreground object locations in the source frame by merging the K object parts $\Ovect_S\!=\!\sum_{k=1}^K\Yvect_S^k$. Occluded locations are then defined as locations corresponding to the background in the target ($\Yvect_T^{K+1}\!=\!1$) but to the foreground in the source frame.  Formally, the background visibility mask is: \vspace{-0.2cm}
\begin{equation}
\vspace{-0.3cm}
\Vvect= 1 - (\Yvect_T^{K+1}\otimes\Ovect_S)
\end{equation}

Note that the background visibility provides an important regularization necessary for reducing parts of the background incorrectly classified as the foreground, which we later refer as \emph{false fg}. Indeed, a large portion of \emph{false fg} in the source frame means significant amount of information useful for background reconstruction will be suppressed. In turn, optimization of the reconstruction error will lead to shrinkage of \emph{false fg}. Compared to the occlusion map of~\cite{siarohin2019neurips} that specifies the parts that need to be inpainted thus improving the quality of reconstruction, our background visibility does not improve reconstruction. In fact, it is even harms it. We note that, providing background visibility to the generator is be problematic since information about the target frame will \emph{leak} through $\Yvect_T^{K+1}$. To this end, we prevent the gradient flow through the $\Yvect_T^{K+1}$.
% by detaching $\Yvect_T^{K+1}$ from the computational graph.

\subsection{Reconstruction and Training}
\label{sec:recTrain}

In this section, we describe how we make use of the estimated optical flow to reconstruct the target frame. Then, we provide the details of our training procedure.
We employ a generator network $G$ that reconstructs the target frame by warping the source frame features according to the estimated optical flow. Similar strategy has been adapted in several recent pose-guided generation frameworks~\cite{zablotskaia2019dwnet,siarohin2018deformable,zanfir2019human,siarohin2018animating,grigorev2019coordinate}. $G$ has an encoder-decoder architecture with a deformation layer in its bottleneck. This deformation layer takes as input a feature map $\xivect\in\mathbb{R}^{C\times H'\times W'}$, the estimated optical flow $\hat{\Fvect}\in\mathbb{R}^{2\times H'\times W'}$ and the background visibility mask ${\Vvect}$. We use ${\Vvect}$ to mask out the feature map locations not visible according to the segmentation maps. Thus, we reduce the impact of the features located in the non-visible parts. Note that if the features in that part were useful for reconstruction (this happens in the case of \emph{false fg}), the source segmentation will be penalized. The deformation layer output is defined as follows:
\vspace{-0.2cm}
\begin{equation}
\vspace{-0.1cm}
\xivect'=\Vvect\odot \mathcal{W}(\xivect,\hat{\Fvect})\label{eq:wrap}
\end{equation}
where $\mathcal{W}(\cdot,\cdot)$ denotes the back-warping operation and $\odot$ denotes the Hadamard product.
The warping operation is implemented in a differentiable manner using the bilinear sampler of Jaderberg~\etal\cite{jaderberg2015spatial}. 

\noindent\textbf{Training Losses.}
We train our model in an end-to-end fashion.
The reconstruction loss is our main driving loss. It is based on the perceptual loss of Johnson \etal~\cite{johnson2016perceptual} and uses a pre-trained VGG-19~\cite{DBLP:journals/corr/SimonyanZ14a} network to asses reconstruction quality. 
%\alex{We use the following 5 layers: conv1\_2, conv2\_2, conv3\_2, conv4\_2,
%conv5\_2, similarly to~\cite{wang2018video}.}\steph{If it is the same than 35 I would not list these layers since the notations are not clear.}\sergey{agree with Steph, we can move them to supplement}
Given the input target frame $\Xvect_T$ and its corresponding reconstructed frame $\bar{\Xvect_T}$, the reconstruction loss is defined as:
\vspace{-0.3cm}
  \begin{equation}
  \vspace{-0.2cm}
    \mathcal{L}_{rec}(\bar{\Xvect_T}, \Xvect_T)  = \sum_{i=1}^{I} \left|\Phi_i(\bar{\Xvect_T}) - \Phi_i(\Xvect_T) \right|,
\end{equation}
where $\Phi_i(\cdot)$ denotes the $i^{th}$ channel feature extracted from a specific VGG-19~\cite{DBLP:journals/corr/SimonyanZ14a} layer and $I$ denotes the number of feature channels in this layer. In addition, this loss is computed at several resolutions, similarly to MS-SSIM~\cite{wang2003multiscale,tang2018dual}.  

%% \alex{This loss we don't use}
Furthermore, following previous unsupervised keypoint detection methods \cite{jakabunsupervised,Zhang_2018_CVPR} we employ the equivariance loss to force the network to predict segment motion representations that are consistent with respect to known geometric transformations. As in~\cite{siarohin2019neurips}, we impose the equivariance loss with respect to affine transformations defined by $p^k$ and $\Avect^k$, we refer to this loss as  $\mathcal{L}_{eq}$.
Finally, the overall loss function is the summation of $\mathcal{L}_{rec}$ and $\mathcal{L}_{eq}$
%\begin{equation}
%\mathcal{L}_{tot}=\mathcal{L}_{rec}+\mathca%l{L}_{eq}.
%  \end{equation}
To facilitate the convergence and improve the performance we start with the pre-trained weights of~\cite{siarohin2019neurips}, but we believe any unsupervised keypoint detection method~\cite{siarohin2018animating,lorenz2019unsupervised,jakabunsupervised} can be used. We also experimented with training from scratch. We observed that the model does learn meaningful segments, however, their precision is sub-optimal.

\section{Experiments}
\label{sec:exp}

% \begin{table}[t]
%     \centering
%     \begin{tabular}{lcc c cc}
%         \specialrule{1.5pt}{1pt}{1pt}
%          & \multicolumn{2}{c}{\emph{Tai-Chi-HD}} && \multicolumn{2}{c}{\emph{VoxCeleb}}\\
%          \cline{2-3} \cline{5-6}
%          Model & MAE $\downarrow$ &  IoU $\uparrow$ && MAE $\downarrow$ & IoU $\uparrow$\\
%          \specialrule{1.5pt}{1pt}{1pt}
%          \emph{Naive} & 629.10 & 0.1956 && 2245.81 & 0.7896\\
%          \emph{Shift-only} & 365.56 & 0.6698 && 525.92 & 0.8944\\
%          \emph{Affine-only} & \bf 363.13 & 0.6666 && 433.80 & 0.8915\\
%          \emph{$\Vvect$-back-prop} & 411.80 & 0.1956 && 600.27 & 0.7896\\
%          \emph{Full} & 389.78 & \bf 0.7686 && \bf 424.96 & \bf 0.9135\\
%          \specialrule{1.5pt}{1pt}{1pt}
%          & 
%     \end{tabular}
%     \vspace{-0.3cm}
%     \caption{Comparison of \emph{landmark regression MAE} and \emph{foreground segmentation IoU} scores between our and baselines methods on \emph{Tai-Chi-HD} and \emph{VoxCeleb}.}
%     \label{tab:main}
%     \vspace{-0.3cm}
% \end{table}

\begin{table*}[t]
    \centering
    \def\arraystretch{0.5}
    \resizebox{\linewidth}{!}{
    \begin{tabular}{lcccc}
         \rotatebox{90}{\hspace{0.50cm}\emph{Naive}} & \includegraphics[width=0.15\columnwidth]{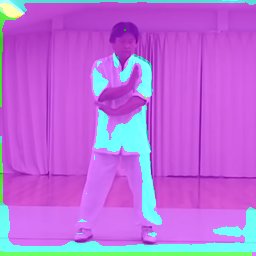}\includegraphics[width=0.15\columnwidth]{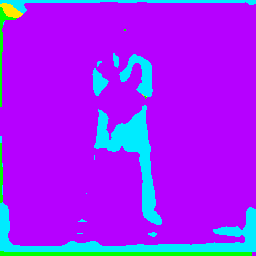} & \includegraphics[width=0.15\columnwidth]{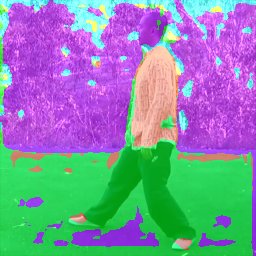}\includegraphics[width=0.15\columnwidth]{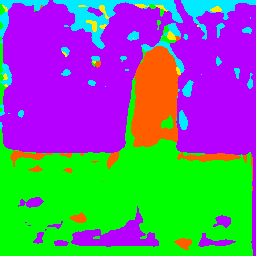} & \includegraphics[width=0.15\columnwidth]{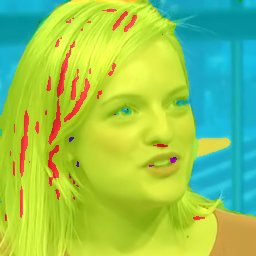}\includegraphics[width=0.15\columnwidth]{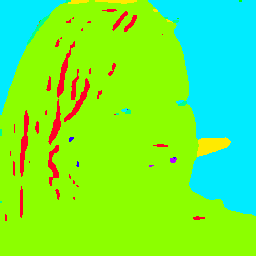} & \includegraphics[width=0.15\columnwidth]{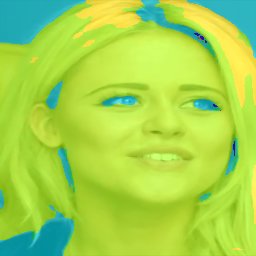}\includegraphics[width=0.15\columnwidth]{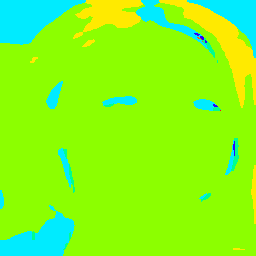} \\
         
         \rotatebox{90}{\hspace{0.25cm}\emph{Shift-only}} & \includegraphics[width=0.15\columnwidth]{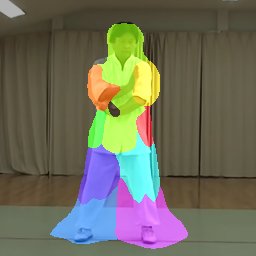}\includegraphics[width=0.15\columnwidth]{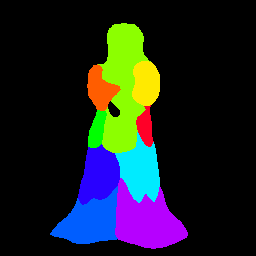} & \includegraphics[width=0.15\columnwidth]{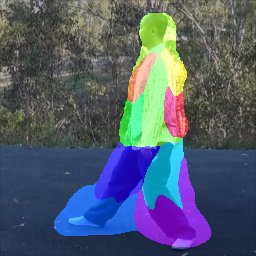}\includegraphics[width=0.15\columnwidth]{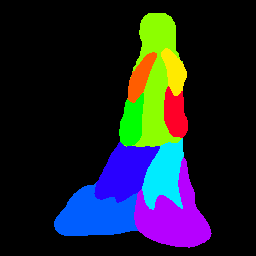} & \includegraphics[width=0.15\columnwidth]{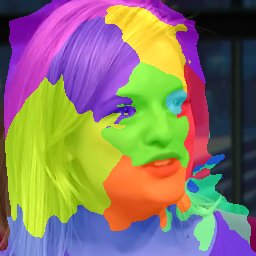}\includegraphics[width=0.15\columnwidth]{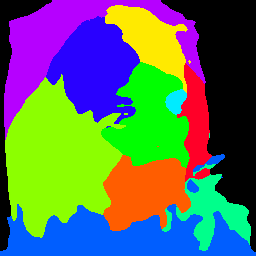} & \includegraphics[width=0.15\columnwidth]{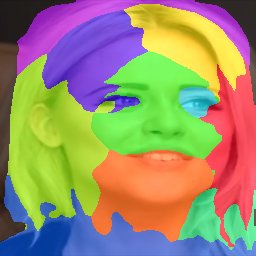}\includegraphics[width=0.15\columnwidth]{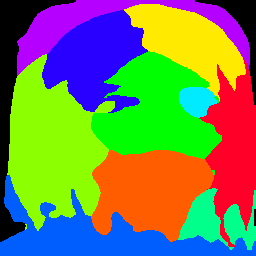} \\
         
         \rotatebox{90}{\hspace{0.15cm}\emph{Affine-only}} & \includegraphics[width=0.15\columnwidth]{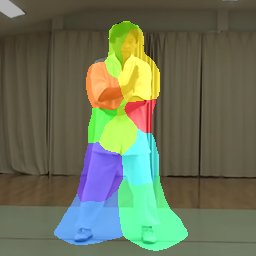}\includegraphics[width=0.15\columnwidth]{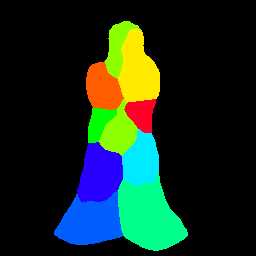} & \includegraphics[width=0.15\columnwidth]{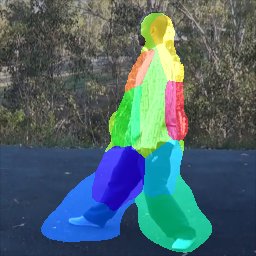}\includegraphics[width=0.15\columnwidth]{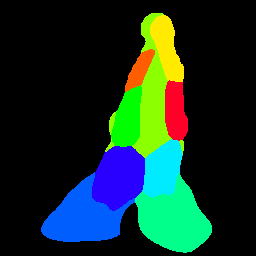} & \includegraphics[width=0.15\columnwidth]{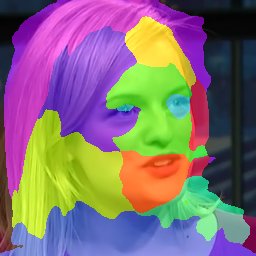}\includegraphics[width=0.15\columnwidth]{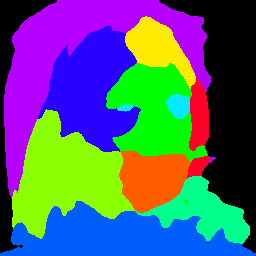} & \includegraphics[width=0.15\columnwidth]{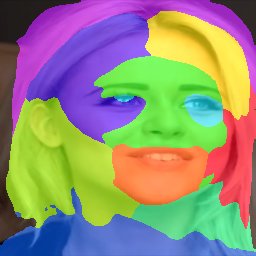}\includegraphics[width=0.15\columnwidth]{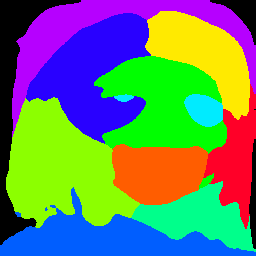} \\
         
         \rotatebox{90}{\hspace{0.05cm}\emph{$\Ovect$-back-prop}} & \includegraphics[width=0.15\columnwidth]{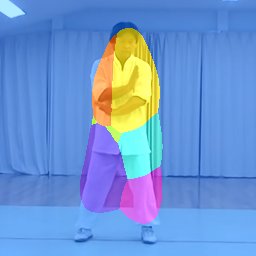}\includegraphics[width=0.15\columnwidth]{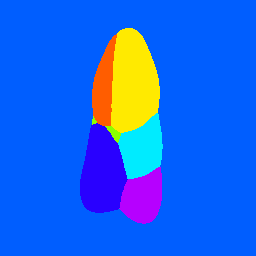} & \includegraphics[width=0.15\columnwidth]{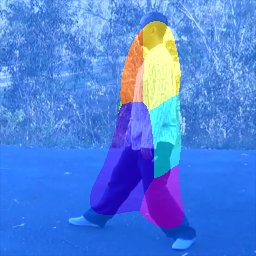}\includegraphics[width=0.15\columnwidth]{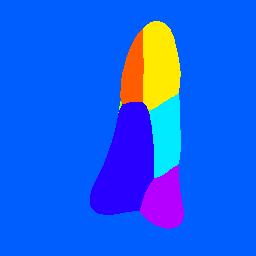} & \includegraphics[width=0.15\columnwidth]{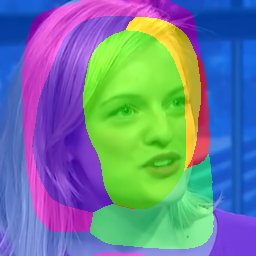}\includegraphics[width=0.15\columnwidth]{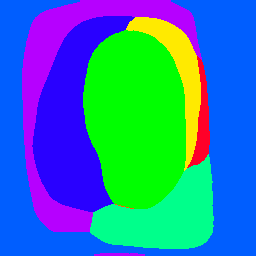} & \includegraphics[width=0.15\columnwidth]{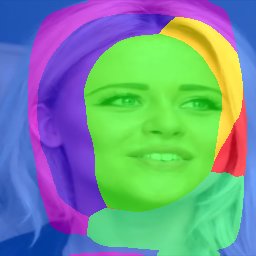}\includegraphics[width=0.15\columnwidth]{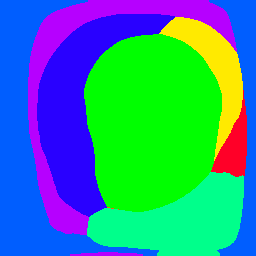} \\
         \rotatebox{90}{\hspace{0.6cm}\emph{Full}} & \includegraphics[width=0.15\columnwidth]{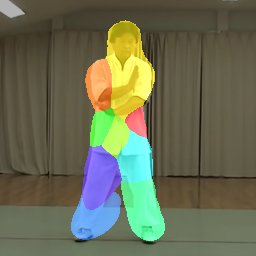}\includegraphics[width=0.15\columnwidth]{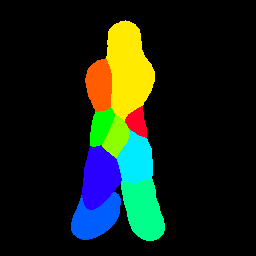} & \includegraphics[width=0.15\columnwidth]{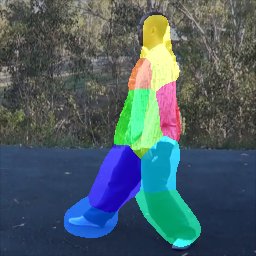}\includegraphics[width=0.15\columnwidth]{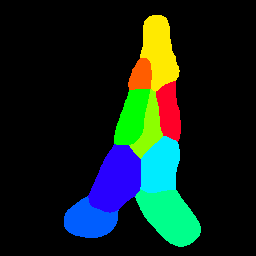} & \includegraphics[width=0.15\columnwidth]{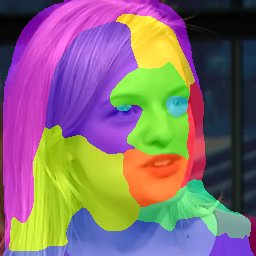}\includegraphics[width=0.15\columnwidth]{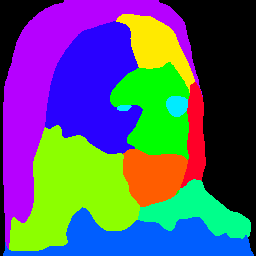} & \includegraphics[width=0.15\columnwidth]{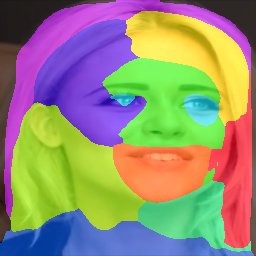}\includegraphics[width=0.15\columnwidth]{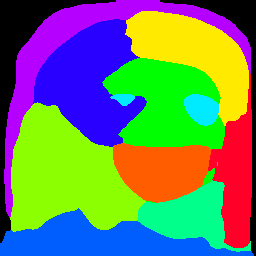} \\
    \end{tabular}}
    \captionof{figure}{\textbf{Visual comparison of our method and baselines}. In the even columns segmentation mask is depicted, while in the odd columns original image with overlayed segmentation is shown. Columns one through four corresponds to \emph{Tai-Chi-HD} dataset, and columns five through  eight show \emph{VoxCeleb} dataset.}
    \label{fig:ablations}
    \vspace{-0.7cm}
\end{table*}{}

\begin{table}[t]
    \centering
    \caption{Comparison of \emph{landmark regression MAE} and \emph{foreground segmentation IoU} scores between our and competitors methods (a). Comparison between our and baselines methods (b). Comparison of \emph{Full} model with a different number of segments on \emph{VoxCeleb} dataset (c).}
    \vspace{0.1cm}
    \resizebox{\linewidth}{!}{
    \begin{tabular}{c@{\hskip 1cm}c}
         \begin{tabular}{lcc c cc}
            \specialrule{1.5pt}{1pt}{1pt}
             & \multicolumn{2}{c}{\emph{Tai-Chi-HD}} && \multicolumn{2}{c}{\emph{VoxCeleb}}  \\
             \cline{2-3} \cline{5-6}
             Method & MAE $\downarrow$ & IoU $\uparrow$ && MAE $\downarrow$ & IoU $\uparrow$\\
             \specialrule{1.5pt}{1pt}{1pt}
             
             DFF~\cite{collins2018} & 494.48 & - && 1254.25 & - \\
             SCOPS~\cite{Hung_2019_CVPR} & 411.38 & 0.5485 && 663.04 & 0.5045\\
             Ours & \bf 389.78 & \bf 0.7686 && \bf 424.96 & \bf 0.9135 \\
             \specialrule{1.5pt}{1pt}{1pt}
        \end{tabular} &
        \multirow{4}{*}{ \begin{tabular}{lcc c cc}
        \specialrule{1.5pt}{1pt}{1pt}
         & \multicolumn{2}{c}{\emph{Tai-Chi-HD}} && \multicolumn{2}{c}{\emph{VoxCeleb}}\\
         \cline{2-3} \cline{5-6}
         Model & MAE $\downarrow$ &  IoU $\uparrow$ && MAE $\downarrow$ & IoU $\uparrow$\\
         \specialrule{1.5pt}{1pt}{1pt}
         \emph{Naive} & 629.10 & 0.1956 && 2245.81 & 0.7896\\
         \emph{Shift-only} & 365.56 & 0.6698 && 525.92 & 0.8944\\
         \emph{Affine-only} & \bf 363.13 & 0.6666 && 433.80 & 0.8915\\
         \emph{$\Vvect$-back-prop} & 411.80 & 0.1956 && 600.27 & 0.7896\\
         \emph{Full} & 389.78 & \bf 0.7686 && \bf 424.96 & \bf 0.9135\\
         \specialrule{1.5pt}{1pt}{1pt}
         &&(b)&&
         \end{tabular}}\\
         (a) & \\ \\
        \begin{tabular}{ccccc}
        \specialrule{1.5pt}{1pt}{1pt}
         & \multicolumn{4}{c}{\emph{ Number of segments}}  \\
        & 1 & 5 & 10 & 15\\
         \specialrule{1.5pt}{1pt}{1pt}
         MAE $\downarrow$ \ \  & 2177.42 \ \ & 656.63 \ \ & 424.96 \ \ & \textbf{355.38} \ \ \\
         IoU $\uparrow$ \ \ & 0.6834 \ \ & 0.9036 \ \ &  \bf 0.9135 \ \ & 0.9087 \ \\
         \specialrule{1.5pt}{1pt}{1pt} 
         \end{tabular} \\
         (c)&
    \end{tabular}}

    \label{tab:scores}
    \vspace{-0.7cm}
\end{table}

\textbf{Datasets.} We evaluate our method on two datasets: \emph{Vox-Celeb} and \emph{Tai-Chi-HD}.
\emph{VoxCeleb} is a large scale video dataset of human faces, extracted from YouTube video interviews. We use the preprocessing described in Siarohin~\etal\cite{siarohin2019neurips}. After the preprocessing, we obtain 12331 training videos and 444 test videos. The length of each video varies from 64 to 1024 frames. \emph{Tai-Chi-HD} dataset is a collection of videos. Each video depicts a person performing Tai-Chi actions. This dataset consists of 3049 training videos and 285 test videos, with length varying from 128 to 1024 frames. These datasets are selected because we require many videos containing objects of the same class but with diverse appearances.

\noindent\textbf{Evaluation Protocol.} Unsupervised part segmentation cannot be trivially evaluated with annotated segmentation or with segmentation obtained using a supervised approach. Indeed, when assigning ground-truth labels to pixels, the definition of each co-part is very subjective. Specially, depending on the application, we may prefer to define the co-part segments in a different way. Consequently, the segmentation maps obtained by the unsupervised method may not correspond to the segmentation masks that a human would draw. Therefore, in this work we follow the evaluation protocol indicated by Hung~\etal\cite{Hung_2019_CVPR} and we adopt two proxy metrics: \emph{landmark regression MAE} and \emph{foreground segmentation IoU}.

The aim of the \emph{landmark regression MAE} is to evaluate whether the extracted semantic parts are meaningful and consistent among different images. In more details, we compute the center of mass $M^k$ for each foreground segment $k$. Then we fit a linear regression model from $M^k$ to the ground truth landmarks and compute the mean average error (MAE). Following Hung~\etal\cite{Hung_2019_CVPR}, we use 5000 images for fitting the regression model, and 300 other images for computing the MAE. Since the datasets we consider come without any annotations, we use as ground truth, the landmarks computed with the method of Bulat~\etal\cite{Bulat_2017_ICCV} for \emph{VoxCeleb} and  Cao~\etal\cite{cao2017realtime} for \emph{Tai-Chi-HD}.

The other metric, \emph{Foreground segmentation IoU}, evaluates whether extracted parts separate the object from the background. For \emph{VoxCeleb} we extract the corresponding ground truth foreground using the face parser of Lie~\etal\cite{lee2019Mask}, while for \emph{Tai-Chi-HD}, we use Detectron2~\cite{wu2019detectron2} (masks corresponding to \textit{person} class). From the predicted co-part, we estimate the foreground with $\Ovect_T\!=\!\sum_{k=1}^K\Yvect_S^k$ and compute intersection over union (IoU) with the ground truth foreground.

\noindent\textbf{Baselines.} We compare our method with two state-of-the-art methods for co-part image segmentation: SCOPS~\cite{Hung_2019_CVPR} and DFF~\cite{collins2018}. To the best of our knowledge, these baselines are the only methods on unsupervised co-part segmentation based on deep architectures\footnote{Please note that we could not compare with \cite{xu2019unsupervised} since their training code was not available at the time of the submission.}.
Concerning SCOPS~\cite{Hung_2019_CVPR}, we use the implementation provided by the authors considering the same hyper-parameters provided by the authors, except the number of parts that we set to $K=10$. We train SCOPS model on \emph{Tai-Chi-HD} and \emph{VoxCeleb} by treating all the frames of all the videos as images. Importantly, SCOPS~\cite{Hung_2019_CVPR} relies on saliency maps that are obtained using the unsupervised method described in~\cite{zhu_2014_saliency}. However, the SCOPS authors' implementation only provides pre-computed saliency maps for CelebA~\cite{liu2015deep}. To obtain the saliency for our datasets, we estimate the saliency maps using \cite{zhu_2014_saliency}. Additional  details provided in Sup. Mat.~\ref{sec:baselines}.

To compare with DFF~\cite{collins2018}, we use the official implementation and employ the default parameters except for the number of parts that we set to $K=10$. One of the significant limitations of the DFF~\cite{collins2018} is the fact that matrix factorization requires all the images at once without distinction between training and test. This leads to two consequences. First, the dataset must be sub-sampled to fit in the memory. Second, train and test images should be combined for matrix factorization, thereby making it impossible to apply on a single test image. For these reasons we use all $5000 + 300$ images for DFF~\cite{collins2018} matrix factorization. See additional  details in Sup. Mat.~\ref{sec:baselines}.

\subsection{Ablation Study}
\label{sec:ablation}

We first show the results of our ablation study which demonstrate the importance of the architectural choices made in the design of our network. 

First, we implement a \emph{Naive} baseline where we try to directly reconstruct the target image from the target segmentation map $\boldsymbol{Y}_T$ and source image $\boldsymbol{X}_S$. In this case, the segmentation map $\boldsymbol{Y}_T$ is provided to the \textit{Reconstruction Module} via concatenation with the source image feature in the bottleneck layer. Then, we consider another baseline to evaluate a simplified motion representation, where transformations are only shifting and no background visibility masks are used. In this case, we consider that $\boldsymbol{A}_S = \boldsymbol{I}$, $\boldsymbol{A}_T = \boldsymbol{I}$ ($\boldsymbol{I}$ is the identity matrix) and $\Vvect = 1$. We refer to this model as \emph{Shift-only}. 
The third model uses a more complex motion representation since it includes the affine term in Eq.~\eqref{approxaffine}. However it does not use background visibility mask to reconstruct the frame. This model is referred to as \emph{Affine-only}.
Finally we introduce two baselines that use background visibility masks. In \emph{$\Vvect$-back-prop}, we propagate the gradient through the background visibility mask at the training time. Conversely, in \emph{Full}, we stop gradient propagation trough the background visibility mask as explained in Sec.~\ref{sec:recTrain}.

\begin{table*}[t]
    \centering
    \def\arraystretch{0.5}
    \resizebox{\linewidth}{!}{
    \begin{tabular}{lcccc}
         \rotatebox{90}{\hspace{0.35cm}DFF~\cite{collins2018}} & \includegraphics[width=0.15\columnwidth]{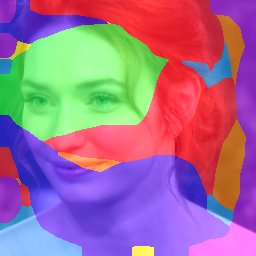}\includegraphics[width=0.15\columnwidth]{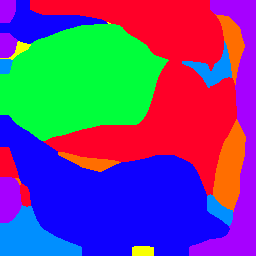} & \includegraphics[width=0.15\columnwidth]{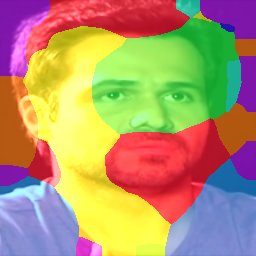}\includegraphics[width=0.15\columnwidth]{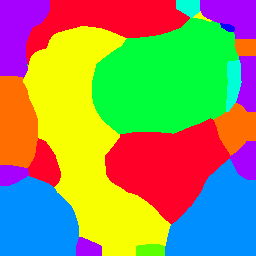} & \includegraphics[width=0.15\columnwidth]{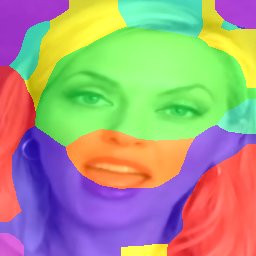}\includegraphics[width=0.15\columnwidth]{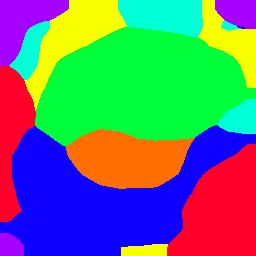} & \includegraphics[width=0.15\columnwidth]{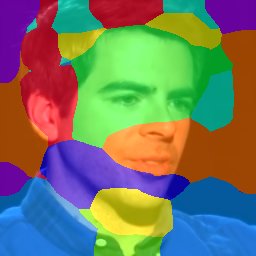}\includegraphics[width=0.15\columnwidth]{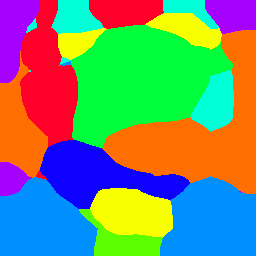} \\
         
         \rotatebox{90}{\hspace{0.05cm}SCOPS~\cite{Hung_2019_CVPR}} & \includegraphics[width=0.15\columnwidth]{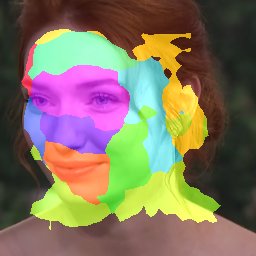}\includegraphics[width=0.15\columnwidth]{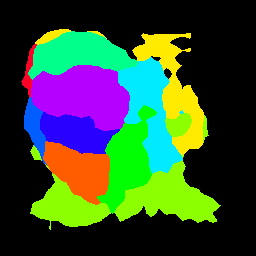} & \includegraphics[width=0.15\columnwidth]{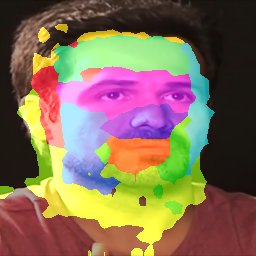}\includegraphics[width=0.15\columnwidth]{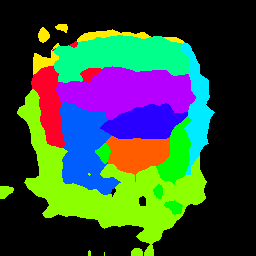} & \includegraphics[width=0.15\columnwidth]{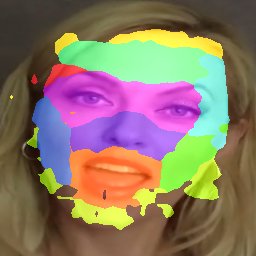}\includegraphics[width=0.15\columnwidth]{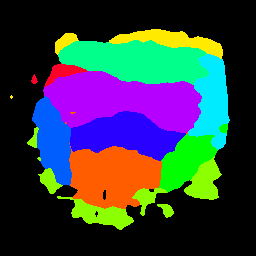} & \includegraphics[width=0.15\columnwidth]{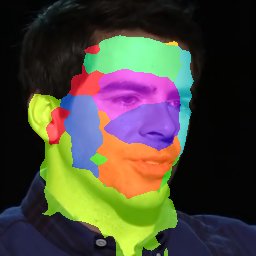}\includegraphics[width=0.15\columnwidth]{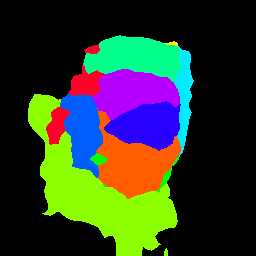}\\
         
         \rotatebox{90}{\hspace{0.55cm}Ours} & \includegraphics[width=0.15\columnwidth]{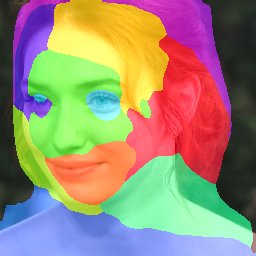}\includegraphics[width=0.15\columnwidth]{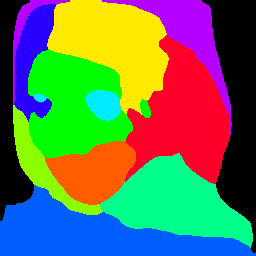} & \includegraphics[width=0.15\columnwidth]{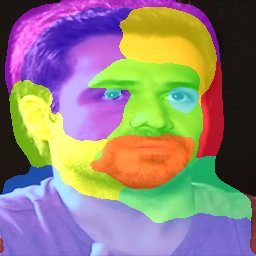}\includegraphics[width=0.15\columnwidth]{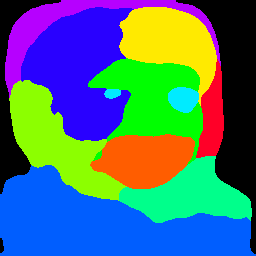} & \includegraphics[width=0.15\columnwidth]{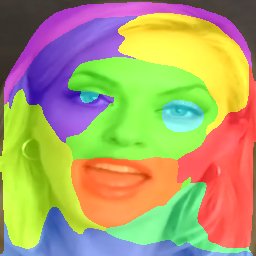}\includegraphics[width=0.15\columnwidth]{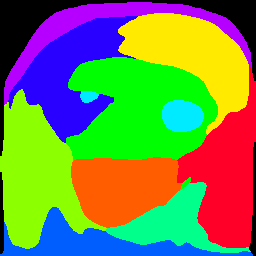} & \includegraphics[width=0.15\columnwidth]{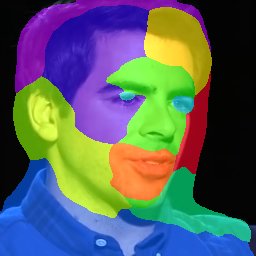}\includegraphics[width=0.15\columnwidth]{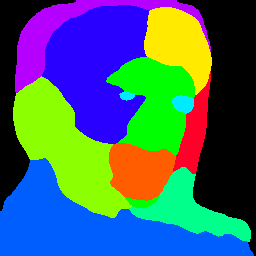}
    \end{tabular}}
    \captionof{figure}{\textbf{Visual results on \emph{VoxCeleb} dataset}. In the even columns the segmentation mask is depicted, while in the odd columns the overlayed segmentation is shown. Our method produces more meaningful face segments that cover most part of the face, when compared to the state-of-the-art methods.}
    \vspace{-0.7cm}
    \label{fig:visual_vox_sota}
\end{table*}{}

The quantitative results associated to our ablation study are reported in Tab.~\ref{tab:scores}~(b), while qualitative results can be found in Fig.~\ref{fig:ablations}. 
First, from Fig.~\ref{fig:ablations}, we clearly see that the \emph{Naive} model outputs really poor segmentation maps that do not encode any semantic information. To better understand this result, we inspect the reconstruction loss on the training set. We observe that the obtained reconstruction loss is significantly lower than the one obtained by our full model (\ie 39.63 vs 87.70 in \emph{\emph{VoxCeleb}}  dataset). This indicates that, in the case of \textit{Naive}, appearance information \emph{leaks} into the semantic map leading to really good reconstruction of the target image but very poor segmentation quality.

\begin{table*}[t]
    \centering
    \def\arraystretch{0.5}
    \resizebox{\linewidth}{!}{
    \begin{tabular}{lcccc}
         \rotatebox{90}{\hspace{0.35cm}DFF~\cite{collins2018}}  & \includegraphics[width=0.15\columnwidth]{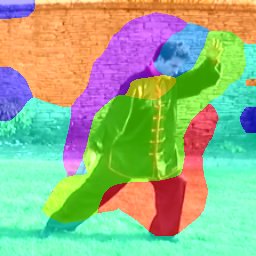}\includegraphics[width=0.15\columnwidth]{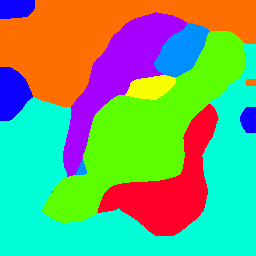} & \includegraphics[width=0.15\columnwidth]{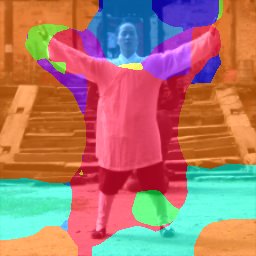}\includegraphics[width=0.15\columnwidth]{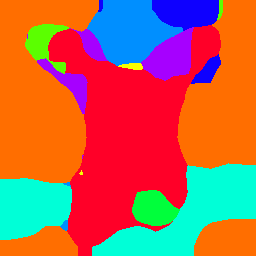} & \includegraphics[width=0.15\columnwidth]{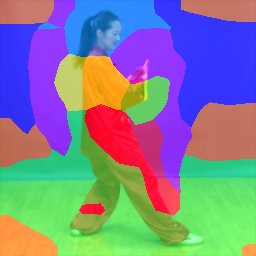}\includegraphics[width=0.15\columnwidth]{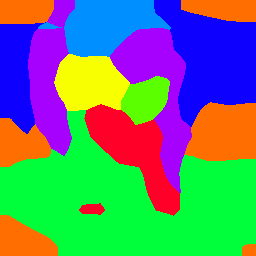} & \includegraphics[width=0.15\columnwidth]{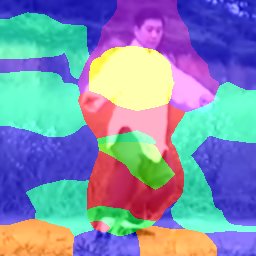}\includegraphics[width=0.15\columnwidth]{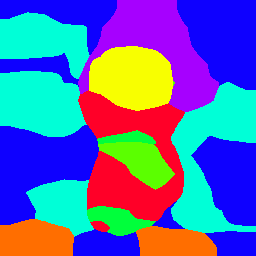} \\
        
        \rotatebox{90}{\hspace{0.05cm}SCOPS~\cite{Hung_2019_CVPR}} & \includegraphics[width=0.15\columnwidth]{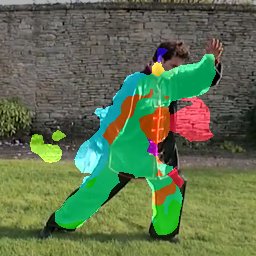}\includegraphics[width=0.15\columnwidth]{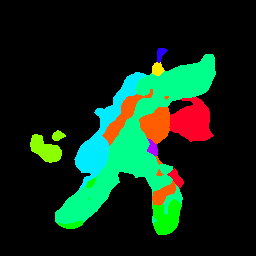} & \includegraphics[width=0.15\columnwidth]{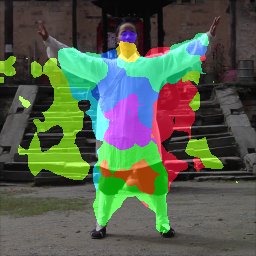}\includegraphics[width=0.15\columnwidth]{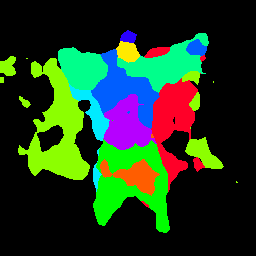} & \includegraphics[width=0.15\columnwidth]{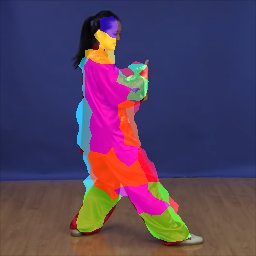}\includegraphics[width=0.15\columnwidth]{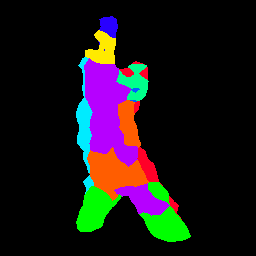} & \includegraphics[width=0.15\columnwidth]{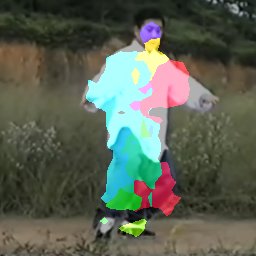}\includegraphics[width=0.15\columnwidth]{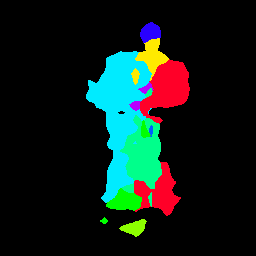} \\
        
        \rotatebox{90}{\hspace{0.55cm}Ours} & \includegraphics[width=0.15\columnwidth]{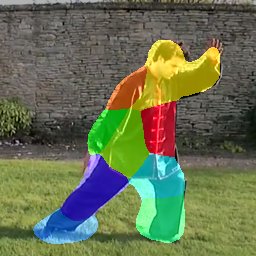}\includegraphics[width=0.15\columnwidth]{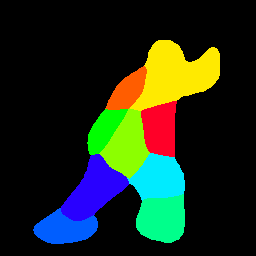} & \includegraphics[width=0.15\columnwidth]{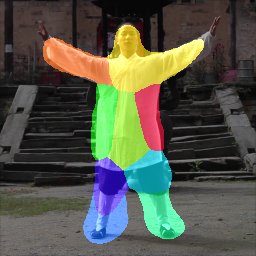}\includegraphics[width=0.15\columnwidth]{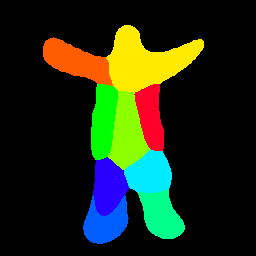} & \includegraphics[width=0.15\columnwidth]{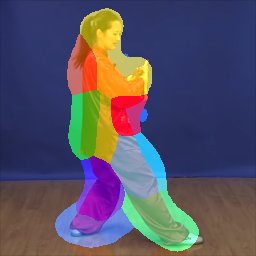}\includegraphics[width=0.15\columnwidth]{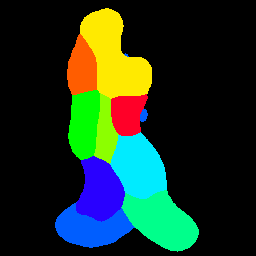} & \includegraphics[width=0.15\columnwidth]{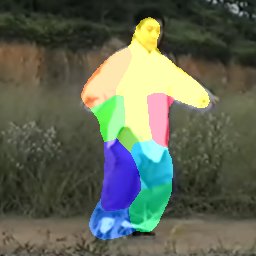}\includegraphics[width=0.15\columnwidth]{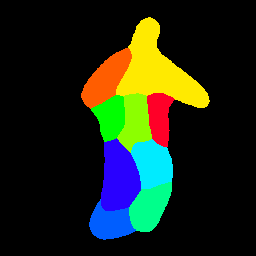} \\
    \end{tabular}
}
    \captionof{figure}{\textbf{Visual results on \emph{Tai-Chi-HD} dataset}. In the even columns the segmentation mask is depicted, while in the odd columns the overlayed segmentation is shown. Our method produces more meaningful and seamless body segments when compared to the state-of-the-art methods.}
    \label{fig:visual_taichi_sota}
    \vspace{-0.7cm}
\end{table*}{}

Compared to \emph{Naive}, we observe that \emph{Shift-only} predicts segments with clearer semantic correspondences. This shows the effectiveness of our part based motion decomposition. Nevertheless, because of simplified motion representation, we notice that the \emph{Shift-only} model can not predict precise segments (for e.g. the legs in \emph{Tai-Chi-HD}, see Fig.~\ref{fig:ablations}). Moreover, the borders of the segments are not smooth (see the red segment in the Fig.~\ref{fig:ablations} for \emph{VoxCeleb}). This critical observation is confirmed by the quantitative evaluation since we observe a relatively poor \emph{foreground segmentation IoU} score (see Tab.~\ref{tab:scores}~(a)).
When we employ a richer motion representation, corresponding to \emph{Affine-only} baseline, the extracted segments are more accurate (see the legs in the second \emph{Tai-Chi-HD}, Fig.~\ref{fig:ablations}). Quantitatively, we notice better \emph{landmark MAE} scores on both datasets.

Regarding the use of background visibility masks, we observe that the addition of the background visibility masks ($\Vvect$\emph{-back-prop}) leads to the same issue as in the \emph{Naive}. Low-level information \emph{leaks} into the semantic map. Indeed, when computing the reconstruction loss, we observe that $\Vvect$\emph{-back-prop} can reconstruct the image more accurately than our full model (46.30 vs 87.70). Finally, the \emph{Full} model corrects this issue and leads to significant improvements in the localization of the foreground object (see \emph{foreground segmentation IoU} in Tab.~\ref{tab:scores}~(a)).

In the second series of experiments, we evaluate the impact of the number of segments \textit{K}. 
%We performed this evaluation both qualitatively and quantitatively on the \emph{VoxCeleb} dataset. 
Quantitative evaluation is reported in Tab.~\ref{tab:scores}~(c).
%while the associated qualitative results are reported in Fig.~\ref{fig:segments}.
We observe that when we introduce more segments, more fine-grained details appear in the estimated segmentations. Moreover, as more parts are used, the \emph{landmark regression MAE} becomes lower. This behaviour is reasonable since predicting more segments leads to a richer description of the object structure. However, concerning \emph{foreground segmentation IoU}, we observe that the model with 5 segments already acceptably separates foreground and background and the \emph{foreground segmentation IoU} does not improve significantly when more segments are added (see Tab.~\ref{tab:scores}~(c)). For visual results please refer to Sup. Mat.~\ref{sec:segments}.

% \begin{table}[t]
%     \centering
%     \begin{tabular}{ccccc}
%         \specialrule{1.5pt}{1pt}{1pt}
%          & \multicolumn{4}{c}{\emph{ Number of segments}}  \\
%         & 1 & 5 & 10 & 15\\
%          \specialrule{1.5pt}{1pt}{1pt}
%          MAE $\downarrow$ \ \  & 2177.42 \ \ & 656.63 \ \ & 424.96 \ \ & 355.38 \ \ \\
%          IoU $\uparrow$ \ \ & 0.6834 \ \ & 0.9036 \ \ &  \bf 0.9135 \ \ & 0.9087 \ \\
%          \specialrule{1.5pt}{1pt}{1pt}
%          & 
%     \end{tabular}
%     \vspace{-0.4cm}
%     \caption{Comparison of \emph{landmark regression MAE} and \emph{foreground segmentation IoU} scores for a different number of segments on \emph{VoxCeleb} dataset.}
%     \label{tab:segments}
%     \vspace{-0.7cm}
% \end{table}{}

\subsection{Comparison with State of the Art}

We compare our method with two previous approaches for co-part segmentation: SCOPS~\cite{Hung_2019_CVPR} and DFF~\cite{collins2018}. 
Quantitative results are reported in Tab.~\ref{tab:scores}~(a), while qualitative comparison is shown in  Fig.~\ref{fig:visual_vox_sota} and Fig.~\ref{fig:visual_taichi_sota}. From the figures it is easy to observe that DFF is not able to localize the foreground object. Furthermore, many predicted segmentation maps are fragmented. Compared to DFF, our method produces semantic parts which are more consistent over different object and where the main object is clearly separated from the background. Note that since DFF~\cite{collins2018} has no explicit way of handling the background, we do not report \emph{foreground segmentation IoU} for it. In terms of \emph{landmark regression MAE}, the benefit of our approach is clearly evident (see Tab.~\ref{tab:scores}~(b)). 

Concerning SCOPS~\cite{Hung_2019_CVPR}, in case of \emph{Tai-Chi-HD}, the predicted parts are not consistent in the different images and large portions of the background are misclassified. Differently in our method, only small portions of the background near to the object boundaries are included in the foreground segments. Note however, that performance of SCOPS~\cite{Hung_2019_CVPR} heavily depends on unsupervised saliency prediction method~\cite{zhu_2014_saliency}.   Overall, the segments from our method are much more consistent and semantically meaningful. The better performance of our method are also confirmed by quantitative results (see Tab.~\ref{tab:scores}~(b)). For both datasets and both metrics, we obtain consistently better scores. The gain of our method is especially evident on the \emph{VoxCeleb} dataset. For additional visual results please refer to Sup. Mat.~\ref{sec:compa} and our supplementary video\footnote{\href{https://youtu.be/RJ4Nj1wV5iA}{https://youtu.be/RJ4Nj1wV5iA}} for more qualitative result.

% \begin{table}[t]
%     \centering
%     \begin{tabular}{lcc c cc}
%         \specialrule{1.5pt}{1pt}{1pt}
%          & \multicolumn{2}{c}{\emph{Tai-Chi-HD}} && \multicolumn{2}{c}{\emph{VoxCeleb}}  \\
%          \cline{2-3} \cline{5-6}
%          Method & MAE $\downarrow$ & IoU $\uparrow$ && MAE $\downarrow$ & IoU $\uparrow$\\
%          \specialrule{1.5pt}{1pt}{1pt}
         
%          DFF~\cite{collins2018} & 494.48 & - && 1254.25 & - \\
%          SCOPS~\cite{Hung_2019_CVPR} & 411.38 & 0.5485 && 663.04 & 0.5045\\
%          Ours & \bf 389.78 & \bf 0.7686 && \bf 424.96 & \bf 0.9135 \\
%          \specialrule{1.5pt}{1pt}{1pt}
%          & 
%     \end{tabular}
%     \vspace{-0.3cm}
%     \caption{Comparison of \emph{landmark regression MAE} and \emph{foreground segmentation IoU} scores between our and competitors methods on \emph{Tai-Chi-HD} and \emph{VoxCeleb}.}
%     \label{tab:scores
%     \vspace{-0.3cm}
% \end{table}{}

\begin{table*}[t]
    \centering
    \def\arraystretch{0.5}
    \resizebox{\linewidth}{!}{
    \begin{tabular}{lcc}
         &\includegraphics[width=0.15\columnwidth]{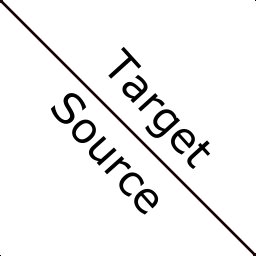}\includegraphics[width=0.15\columnwidth]{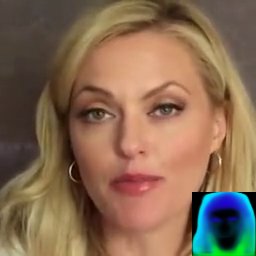}\includegraphics[width=0.15\columnwidth]{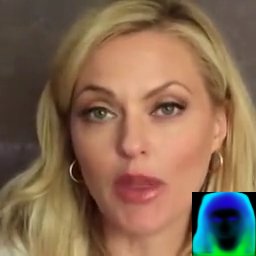}\includegraphics[width=0.15\columnwidth]{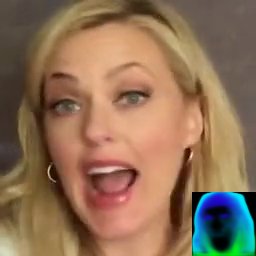} & \includegraphics[width=0.15\columnwidth]{figures/vox/swaps/white}\includegraphics[width=0.15\columnwidth]{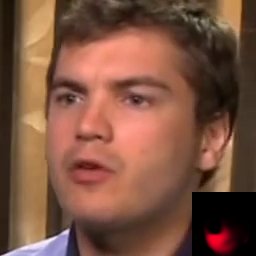}\includegraphics[width=0.15\columnwidth]{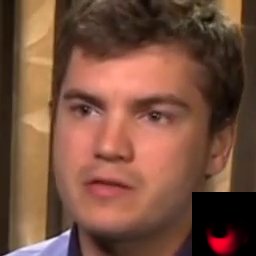}\includegraphics[width=0.15\columnwidth]{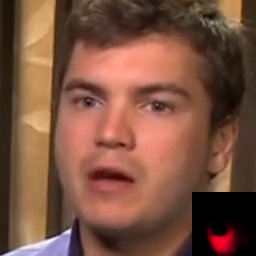} \\
         \rotatebox{90}{\hspace{0.15cm}\emph{5 segments}}&\includegraphics[width=0.15\columnwidth]{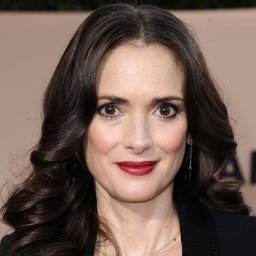}\includegraphics[width=0.15\columnwidth]{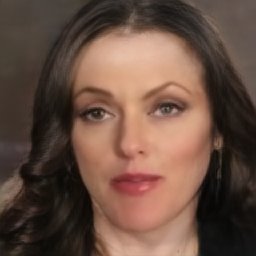}\includegraphics[width=0.15\columnwidth]{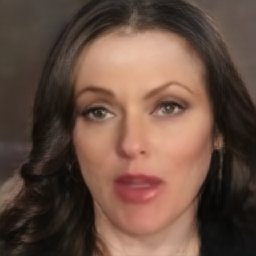}\includegraphics[width=0.15\columnwidth]{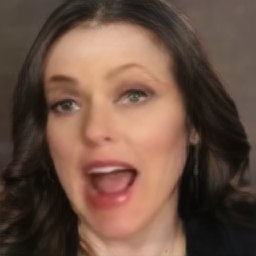} & \includegraphics[width=0.15\columnwidth]{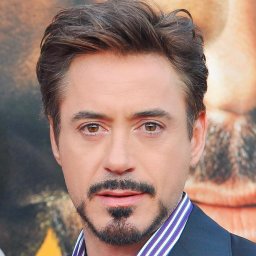}\includegraphics[width=0.15\columnwidth]{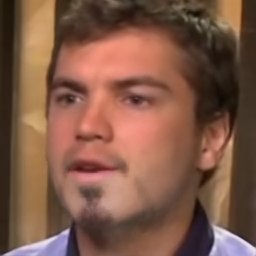}\includegraphics[width=0.15\columnwidth]{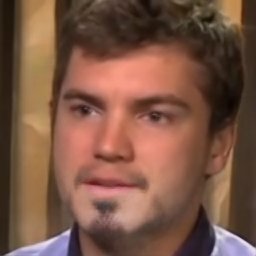}\includegraphics[width=0.15\columnwidth]{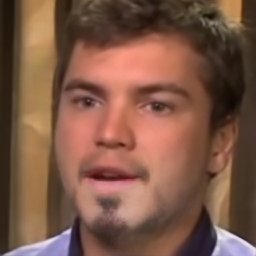} \\
         
         &\includegraphics[width=0.15\columnwidth]{figures/vox/swaps/white}\includegraphics[width=0.15\columnwidth]{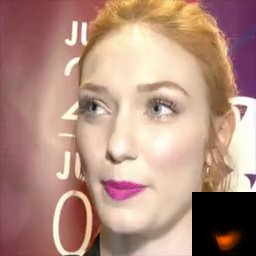}\includegraphics[width=0.15\columnwidth]{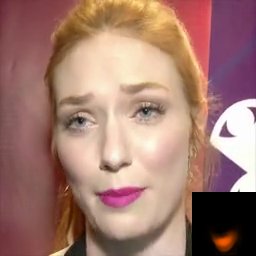}\includegraphics[width=0.15\columnwidth]{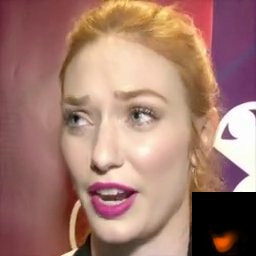} & \includegraphics[width=0.15\columnwidth]{figures/vox/swaps/white}\includegraphics[width=0.15\columnwidth]{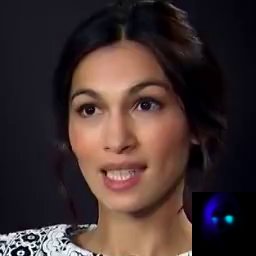}\includegraphics[width=0.15\columnwidth]{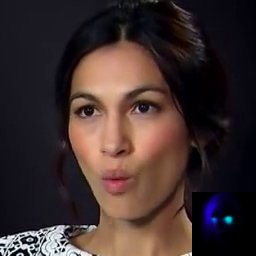}\includegraphics[width=0.15\columnwidth]{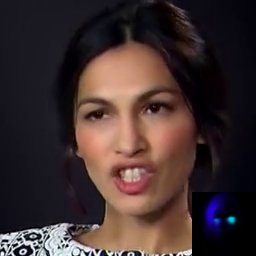} \\
         \rotatebox{90}{\hspace{0.05cm}\emph{10 segments}}&\includegraphics[width=0.15\columnwidth]{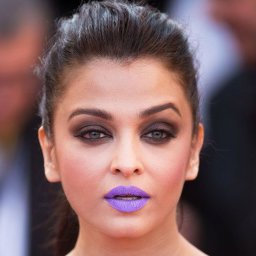}\includegraphics[width=0.15\columnwidth]{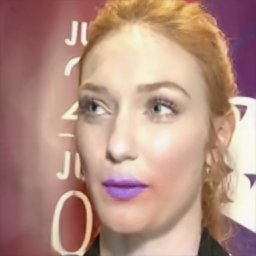}\includegraphics[width=0.15\columnwidth]{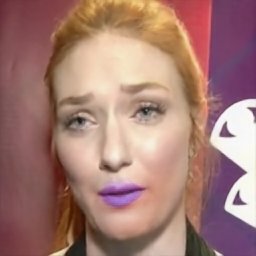}\includegraphics[width=0.15\columnwidth]{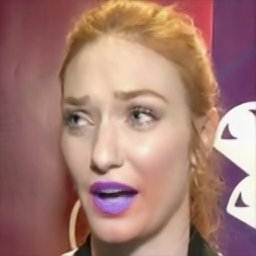} & \includegraphics[width=0.15\columnwidth]{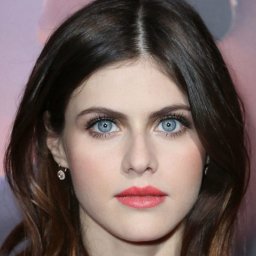}\includegraphics[width=0.15\columnwidth]{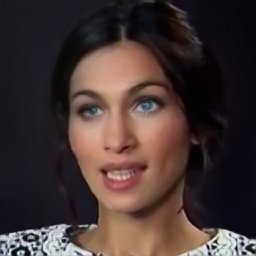}\includegraphics[width=0.15\columnwidth]{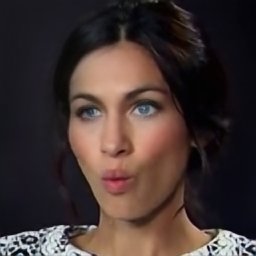}\includegraphics[width=0.15\columnwidth]{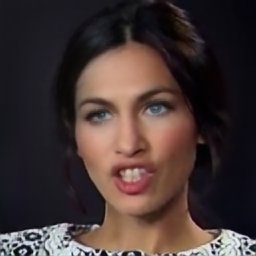} \\
    \end{tabular}}
    
    \captionof{figure}{\textbf{Visual results of video-editing}. In the odd rows the target frames from video sequences are depicted, alongside the masks of interest (in the right bottom corners) intended to be swapped. In the even columns the source image is shown, followed by the generated frames containing the indicated parts swapped from the source. Results achieved with models for \textit{K}=5 and \textit{K}=10 are depicted. Best viewed with digital zoom.}
    \label{fig:visual}
    \vspace{-0.7cm}
\end{table*}{}

\subsection{Application: Face Part Swapping}

Unsupervised discovery of body parts allows us to use our model for interesting media applications such as \textit{video-editing}, \textit{face-swapping}, \textit{face-editing} and so on, without having to rely on externally trained parsers, unlike~\cite{lee2019Mask}. Specifically, in video-editing the user can select one or multiple semantic parts in a video sequence and swap them with the corresponding semantic parts derived from another source image. In Fig.~\ref{fig:visual} we show the results for video-editing obtained with our method, by operating in a completely unsupervised setting. 

In details, the video-editing results are generated on a frame-by-frame basis, wherein our segmentation module predicts the semantic masks for both the target frame and source image, along with the flow $\mathcal{F}$ for each mask. This part-based $\mathcal{F}$ is then used to align the source features with the corresponding target features, by deforming the former. The mask of interest to be swapped (e.g. \textit{lower jaw}) is then utilized to discard the \textit{lower jaw} features from the target frame and fill the void with the relevant features from the source image. The resulting intermediate features are then decoded by our reconstruction module to obtain the resulting image where the \textit{lower jaw} appears to be swapped from the source image (see Fig.~\ref{fig:visual} top right). Since the results of video-editing depends on the application of interest, we tested with two models that are trained to predict semantic parts with \textit{K}=5 and \textit{K}=10, respectively. We use \textit{K}=5 when we want to change large facial parts such as \textit{blonde hair} $\rightarrow$ \textit{black hair} and \textit{clean shave} $\rightarrow$ \textit{goatee beard} (see Fig.~\ref{fig:visual} top). On the other hand, we accomplish a fine-grained editing by using a model with \textit{K}=10 where we alter only small parts of the face, as in \textit{red lips} $\rightarrow$ \textit{purple lips} and \textit{black eyes} $\rightarrow$ \textit{turquoise eyes} (see Fig.~\ref{fig:visual} bottom). Note that, we used continuous masks for video-editing application to obtain a smoother transition between regions and to avoid an abrupt change in skin tone. Please see Sup. Mat.~\ref{sec:part-swap} and supplementary video for additional video-editing results.


\section{Conclusions}

We presented a novel self-supervised approach for co-part segmentation that leverages motion information automatically extracted from video streams in order to predict better segmentation maps from still images. We discussed how the critical issue for developing a self-supervised reconstruction-based model for segmentation lies into the \emph{leaking} issues, and thus, we introduced a novel part-based motion formulation that permits to overcome this problem. Additionally, we add background visibility mask to our model, thus promoting the better separation of the object with respect to the background. Through an extensive experimental evaluation and comparisons with state of the art methods, we demonstrate the validity and superiority of our approach.

\clearpage
\renewcommand{\thesection}{\Alph{section}}
\setcounter{section}{0}

%In this supplementary material, first we provide details about the implementation of our method and the state-of-the-art methods (SCOPS~\cite{Hung_2019_CVPR} and DFF~\cite{collins2018}). In particular, in Sec~\ref{sec:imp-det} we explain the details of the network architectures used in our method. Later in Sec.~\ref{sec:baselines} we explain how we trained and evaluated SCOPS~\cite{Hung_2019_CVPR} and DFF~\cite{collins2018} on each dataset. Then, in Sec.~\ref{sec:segments}, we provide an additional qualitative results for the experiment where we evaluate the impact of the number of segments (referred to as $K$ in the main paper). In Sec.~\ref{sec:compa} we extend the qualitative comparison with the state of the art methods. Finally we provide an additional results for part-swap application in Sec.~\ref{sec:part-swap}, which is also provided in supplementary video. 
%Additionally, this video contains qualitative results for segmentation. It is to be noted that segmentation and part-swap results is produced independently for each frame of the video.

\section{Implementation details}
\label{sec:imp-det}

As stated in the main documents, there are 2 main modules in our method: \textit{Segmentation module} and \textit{Reconstruction module}. 

For the \textit{Segmentation module}, we employ the U-Net~\cite{ronneberger2015u} architecture. Our \textit{Segmentation Module} operates on resolution $H'\!\times\!W'\!=\!64\!\times\!64$ and is composed of five $conv_{3\times3}$ - $bn$ - $relu$ - $avg-pool_{2\times2}$ blocks and five $upsample_{2\times2}$ - $conv_{3\times3}$ - $bn$ - $relu$ blocks. As explained in the main paper, \textit{Segmentation module} estimate shift parameters $p^k$ and affine parameters $\Yvect^k$, along with segmentation map $\Yvect^k$. The shift parameters $p^k$ can be understood as anchor keypoints associated to the segments $\Yvect^k$. Practically, the anchor keypoints are estimated by the encoder network that outputs $K$ additional channels, one per keypoint. From each channel, the anchor keypoint location is estimated via \textit{soft-argmax} as in ~\cite{siarohin2018animating,robinson2019laplace,siarohin2019neurips}. 
%By introducing the matrices $\Avect_S^k$ and ${\Avect_T^k}$, the motion within $\mathcal{Y}_T^k$ of  $p_T^k\in\mathbb{R}^{2}$ is modeled by an affine transformation, thus accounting for transformations that cannot be obtained by simple pixel-shift.
On the other hand, $\Avect^k$ is computed by using four additional outputs. From these 4-channel tensors, we obtain the coefficients of the matrix $\Avect^k$ by computing spatial weighted average using as weights the confidence map of the corresponding keypoint as in~\cite{siarohin2019neurips}. In the end, the encoder outputs a total of $6K\!+\!1$ channels: $K\!+\!1$ for the segmentation, $K$ for the anchor keypoints and $4K$ for the $\Avect^k$ matrices.

For the \textit{Reconstruction Module} we use the architecture of Johnson~\etal\cite{johnson2016perceptual} that contains two down-sampling blocks, six residual blocks and two up-sampling blocks. At train time the reconstruction loss is computed at four different resolutions $256\times256$, $128\times128$, $64\times64$ and $32\times32$ as in \cite{tang2018dual,siarohin2019neurips}. We train the system using Adam~\cite{kingma2014adam} optimizer with learning rate 2$e$-4 and batch size 20 for approximately 10k iterations on 2 TitanX GPUs. In all the experiments we use $K = 10$.

\subsection{State-of-the-art comparison}
\label{sec:baselines}

As explained in the main paper, we compare our method with two state-of-the-art methods for co-part image segmentation: SCOPS~\cite{Hung_2019_CVPR} and DFF~\cite{collins2018}. 

Concerning SCOPS~\cite{Hung_2019_CVPR}, we use the implementation provided by the authors\footnote{\href{https://github.com/NVlabs/SCOPS}{https://github.com/NVlabs/SCOPS}}. We use the hyper-parameters provided by the authors, except the number of parts that we set to $K=10$. We train SCOPS model on \emph{Tai-Chi-HD}~\cite{siarohin2019neurips} and \emph{VoxCeleb}~\cite{Nagrani17} by treating all the frames of all the videos as images. Importantly, SCOPS~\cite{Hung_2019_CVPR} relies on saliency maps that are obtained using the unsupervised method described in~\cite{zhu_2014_saliency}. However, the SCOPS authors' implementation only provides pre-computed saliency maps for CelebA~\cite{liu2015deep}. To obtain the saliency for our datasets, we estimate the saliency maps using the public implementation\footnote{\href{https://github.com/yhenon/pyimgsaliency}{https://github.com/yhenon/pyimgsaliency}} of \cite{zhu_2014_saliency}.

With regards to DFF~\cite{collins2018}, we use the official implementations provided by the authors\footnote{\href{https://github.com/edocollins/DFF}{https://github.com/edocollins/DFF}}. We also employed the default parameters except for the number of parts that we set to $K=10$. One of the significant limitations of the DFF~\cite{collins2018} is the fact that matrix factorization requires all the images at once without distinction between training and test. This leads to two consequences. First, the dataset must be sub-sampled to fit in the memory. Second, train and test images should be combined for matrix factorization, thereby making it impossible to apply on a single test image. For these reasons we use all $5000 + 300$ (see main paper Sec.~4) images for DFF~\cite{collins2018} matrix factorization.

In our preliminary experiments, we also implemented a video-based version of DFF~\cite{collins2018} that performs matrix factorization using all the frames from a single video. The motivation for this experiment is to compare our method with an approach that uses video information. Nevertheless, this comparison is not completely fair since this video-based DFF estimates segmentation in the test frames by combining information from several frames. We include this video-based implementation of DFF for the comparison in the supplementary video.  Despite evaluation biased in favour of DFF, we can observe that our method produces clearly better visual results than this video-based DFF. Our approach obtains much more consistent segments with seamless boundaries. 

Furthermore, video-based DFF must be run independently for each video to prevent memory issues. This leads to segmentation that are not consistent among videos, as we can see in the supplementary video. Consequently, this DFF version could not be included in our quantitative evaluation that uses landmark-base metrics. Indeed, to compute this metric, we train and evaluate a linear landmark regression model on two different subsets without identity overlap. In the case of the video-based DFF, this procedure should be applied for every run of the method, \textit{ie.} every video. For this reason, subsets without overlapping identities cannot be used.

\section{Qualitative results: number of segments}
\label{sec:segments}

In this section, we report qualitative results for the ablation study where we evaluate the impact of the number of segments (see Fig.~\ref{fig:segments}).

%evaluation is performed both qualitatively and quantitatively on the \emph{VoxCeleb} dataset. First, quantitative evaluation is reported in Tab.~\ref{tab:segments}. The qualitative evaluation is reported in Fig.~\ref{fig:segments}.

\begin{table*}[!h]
    \centering
    \def\arraystretch{0.9}
    \resizebox{\linewidth}{!}{
    \begin{tabular}{lcccc}
         \rotatebox{90}{\hspace{0.2cm}\emph{1 segment}} & \includegraphics[width=0.15\columnwidth]{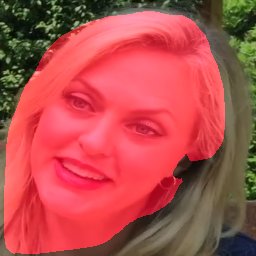}\includegraphics[width=0.15\columnwidth]{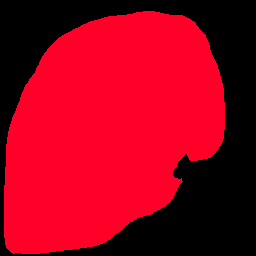}  & \includegraphics[width=0.15\columnwidth]{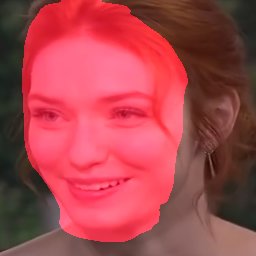}\includegraphics[width=0.15\columnwidth]{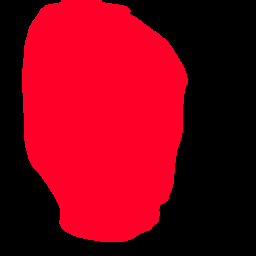} & \includegraphics[width=0.15\columnwidth]{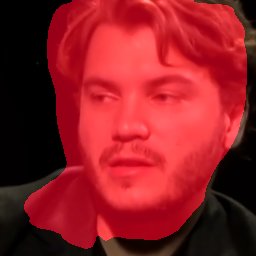}\includegraphics[width=0.15\columnwidth]{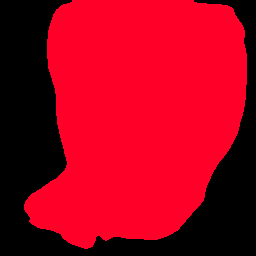} & \includegraphics[width=0.15\columnwidth]{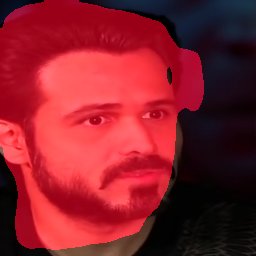}\includegraphics[width=0.15\columnwidth]{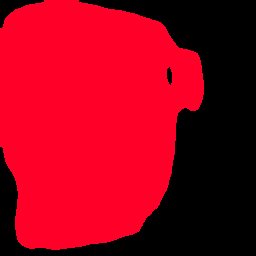}\\
         
         \rotatebox{90}{\hspace{0.16cm}\emph{5 segments}} & \includegraphics[width=0.15\columnwidth]{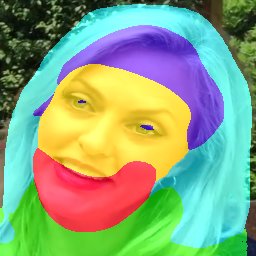}\includegraphics[width=0.15\columnwidth]{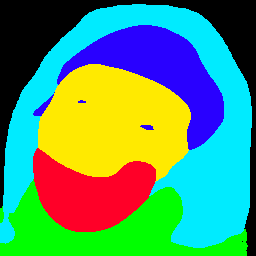}  & \includegraphics[width=0.15\columnwidth]{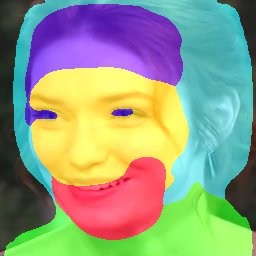}\includegraphics[width=0.15\columnwidth]{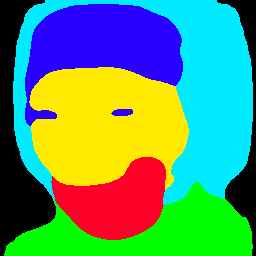} & \includegraphics[width=0.15\columnwidth]{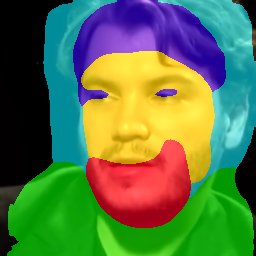}\includegraphics[width=0.15\columnwidth]{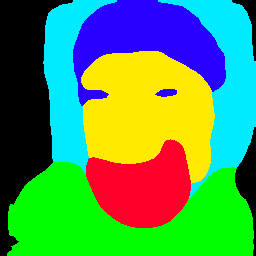} & \includegraphics[width=0.15\columnwidth]{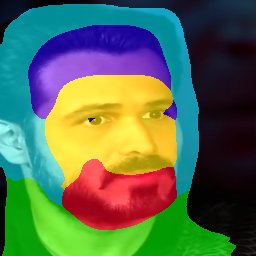}\includegraphics[width=0.15\columnwidth]{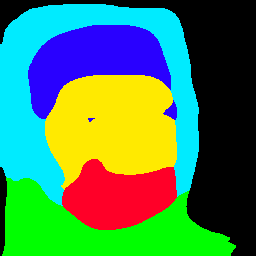}\\
         
         \rotatebox{90}{\hspace{0.00cm}\emph{10 segments}} & \includegraphics[width=0.15\columnwidth]{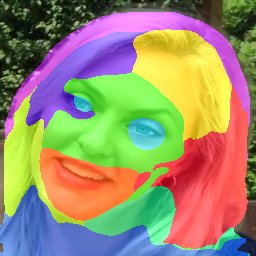}\includegraphics[width=0.15\columnwidth]{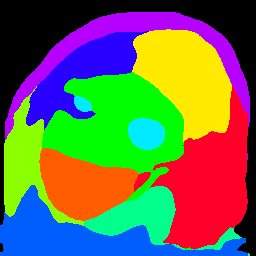}  & \includegraphics[width=0.15\columnwidth]{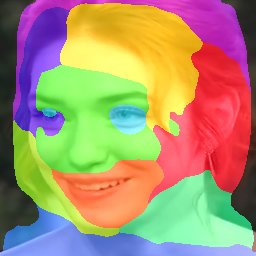}\includegraphics[width=0.15\columnwidth]{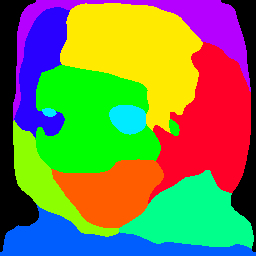} & \includegraphics[width=0.15\columnwidth]{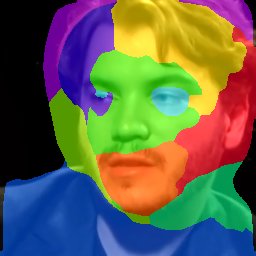}\includegraphics[width=0.15\columnwidth]{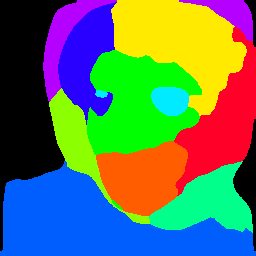} & \includegraphics[width=0.15\columnwidth]{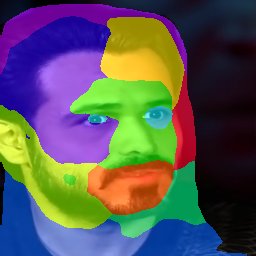}\includegraphics[width=0.15\columnwidth]{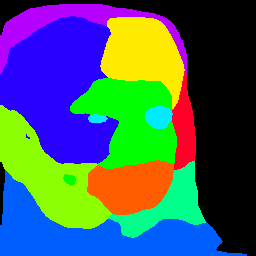}\\
         
         \rotatebox{90}{\hspace{0.00cm}\emph{15 segments}} & \includegraphics[width=0.15\columnwidth]{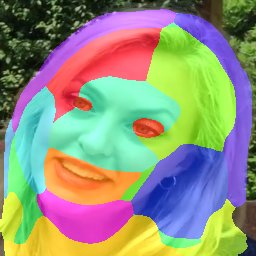}\includegraphics[width=0.15\columnwidth]{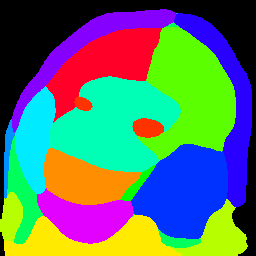}  & \includegraphics[width=0.15\columnwidth]{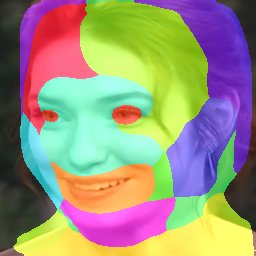}\includegraphics[width=0.15\columnwidth]{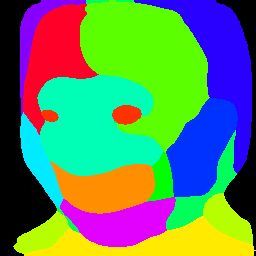} & \includegraphics[width=0.15\columnwidth]{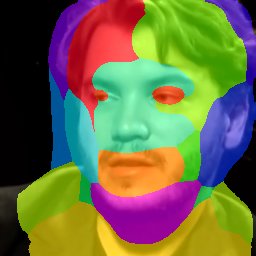}\includegraphics[width=0.15\columnwidth]{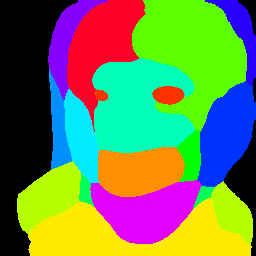} & \includegraphics[width=0.15\columnwidth]{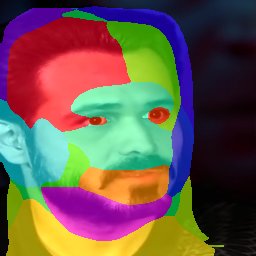}\includegraphics[width=0.15\columnwidth]{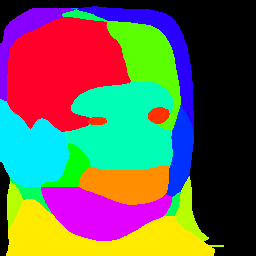}\\
         
    \end{tabular}
    }
    \captionof{figure}{\textbf{Visual comparison of our method with different number of segments on the \emph{VoxCeleb} dataset}. In the even columns, predicted segmentations are depicted, while in the odd columns, the original images with overlayed segmentation are shown.}
    \label{fig:segments}
\end{table*}{}

%We observe that when we introduce more segments, more fine-grained details appear in the estimated segmentations (see Fig.~\ref{fig:segments}). Moreover, as more parts are used, the \emph{landmark regression MAE} becomes lower. This behaviour is reasonable since predicting more segments leads to a richer description of the object structure. However, concerning \emph{foreground segmentation IoU}, we observe in Fig.~\ref{fig:segments} that the model using 5 segments already acceptably separates foreground and background. The \emph{foreground segmentation IoU} does not improve significantly when more segments are added (see Tab.~\ref{tab:segments}).

% \begin{table}[!h]
%     \centering
%     \begin{tabular}{ccc}
%         \specialrule{1.5pt}{1pt}{1pt}
%          & \multicolumn{2}{c}{\emph{VoxCeleb}}  \\
%          Number of segments & MAE $\downarrow$ & IoU $\uparrow$ \\
%          \specialrule{1.5pt}{1pt}{1pt}
         
%          1 & 2177.42 & 0.6834 \\
%          5 & 656.63 & 0.9036 \\
%          10 & 424.96 & \bf 0.9135 \\
%          15 & \bf 355.38 & 0.9087 \\
%          \specialrule{1.5pt}{1pt}{1pt}
%          & 
%     \end{tabular}
%     \vspace{-0.4cm}
%     \caption{Comparison of \emph{landmark regression MAE} and \emph{foreground segmentation IoU} scores for a different number of segments.}
%     \label{tab:segments}
%     \vspace{-0.4cm}
% \end{table}{}

\section{Qualitative comparison with the state-of-the-art}
\label{sec:compa}

In this section, we extend the qualitative comparison with state of the art presented in the main paper (see Fig.~4 of the main paper). In Figs.~\ref{fig:taichi-sota},~\ref{fig:vox-sota}, we provide an additional comparison of the predicted co-part segmentation obtained with DFF, SCOPS and our methods on both the \emph{Tai-Chi-HD} and the \emph{VoxCeleb} datasets. These results are well in-line with the results reported in the main paper. We observe that on both datasets, our method outputs segments much more consistent across different images. Furthermore, the boundaries between the segments are cleaner and the foreground is better separated from the background. Similar observations can also be made from the video attached to this supplementary material. In this video, we also observe that our method outputs stable (over time) segmentations even given that these segmentations are estimated independently for each frame.

\begin{table*}[h!]
    \centering
    \def\arraystretch{0.9}
    \resizebox{\linewidth}{!}{
    \begin{tabular}{cccc}
        Input & DFF~\cite{collins2018} & SCOPS~\cite{Hung_2019_CVPR} & Ours\\\\
        \includegraphics[width=0.23\columnwidth]{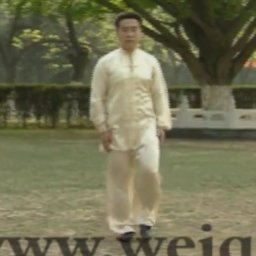} & \includegraphics[width=0.23\columnwidth]{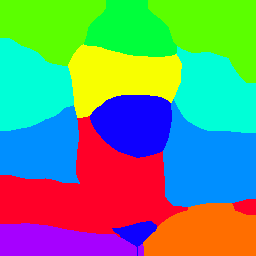}\includegraphics[width=0.23\columnwidth]{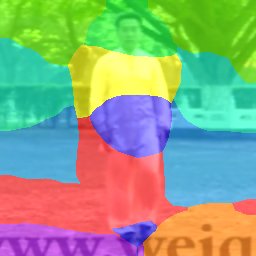} & \includegraphics[width=0.23\columnwidth]{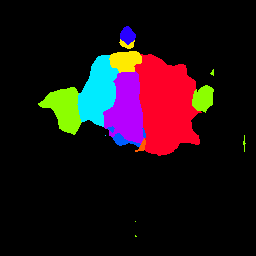}\includegraphics[width=0.23\columnwidth]{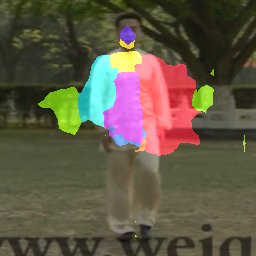} & \includegraphics[width=0.23\columnwidth]{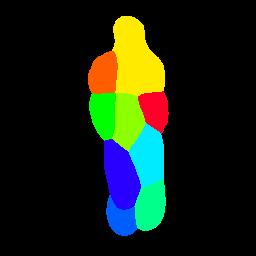}\includegraphics[width=0.23\columnwidth]{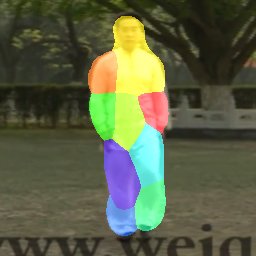}\\
        \includegraphics[width=0.23\columnwidth]{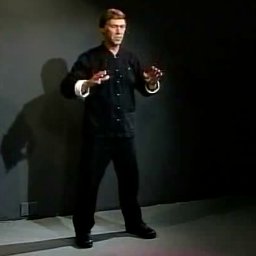} & \includegraphics[width=0.23\columnwidth]{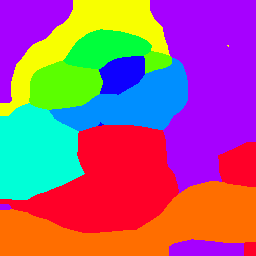}\includegraphics[width=0.23\columnwidth]{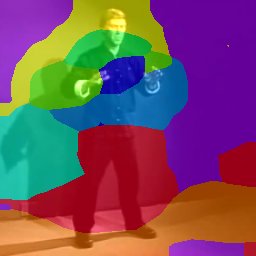} & \includegraphics[width=0.23\columnwidth]{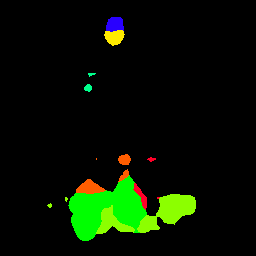}\includegraphics[width=0.23\columnwidth]{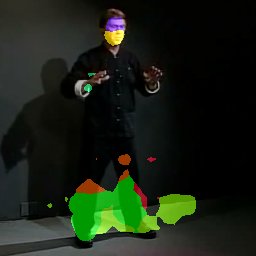} & \includegraphics[width=0.23\columnwidth]{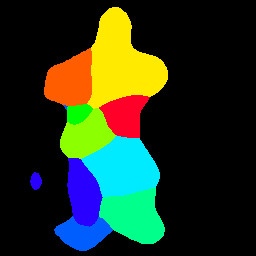}\includegraphics[width=0.23\columnwidth]{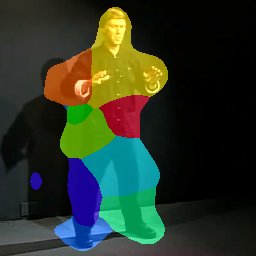}\\
        
        \includegraphics[width=0.23\columnwidth]{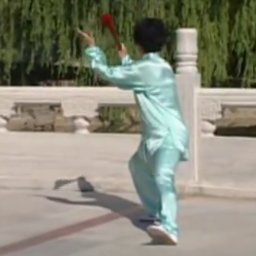} & \includegraphics[width=0.23\columnwidth]{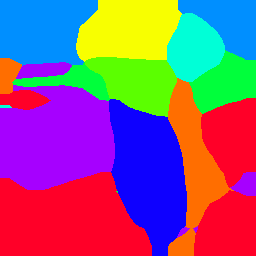}\includegraphics[width=0.23\columnwidth]{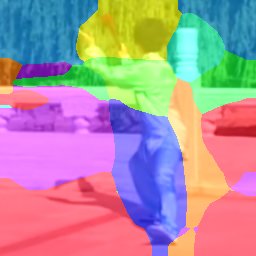} & \includegraphics[width=0.23\columnwidth]{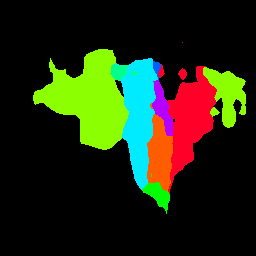}\includegraphics[width=0.23\columnwidth]{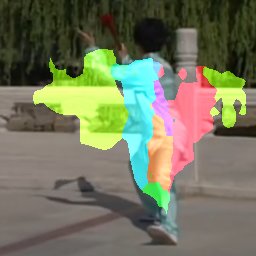} & \includegraphics[width=0.23\columnwidth]{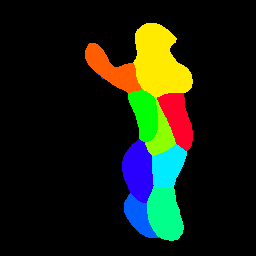}\includegraphics[width=0.23\columnwidth]{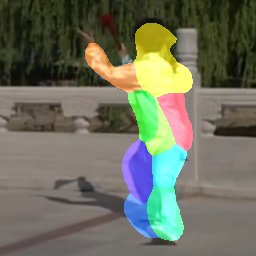}\\
        
        \includegraphics[width=0.23\columnwidth]{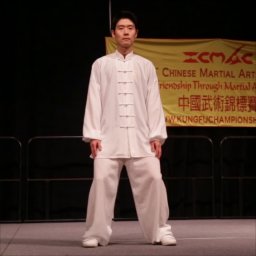} & \includegraphics[width=0.23\columnwidth]{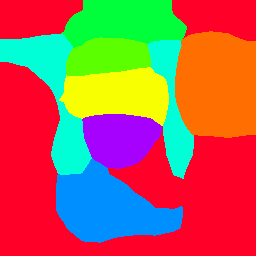}\includegraphics[width=0.23\columnwidth]{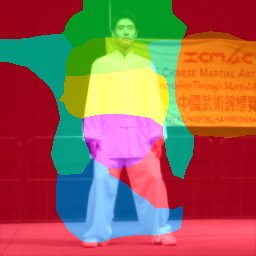} & \includegraphics[width=0.23\columnwidth]{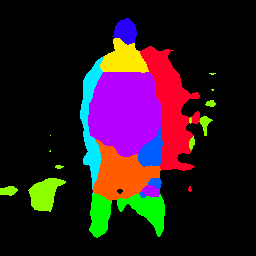}\includegraphics[width=0.23\columnwidth]{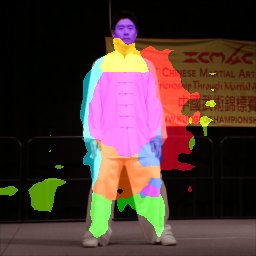} & \includegraphics[width=0.23\columnwidth]{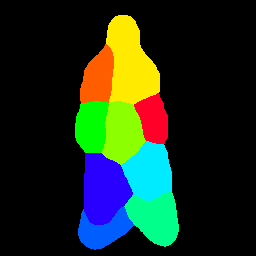}\includegraphics[width=0.23\columnwidth]{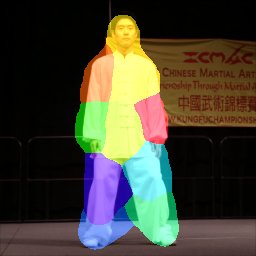}\\
        
        \includegraphics[width=0.23\columnwidth]{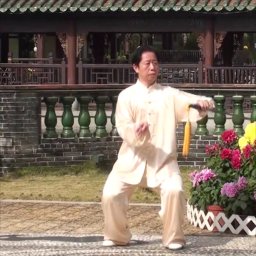} & \includegraphics[width=0.23\columnwidth]{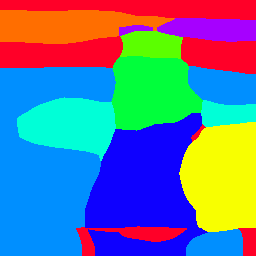}\includegraphics[width=0.23\columnwidth]{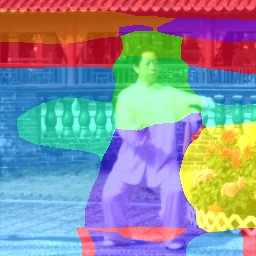} & \includegraphics[width=0.23\columnwidth]{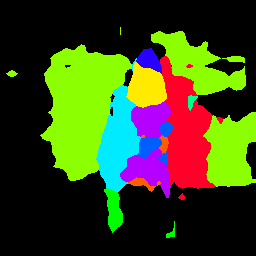}\includegraphics[width=0.23\columnwidth]{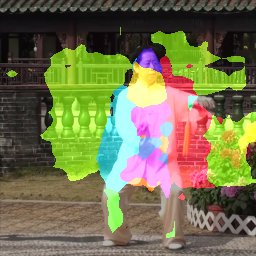} & \includegraphics[width=0.23\columnwidth]{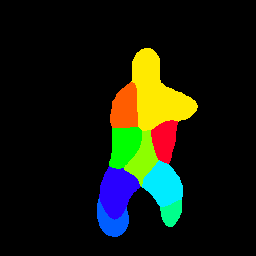}\includegraphics[width=0.23\columnwidth]{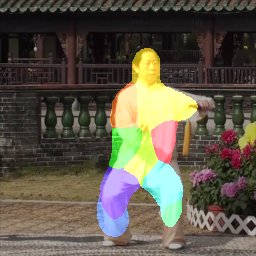}\\
        
        \includegraphics[width=0.23\columnwidth]{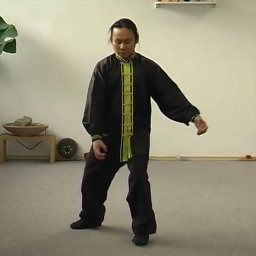} & \includegraphics[width=0.23\columnwidth]{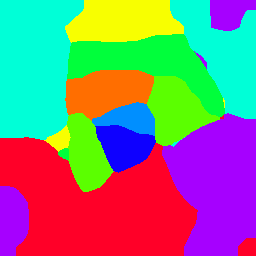}\includegraphics[width=0.23\columnwidth]{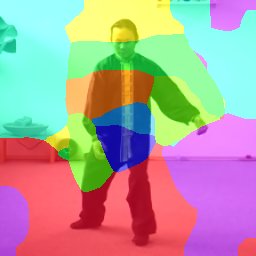} & \includegraphics[width=0.23\columnwidth]{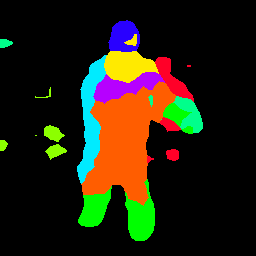}\includegraphics[width=0.23\columnwidth]{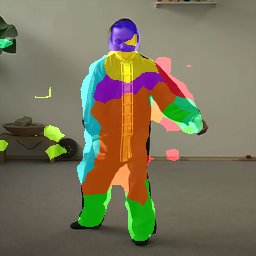} & \includegraphics[width=0.23\columnwidth]{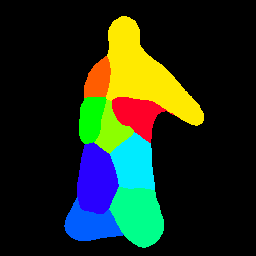}\includegraphics[width=0.23\columnwidth]{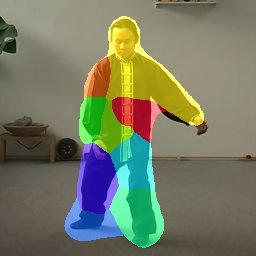}\\
        
        \includegraphics[width=0.23\columnwidth]{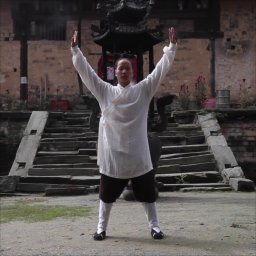} & \includegraphics[width=0.23\columnwidth]{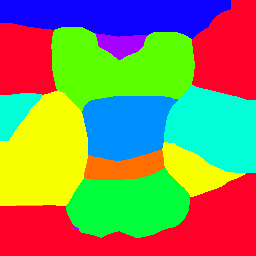}\includegraphics[width=0.23\columnwidth]{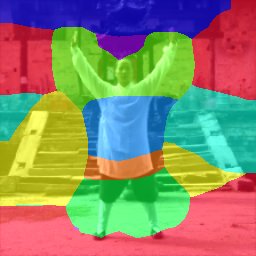} & \includegraphics[width=0.23\columnwidth]{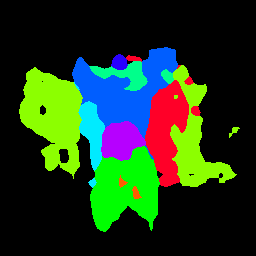}\includegraphics[width=0.23\columnwidth]{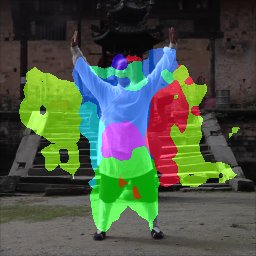} & \includegraphics[width=0.23\columnwidth]{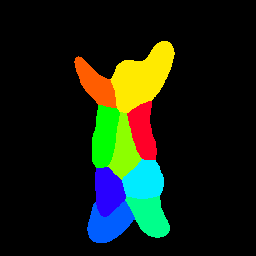}\includegraphics[width=0.23\columnwidth]{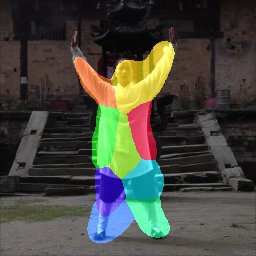}\\
        
        \includegraphics[width=0.23\columnwidth]{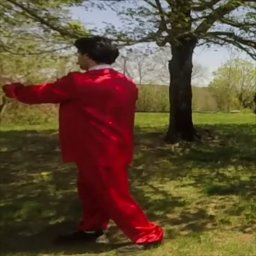} & \includegraphics[width=0.23\columnwidth]{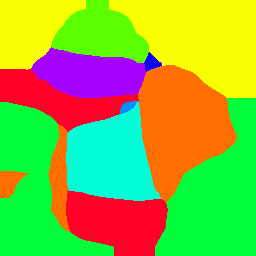}\includegraphics[width=0.23\columnwidth]{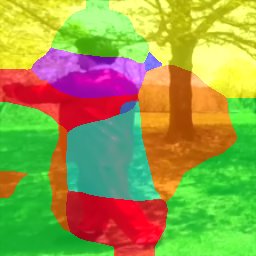} & \includegraphics[width=0.23\columnwidth]{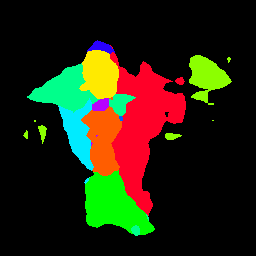}\includegraphics[width=0.23\columnwidth]{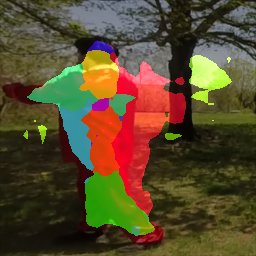} & \includegraphics[width=0.23\columnwidth]{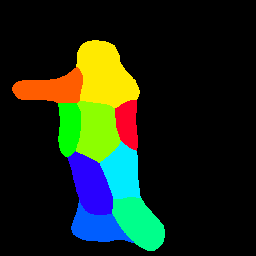}\includegraphics[width=0.23\columnwidth]{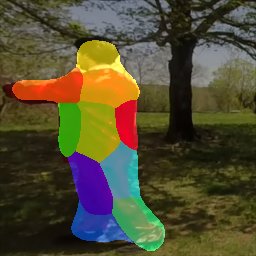}\\
    \end{tabular}
    }
    \captionof{figure}{\textbf{Visual comparison of our method and the state-of-the-art baselines on \emph{Tai-Chi-HD} dataset}. First column is an input. Next columns depict DFF~\cite{collins2018}, SCOPS~\cite{Hung_2019_CVPR} and our method. For every method segmentation mask and image with overlayed segmentation is shown.}
    \label{fig:taichi-sota}
\end{table*}{}

\begin{table*}[h!]
    \centering
    \def\arraystretch{0.9}
    \resizebox{\linewidth}{!}{
    \begin{tabular}{cccc}
        Input & DFF~\cite{collins2018} & SCOPS~\cite{Hung_2019_CVPR} & Ours\\\\
        \includegraphics[width=0.23\columnwidth]{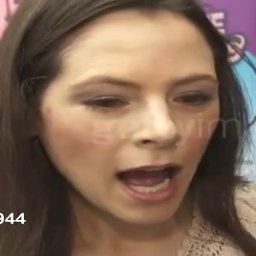} & \includegraphics[width=0.23\columnwidth]{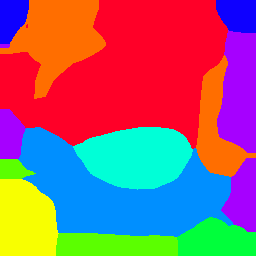}\includegraphics[width=0.23\columnwidth]{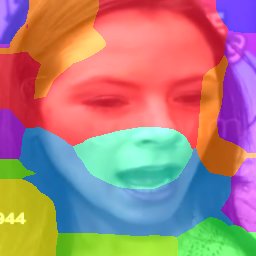} & \includegraphics[width=0.23\columnwidth]{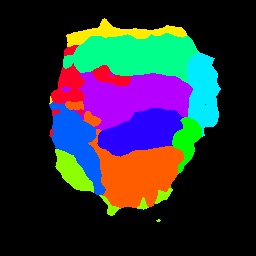}\includegraphics[width=0.23\columnwidth]{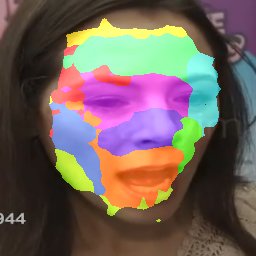} & \includegraphics[width=0.23\columnwidth]{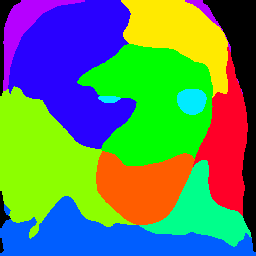}\includegraphics[width=0.23\columnwidth]{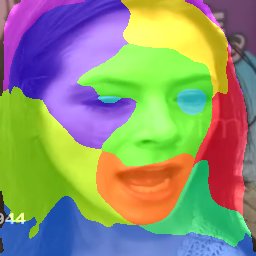}\\
        
        \includegraphics[width=0.23\columnwidth]{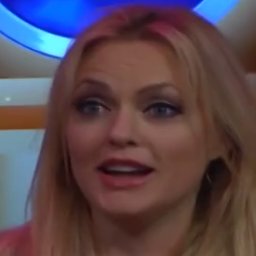} & \includegraphics[width=0.23\columnwidth]{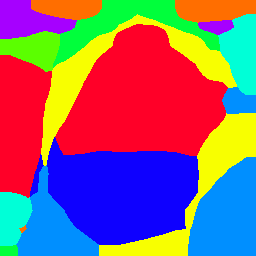}\includegraphics[width=0.23\columnwidth]{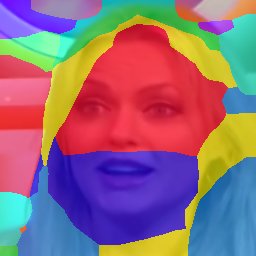} & \includegraphics[width=0.23\columnwidth]{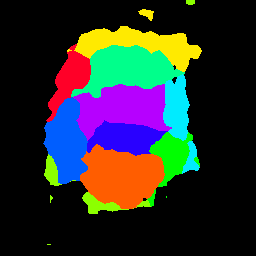}\includegraphics[width=0.23\columnwidth]{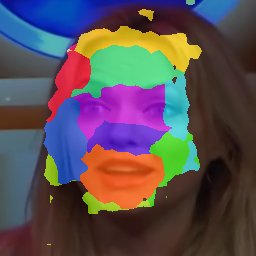} & \includegraphics[width=0.23\columnwidth]{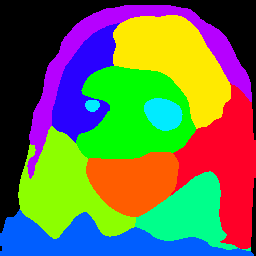}\includegraphics[width=0.23\columnwidth]{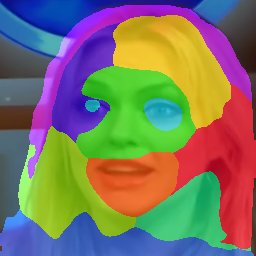}\\
        
        \includegraphics[width=0.23\columnwidth]{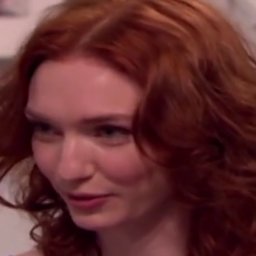} & \includegraphics[width=0.23\columnwidth]{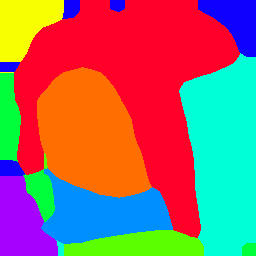}\includegraphics[width=0.23\columnwidth]{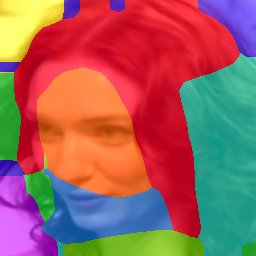} & \includegraphics[width=0.23\columnwidth]{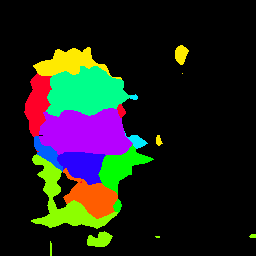}\includegraphics[width=0.23\columnwidth]{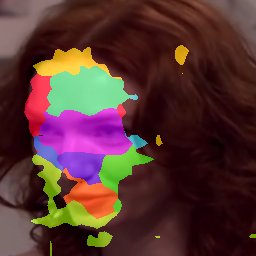} & \includegraphics[width=0.23\columnwidth]{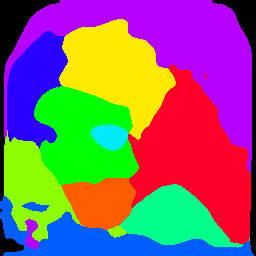}\includegraphics[width=0.23\columnwidth]{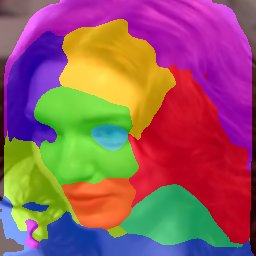}\\
        
        \includegraphics[width=0.23\columnwidth]{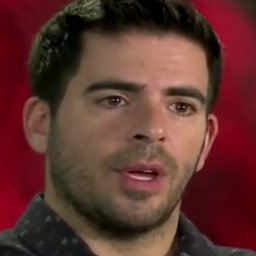} & \includegraphics[width=0.23\columnwidth]{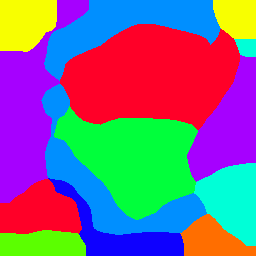}\includegraphics[width=0.23\columnwidth]{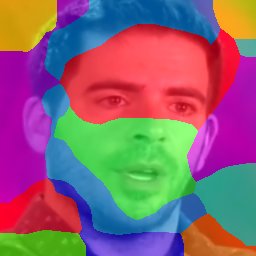} & \includegraphics[width=0.23\columnwidth]{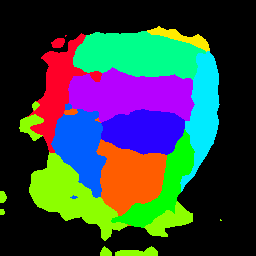}\includegraphics[width=0.23\columnwidth]{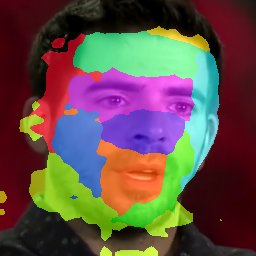} & \includegraphics[width=0.23\columnwidth]{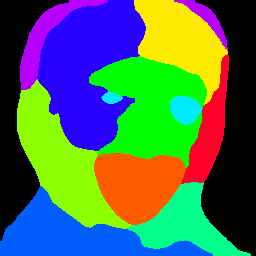}\includegraphics[width=0.23\columnwidth]{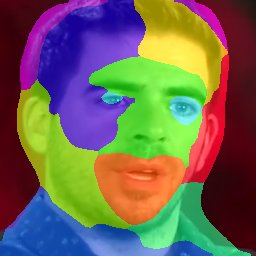}\\
        
        \includegraphics[width=0.23\columnwidth]{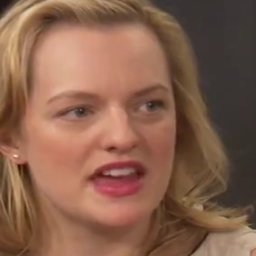} & \includegraphics[width=0.23\columnwidth]{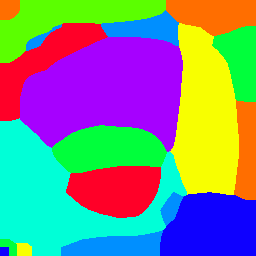}\includegraphics[width=0.23\columnwidth]{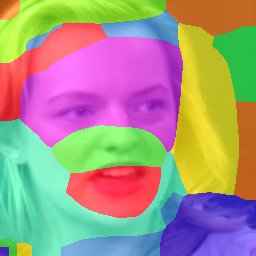} & \includegraphics[width=0.23\columnwidth]{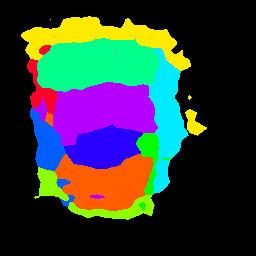}\includegraphics[width=0.23\columnwidth]{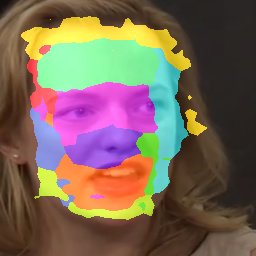} & \includegraphics[width=0.23\columnwidth]{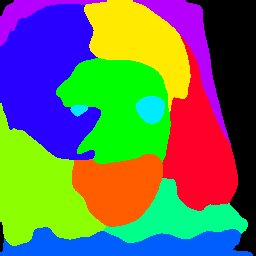}\includegraphics[width=0.23\columnwidth]{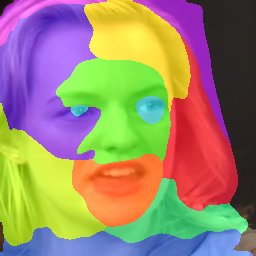}\\
        
        \includegraphics[width=0.23\columnwidth]{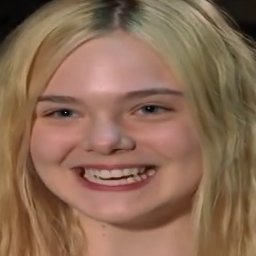} & \includegraphics[width=0.23\columnwidth]{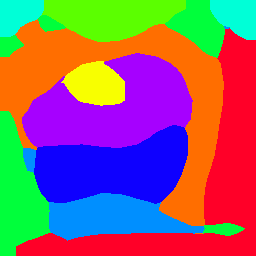}\includegraphics[width=0.23\columnwidth]{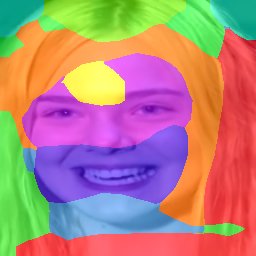} & \includegraphics[width=0.23\columnwidth]{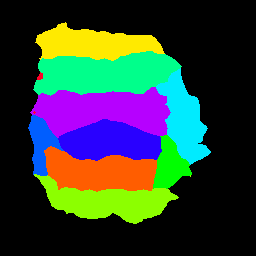}\includegraphics[width=0.23\columnwidth]{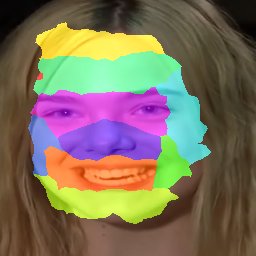} & \includegraphics[width=0.23\columnwidth]{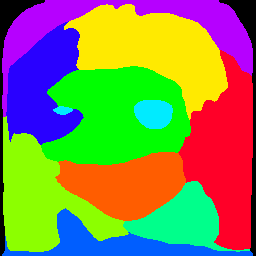}\includegraphics[width=0.23\columnwidth]{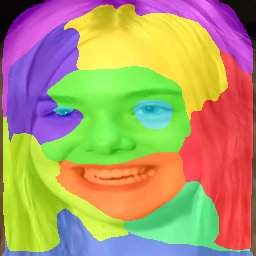}\\
        
        \includegraphics[width=0.23\columnwidth]{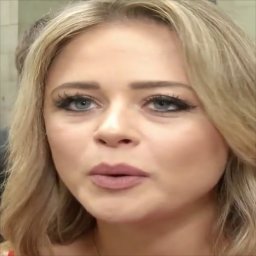} & \includegraphics[width=0.23\columnwidth]{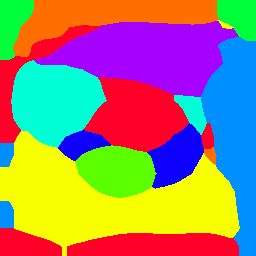}\includegraphics[width=0.23\columnwidth]{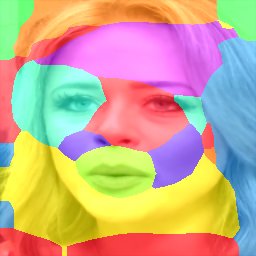} & \includegraphics[width=0.23\columnwidth]{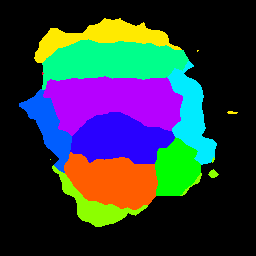}\includegraphics[width=0.23\columnwidth]{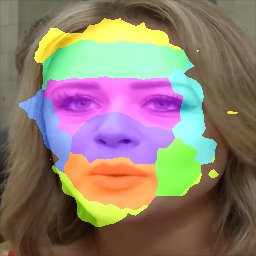} & \includegraphics[width=0.23\columnwidth]{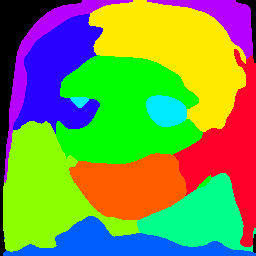}\includegraphics[width=0.23\columnwidth]{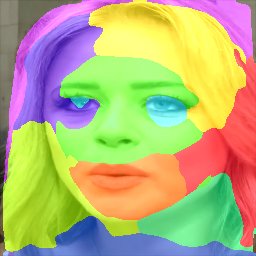}\\
        
        \includegraphics[width=0.23\columnwidth]{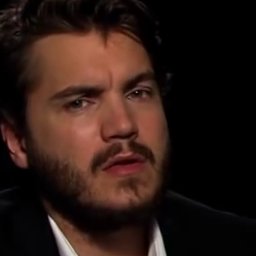} & \includegraphics[width=0.23\columnwidth]{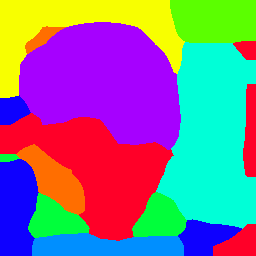}\includegraphics[width=0.23\columnwidth]{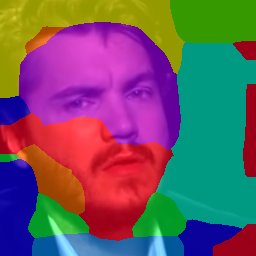} & \includegraphics[width=0.23\columnwidth]{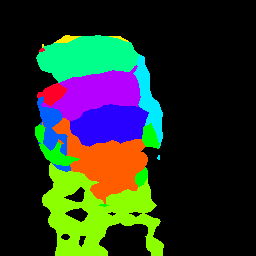}\includegraphics[width=0.23\columnwidth]{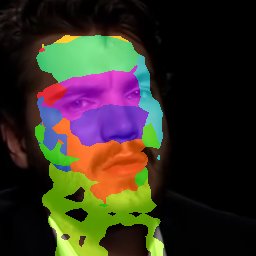} & \includegraphics[width=0.23\columnwidth]{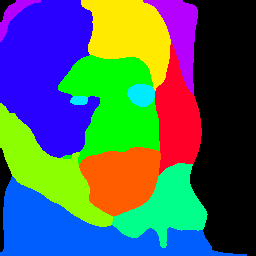}\includegraphics[width=0.23\columnwidth]{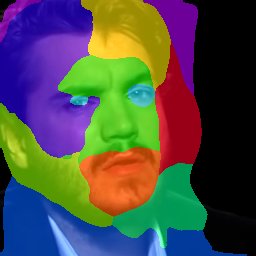}\\
    \end{tabular}
    }
    \captionof{figure}{\textbf{Visual comparison of our method and the state-of-the-art baselines on \emph{VoxCeleb} dataset}. First column is an input. Next columns depict DFF~\cite{collins2018}, SCOPS~\cite{Hung_2019_CVPR} and our method. For every method segmentation mask and image with overlayed segmentation is shown.}
    \label{fig:vox-sota}
\end{table*}{}

\section{Additional part-swap results}
\label{sec:part-swap}
In this section, we provide additional results for part-swap application, that was introduced in Sec.~4.3 of the main paper. In Figs.~\ref{fig:visual_actors},~\ref{fig:visual_garry} we demonstrate an additional visual results using the model trained on \emph{VoxCeleb} dataset. In Fig.~\ref{fig:visual_actors} we swap the hair (top rows) and the top part of the face (bottom rows), using $K=5$ segment model. On the other hand in Fig.~\ref{fig:visual_garry}, we show hair swaps with $K=10$ segments model. Interestingly, we observe that for persons with short hair the best results are archived when we use 4 segments from the source image, while for persons with long hair the best result archived when we use 5 segments. Finally in Fig.~\ref{fig:visual_taichi_swaps} we demonstrate results of our method on \emph{Tai-Chi-HD} dataset. In top rows we change the upper part of the body, while in bottom 2 we alter an appearance of the legs. Overall, we observe that without fine-tuning or special training our model can perform a range of different part-swap operations.

\begin{table*}[t]
    \centering
    \def\arraystretch{0.5}
    \resizebox{\linewidth}{!}{
    \begin{tabular}{c}
         \includegraphics[width=0.15\columnwidth]{figures/vox/swaps/white}\includegraphics[width=0.15\columnwidth]{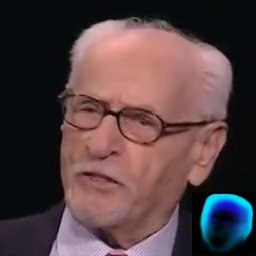}\includegraphics[width=0.15\columnwidth]{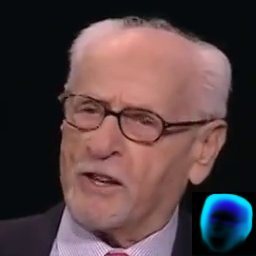}\includegraphics[width=0.15\columnwidth]{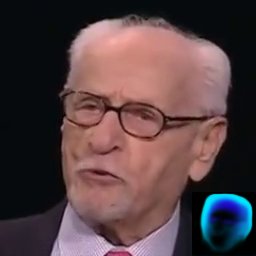}\includegraphics[width=0.15\columnwidth]{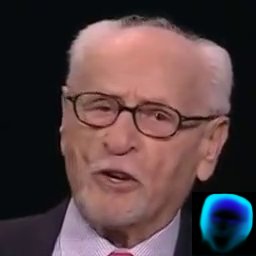}\includegraphics[width=0.15\columnwidth]{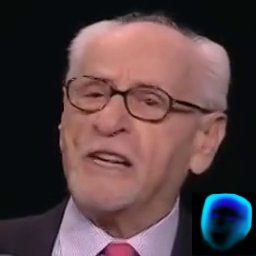}\includegraphics[width=0.15\columnwidth]{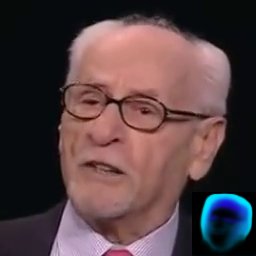}\includegraphics[width=0.15\columnwidth]{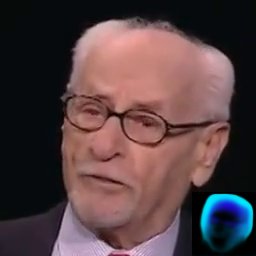} \\
         
         \includegraphics[width=0.15\columnwidth]{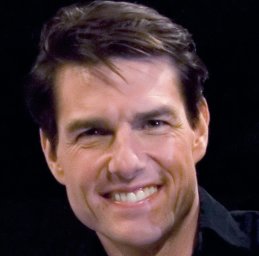}\includegraphics[width=0.15\columnwidth]{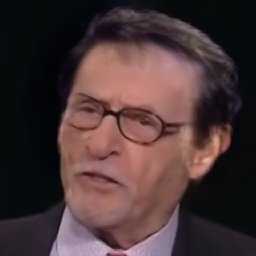}\includegraphics[width=0.15\columnwidth]{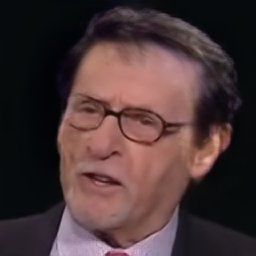}\includegraphics[width=0.15\columnwidth]{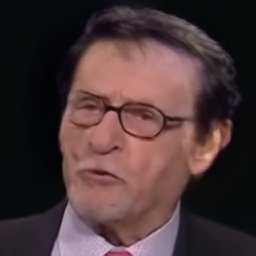}\includegraphics[width=0.15\columnwidth]{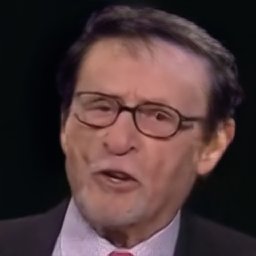}\includegraphics[width=0.15\columnwidth]{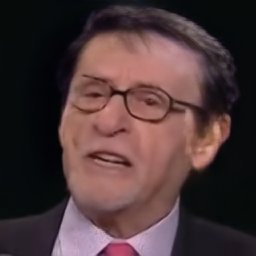}\includegraphics[width=0.15\columnwidth]{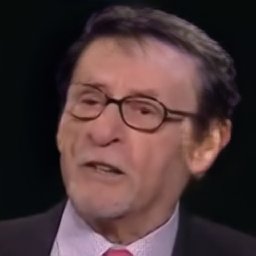}\includegraphics[width=0.15\columnwidth]{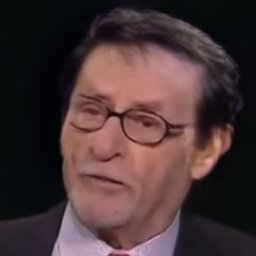} \\
         
         \includegraphics[width=0.15\columnwidth]{figures/vox/swaps/white}\includegraphics[width=0.15\columnwidth]{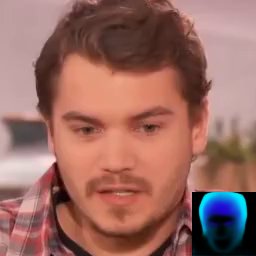}\includegraphics[width=0.15\columnwidth]{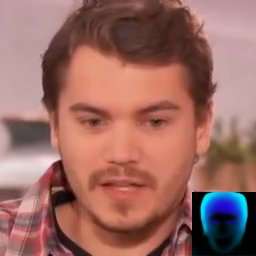}\includegraphics[width=0.15\columnwidth]{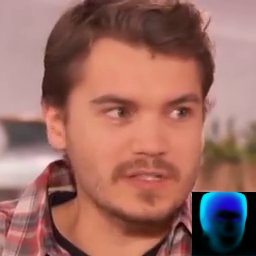}\includegraphics[width=0.15\columnwidth]{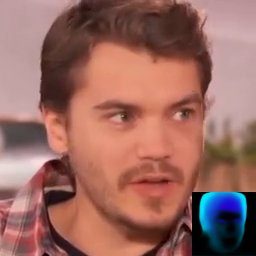}\includegraphics[width=0.15\columnwidth]{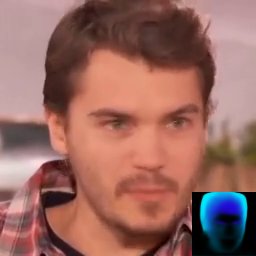}\includegraphics[width=0.15\columnwidth]{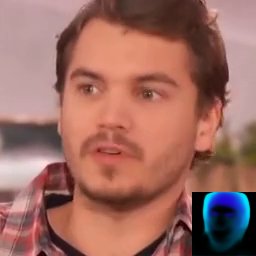}\includegraphics[width=0.15\columnwidth]{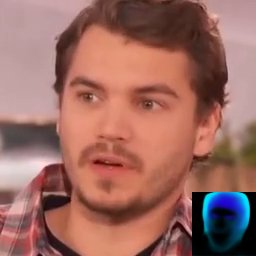} \\
         
         \includegraphics[width=0.15\columnwidth]{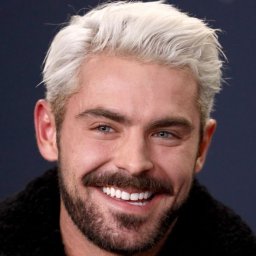}\includegraphics[width=0.15\columnwidth]{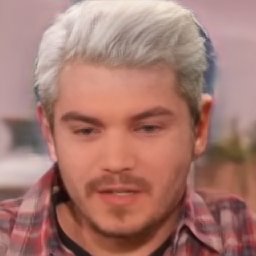}\includegraphics[width=0.15\columnwidth]{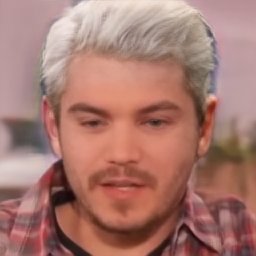}\includegraphics[width=0.15\columnwidth]{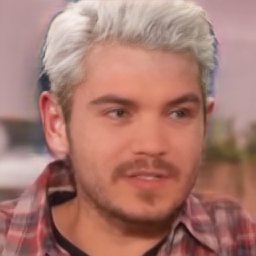}\includegraphics[width=0.15\columnwidth]{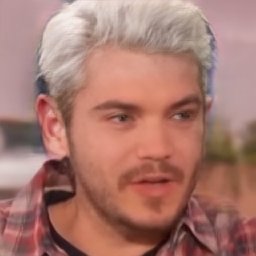}\includegraphics[width=0.15\columnwidth]{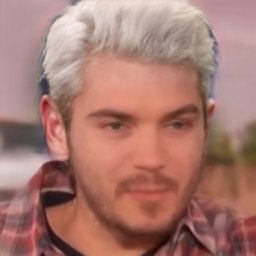}\includegraphics[width=0.15\columnwidth]{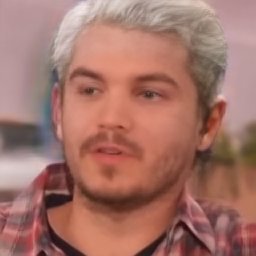}\includegraphics[width=0.15\columnwidth]{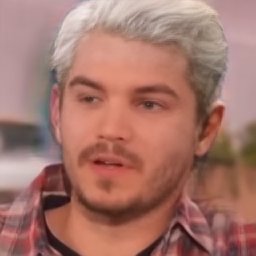} \\
         
         \includegraphics[width=0.15\columnwidth]{figures/vox/swaps/white}\includegraphics[width=0.15\columnwidth]{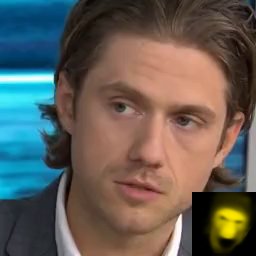}\includegraphics[width=0.15\columnwidth]{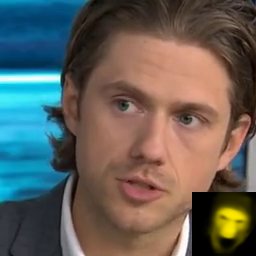}\includegraphics[width=0.15\columnwidth]{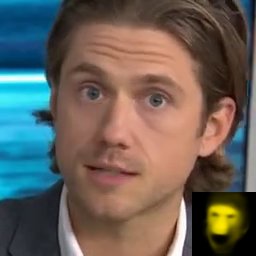}\includegraphics[width=0.15\columnwidth]{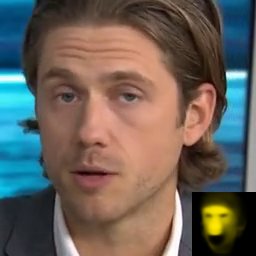}\includegraphics[width=0.15\columnwidth]{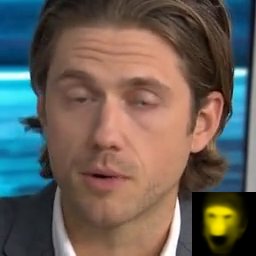}\includegraphics[width=0.15\columnwidth]{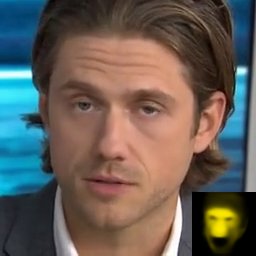}\includegraphics[width=0.15\columnwidth]{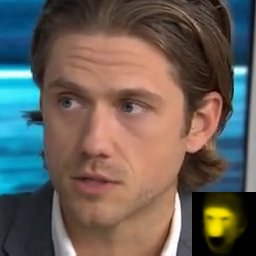} \\
         
         \includegraphics[width=0.15\columnwidth]{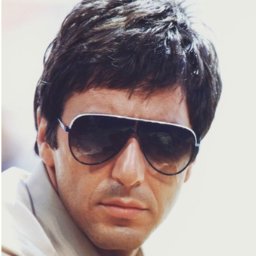}\includegraphics[width=0.15\columnwidth]{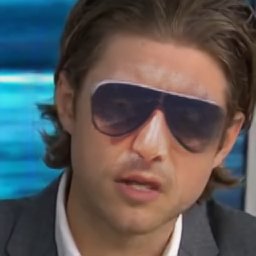}\includegraphics[width=0.15\columnwidth]{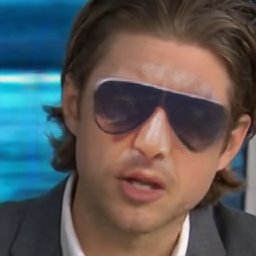}\includegraphics[width=0.15\columnwidth]{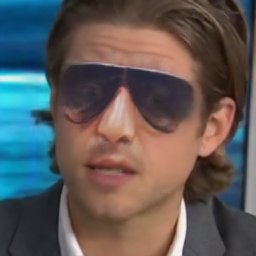}\includegraphics[width=0.15\columnwidth]{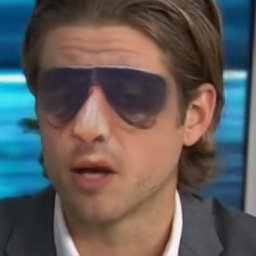}\includegraphics[width=0.15\columnwidth]{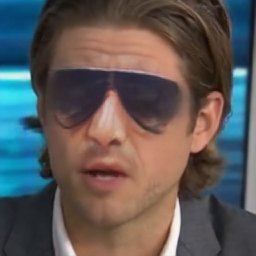}\includegraphics[width=0.15\columnwidth]{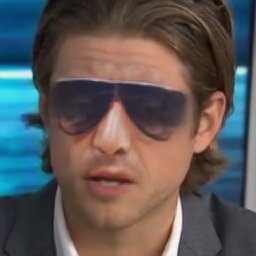}\includegraphics[width=0.15\columnwidth]{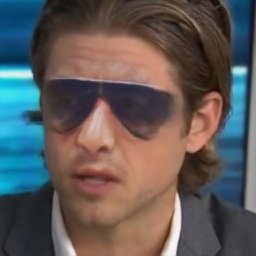} \\
         
         \includegraphics[width=0.15\columnwidth]{figures/vox/swaps/white}\includegraphics[width=0.15\columnwidth]{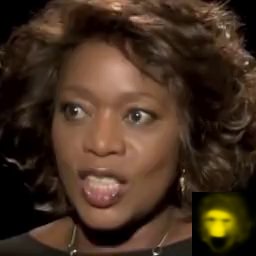}\includegraphics[width=0.15\columnwidth]{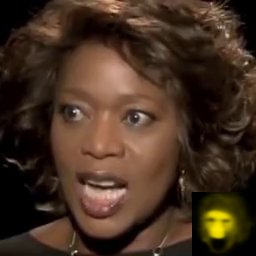}\includegraphics[width=0.15\columnwidth]{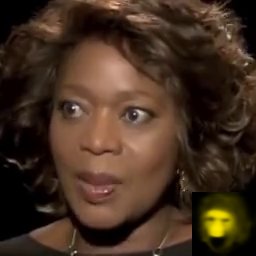}\includegraphics[width=0.15\columnwidth]{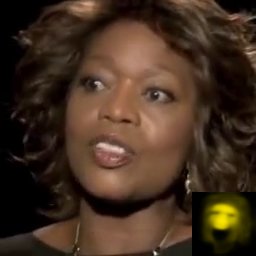}\includegraphics[width=0.15\columnwidth]{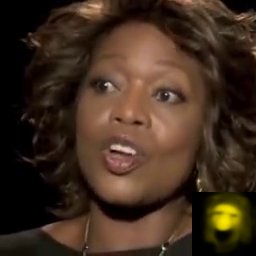}\includegraphics[width=0.15\columnwidth]{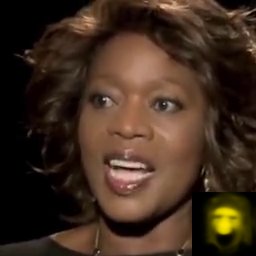}\includegraphics[width=0.15\columnwidth]{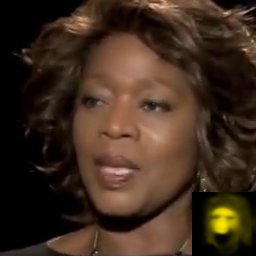} \\
         
         \includegraphics[width=0.15\columnwidth]{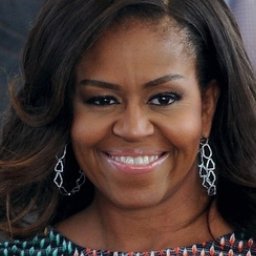}\includegraphics[width=0.15\columnwidth]{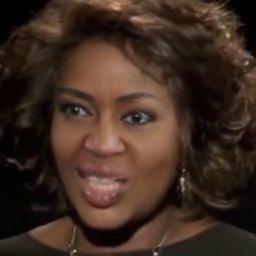}\includegraphics[width=0.15\columnwidth]{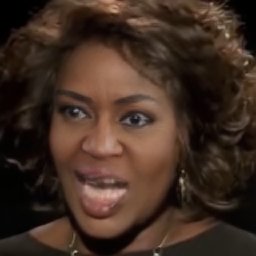}\includegraphics[width=0.15\columnwidth]{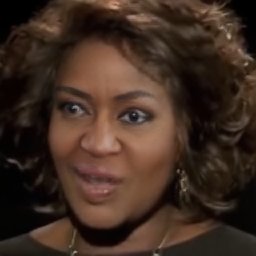}\includegraphics[width=0.15\columnwidth]{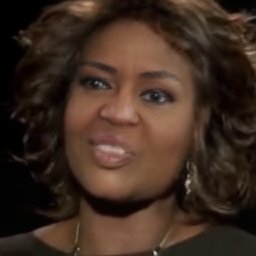}\includegraphics[width=0.15\columnwidth]{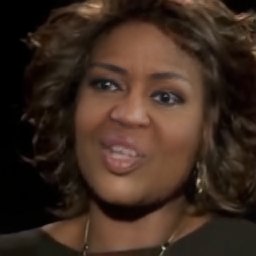}\includegraphics[width=0.15\columnwidth]{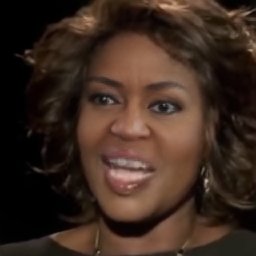}\includegraphics[width=0.15\columnwidth]{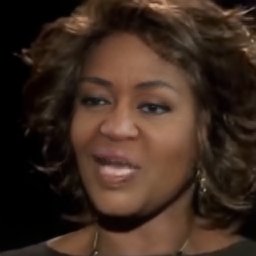} \\
         
         \includegraphics[width=0.15\columnwidth]{figures/vox/swaps/white}\includegraphics[width=0.15\columnwidth]{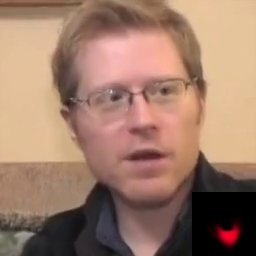}\includegraphics[width=0.15\columnwidth]{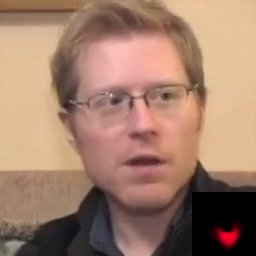}\includegraphics[width=0.15\columnwidth]{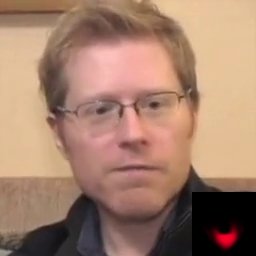}\includegraphics[width=0.15\columnwidth]{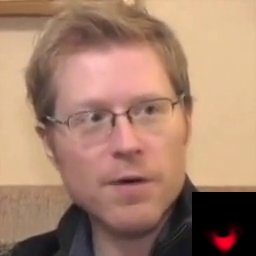}\includegraphics[width=0.15\columnwidth]{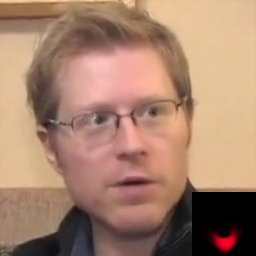}\includegraphics[width=0.15\columnwidth]{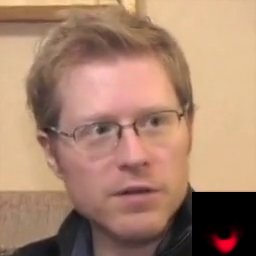}\includegraphics[width=0.15\columnwidth]{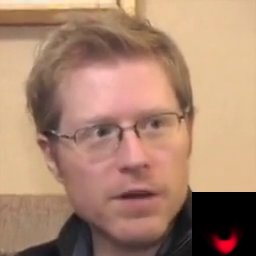} \\
         
         \includegraphics[width=0.15\columnwidth]{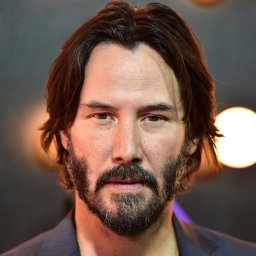}\includegraphics[width=0.15\columnwidth]{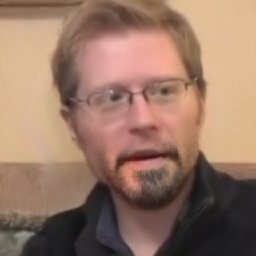}\includegraphics[width=0.15\columnwidth]{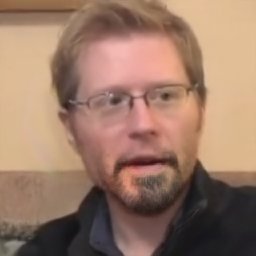}\includegraphics[width=0.15\columnwidth]{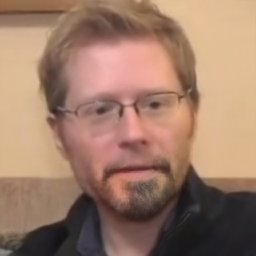}\includegraphics[width=0.15\columnwidth]{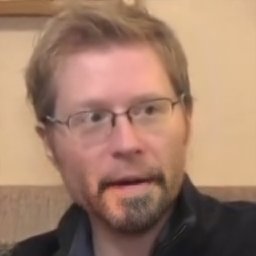}\includegraphics[width=0.15\columnwidth]{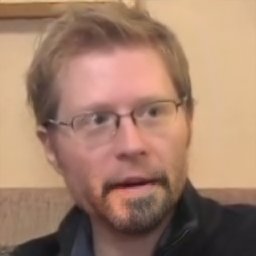}\includegraphics[width=0.15\columnwidth]{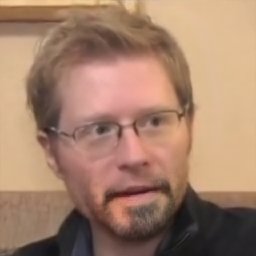}\includegraphics[width=0.15\columnwidth]{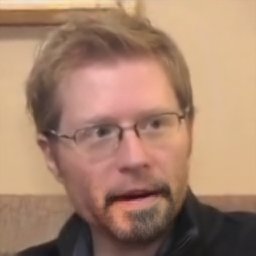} \\
    \end{tabular}}
    
    \captionof{figure}{\textbf{Visual results of video-editing for \textit{VoxCeleb}}. In the odd rows the target frames from video sequences are depicted, alongside the masks of interest (in the right bottom corners) intended to be swapped. In the even rows the source image is shown, followed by the generated frames containing the indicated parts swapped from the source. Results achieved with model for \textit{K}=5 are depicted. Please note that the source images are downloaded from Google Images. Best viewed with digital zoom.}
    \label{fig:visual_actors}

\end{table*}{}

\begin{table*}[t]
    \centering
    \def\arraystretch{0.5}
    \resizebox{\linewidth}{!}{
    \begin{tabular}{c}
         \includegraphics[width=0.15\columnwidth]{figures/vox/swaps/white}\includegraphics[width=0.15\columnwidth]{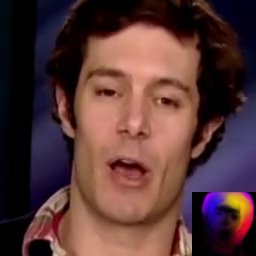}\includegraphics[width=0.15\columnwidth]{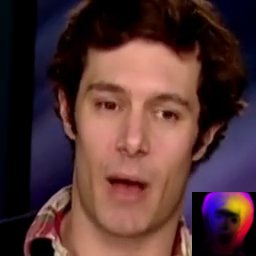}\includegraphics[width=0.15\columnwidth]{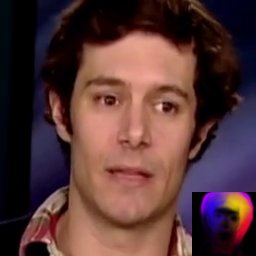}\includegraphics[width=0.15\columnwidth]{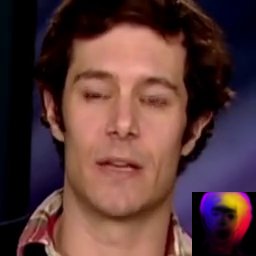}\includegraphics[width=0.15\columnwidth]{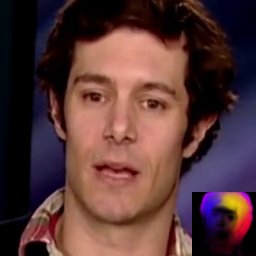}\includegraphics[width=0.15\columnwidth]{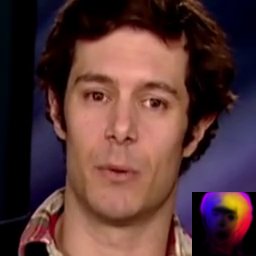}\includegraphics[width=0.15\columnwidth]{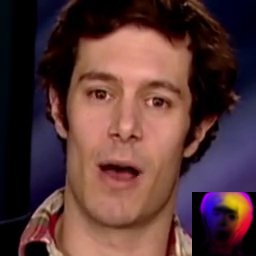} \\
         
         \includegraphics[width=0.15\columnwidth]{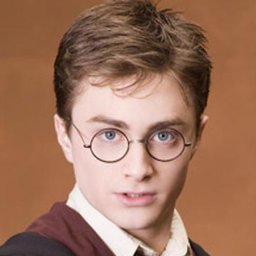}\includegraphics[width=0.15\columnwidth]{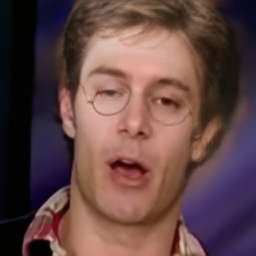}\includegraphics[width=0.15\columnwidth]{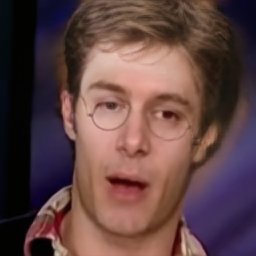}\includegraphics[width=0.15\columnwidth]{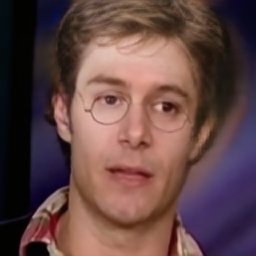}\includegraphics[width=0.15\columnwidth]{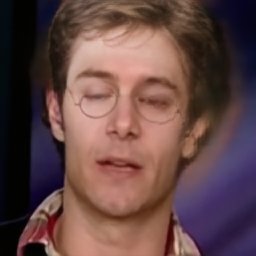}\includegraphics[width=0.15\columnwidth]{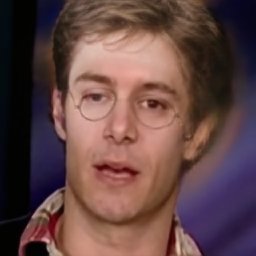}\includegraphics[width=0.15\columnwidth]{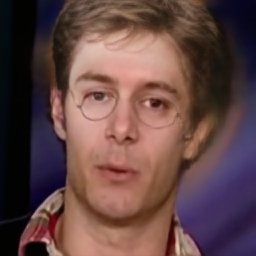}\includegraphics[width=0.15\columnwidth]{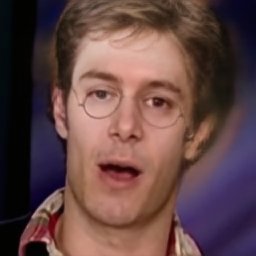} \\
         
         \includegraphics[width=0.15\columnwidth]{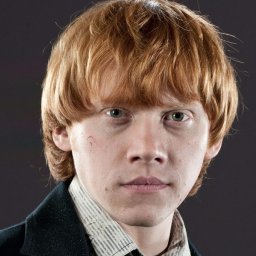}\includegraphics[width=0.15\columnwidth]{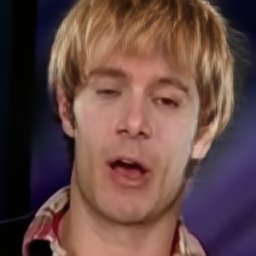}\includegraphics[width=0.15\columnwidth]{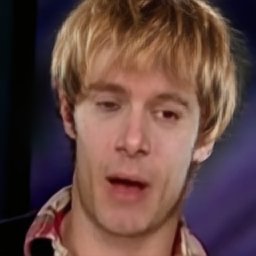}\includegraphics[width=0.15\columnwidth]{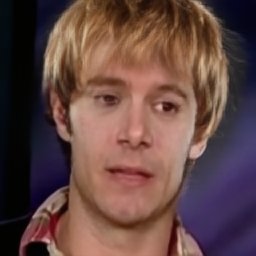}\includegraphics[width=0.15\columnwidth]{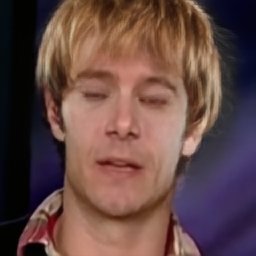}\includegraphics[width=0.15\columnwidth]{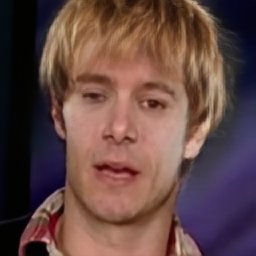}\includegraphics[width=0.15\columnwidth]{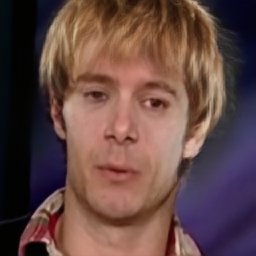}\includegraphics[width=0.15\columnwidth]{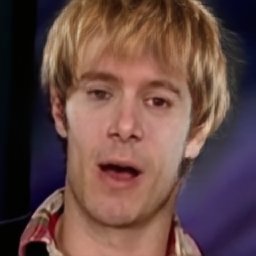} \\
         
         \includegraphics[width=0.15\columnwidth]{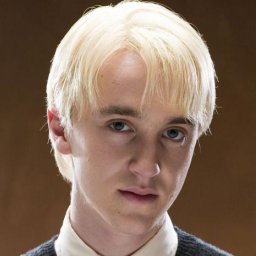}\includegraphics[width=0.15\columnwidth]{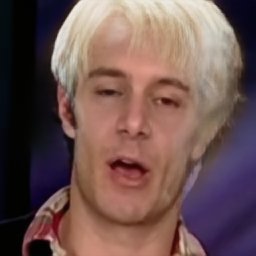}\includegraphics[width=0.15\columnwidth]{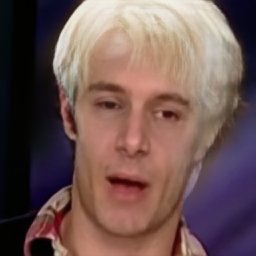}\includegraphics[width=0.15\columnwidth]{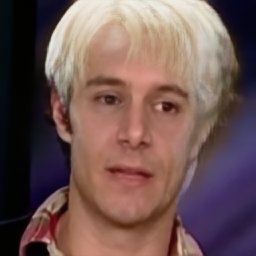}\includegraphics[width=0.15\columnwidth]{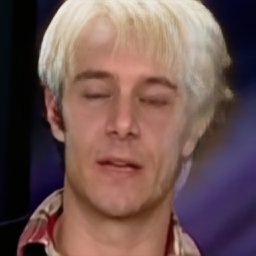}\includegraphics[width=0.15\columnwidth]{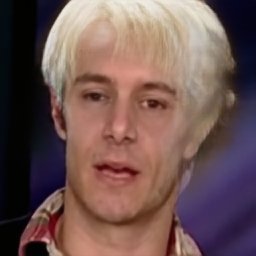}\includegraphics[width=0.15\columnwidth]{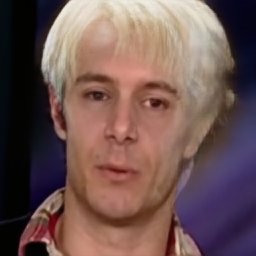}\includegraphics[width=0.15\columnwidth]{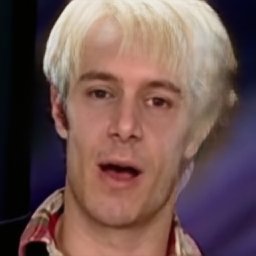} \\
         
         \includegraphics[width=0.15\columnwidth]{figures/vox/swaps/white}\includegraphics[width=0.15\columnwidth]{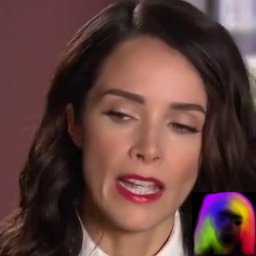}\includegraphics[width=0.15\columnwidth]{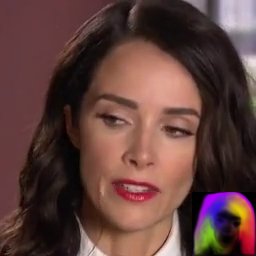}\includegraphics[width=0.15\columnwidth]{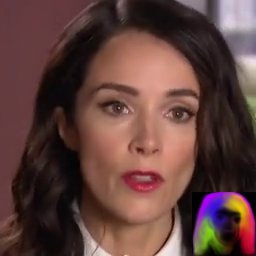}\includegraphics[width=0.15\columnwidth]{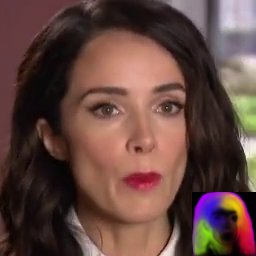}\includegraphics[width=0.15\columnwidth]{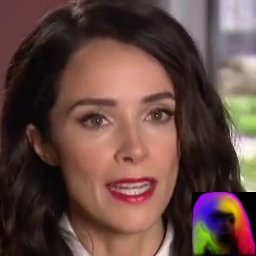}\includegraphics[width=0.15\columnwidth]{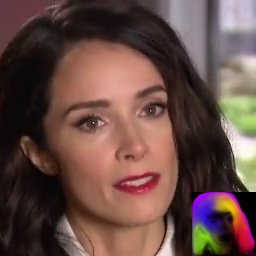}\includegraphics[width=0.15\columnwidth]{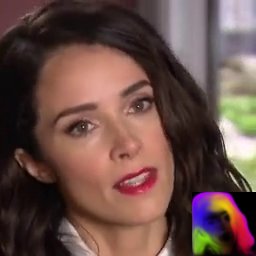} \\
         
         \includegraphics[width=0.15\columnwidth]{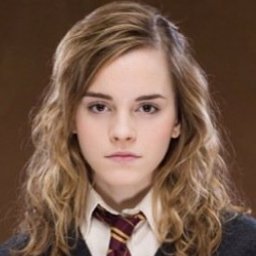}\includegraphics[width=0.15\columnwidth]{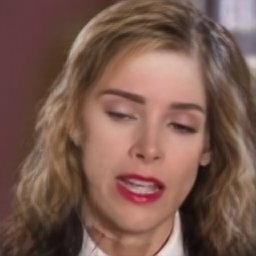}\includegraphics[width=0.15\columnwidth]{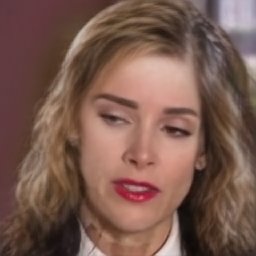}\includegraphics[width=0.15\columnwidth]{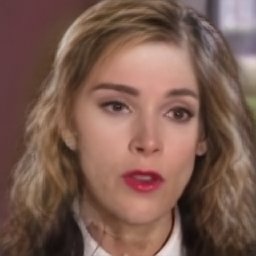}\includegraphics[width=0.15\columnwidth]{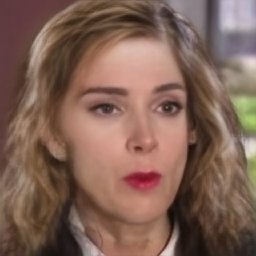}\includegraphics[width=0.15\columnwidth]{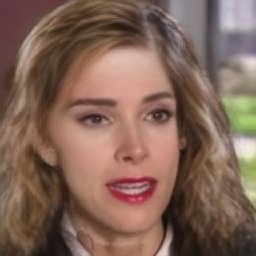}\includegraphics[width=0.15\columnwidth]{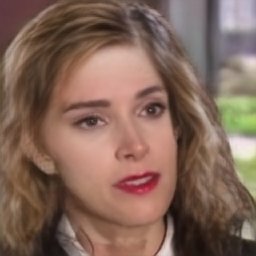}\includegraphics[width=0.15\columnwidth]{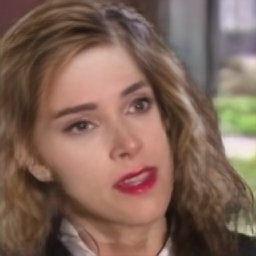} \\
         
         \includegraphics[width=0.15\columnwidth]{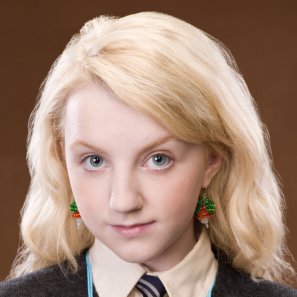}\includegraphics[width=0.15\columnwidth]{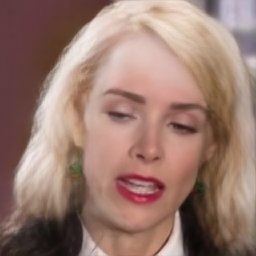}\includegraphics[width=0.15\columnwidth]{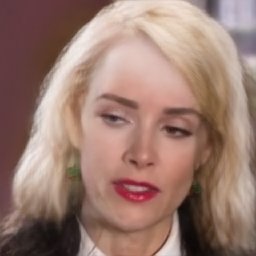}\includegraphics[width=0.15\columnwidth]{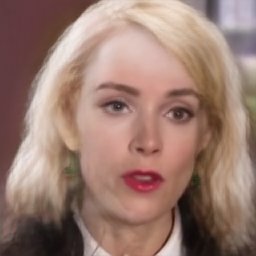}\includegraphics[width=0.15\columnwidth]{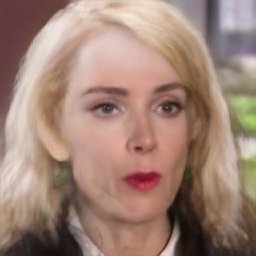}\includegraphics[width=0.15\columnwidth]{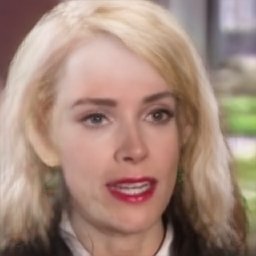}\includegraphics[width=0.15\columnwidth]{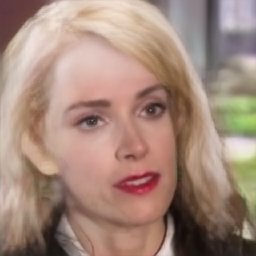}\includegraphics[width=0.15\columnwidth]{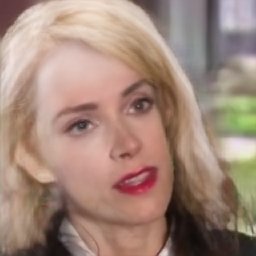} \\
         
         \includegraphics[width=0.15\columnwidth]{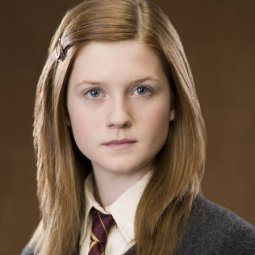}\includegraphics[width=0.15\columnwidth]{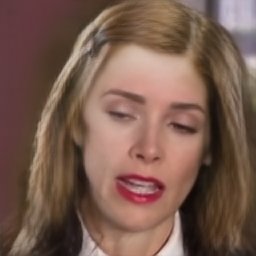}\includegraphics[width=0.15\columnwidth]{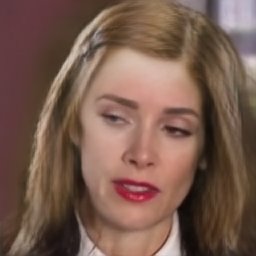}\includegraphics[width=0.15\columnwidth]{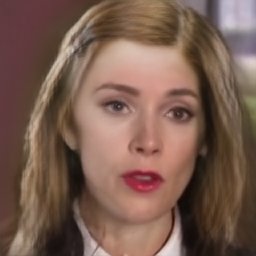}\includegraphics[width=0.15\columnwidth]{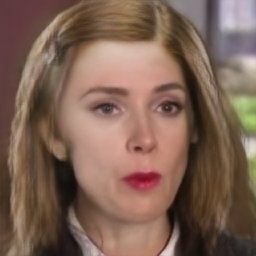}\includegraphics[width=0.15\columnwidth]{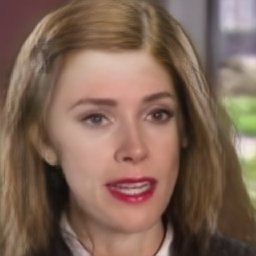}\includegraphics[width=0.15\columnwidth]{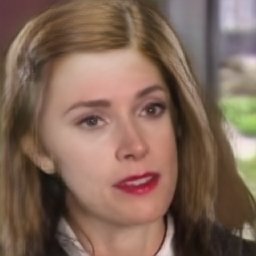}\includegraphics[width=0.15\columnwidth]{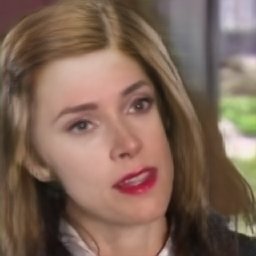} \\
    \end{tabular}}
    
    \captionof{figure}{\textbf{Visual results of video-editing for \textit{VoxCeleb}}. In the first and the fifth rows the target frames from video sequences are depicted, alongside the masks of interest (in the right bottom corners) intended to be swapped. In the rest of the rows the source images are shown, followed by the generated frames containing the indicated parts swapped from the source. Results achieved with model for \textit{K}=10 are depicted. Please note that the source images are downloaded from Google Images. Best viewed with digital zoom.}
    \label{fig:visual_garry}

\end{table*}{}

\begin{table*}[t]
    \centering
    \def\arraystretch{0.5}
    \resizebox{\linewidth}{!}{
    \begin{tabular}{c}
         \includegraphics[width=0.15\columnwidth]{figures/vox/swaps/white}\includegraphics[width=0.15\columnwidth]{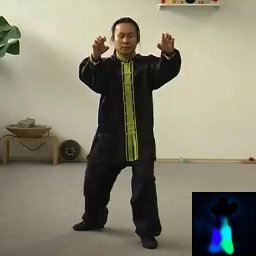}\includegraphics[width=0.15\columnwidth]{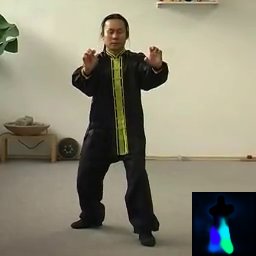}\includegraphics[width=0.15\columnwidth]{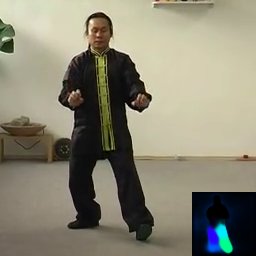}\includegraphics[width=0.15\columnwidth]{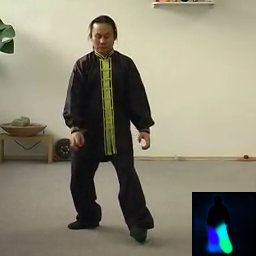}\includegraphics[width=0.15\columnwidth]{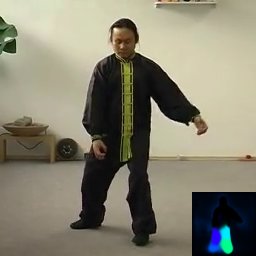}\includegraphics[width=0.15\columnwidth]{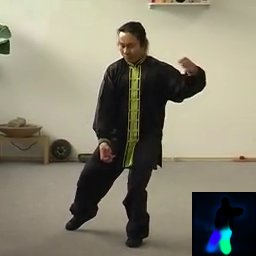}\includegraphics[width=0.15\columnwidth]{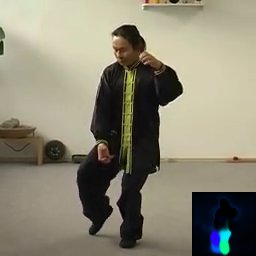} \\
         
         \includegraphics[width=0.15\columnwidth]{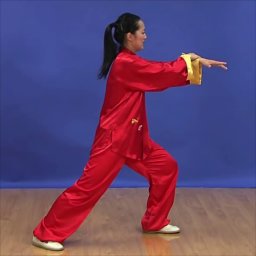}\includegraphics[width=0.15\columnwidth]{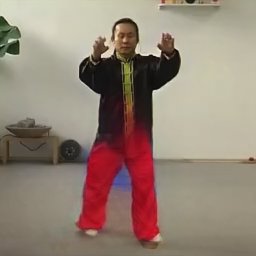}\includegraphics[width=0.15\columnwidth]{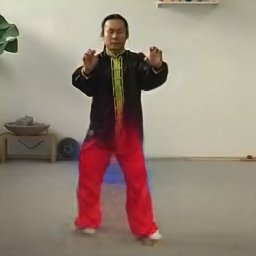}\includegraphics[width=0.15\columnwidth]{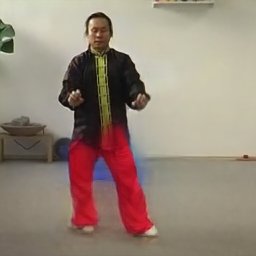}\includegraphics[width=0.15\columnwidth]{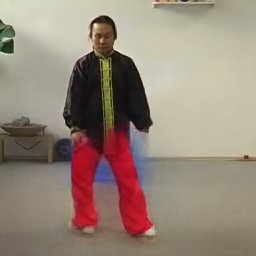}\includegraphics[width=0.15\columnwidth]{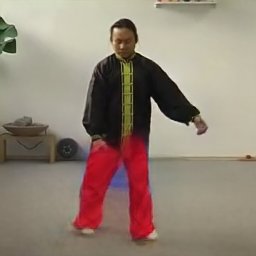}\includegraphics[width=0.15\columnwidth]{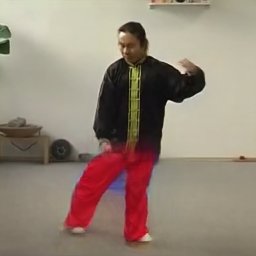}\includegraphics[width=0.15\columnwidth]{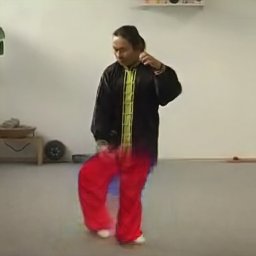} \\
         
         \includegraphics[width=0.15\columnwidth]{figures/vox/swaps/white}\includegraphics[width=0.15\columnwidth]{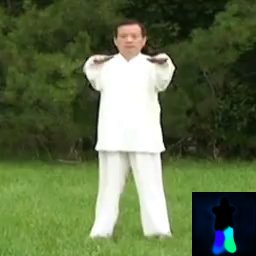}\includegraphics[width=0.15\columnwidth]{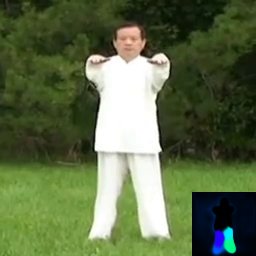}\includegraphics[width=0.15\columnwidth]{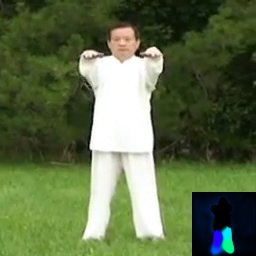}\includegraphics[width=0.15\columnwidth]{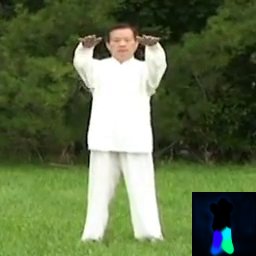}\includegraphics[width=0.15\columnwidth]{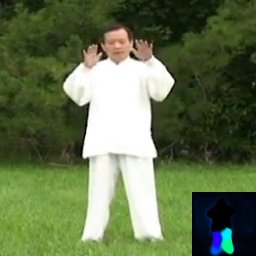}\includegraphics[width=0.15\columnwidth]{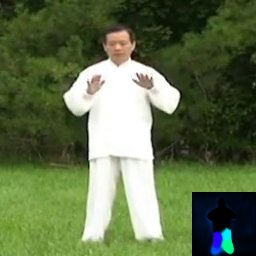}\includegraphics[width=0.15\columnwidth]{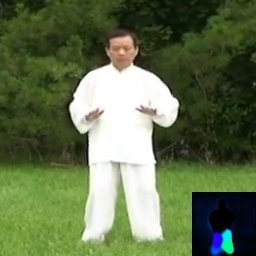} \\
         
         \includegraphics[width=0.15\columnwidth]{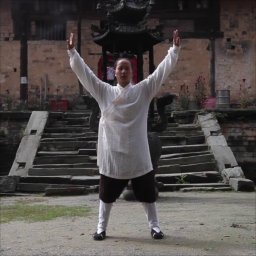}\includegraphics[width=0.15\columnwidth]{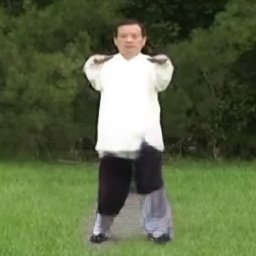}\includegraphics[width=0.15\columnwidth]{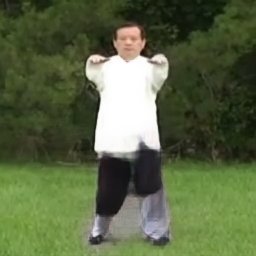}\includegraphics[width=0.15\columnwidth]{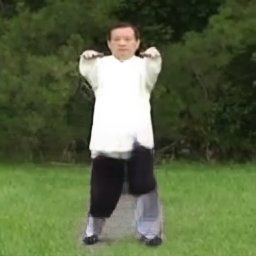}\includegraphics[width=0.15\columnwidth]{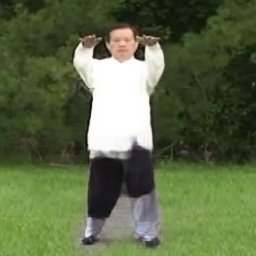}\includegraphics[width=0.15\columnwidth]{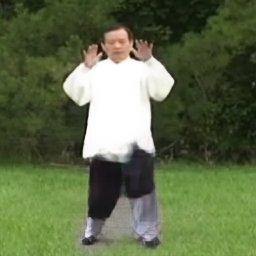}\includegraphics[width=0.15\columnwidth]{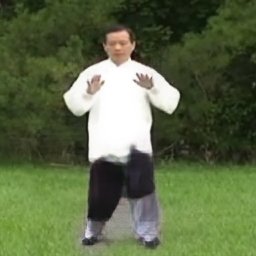}\includegraphics[width=0.15\columnwidth]{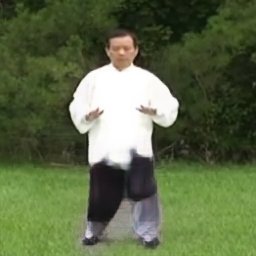} \\

        \includegraphics[width=0.15\columnwidth]{figures/vox/swaps/white}\includegraphics[width=0.15\columnwidth]{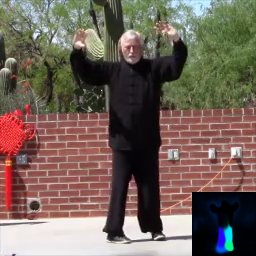}\includegraphics[width=0.15\columnwidth]{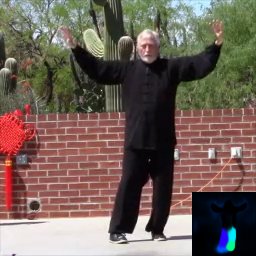}\includegraphics[width=0.15\columnwidth]{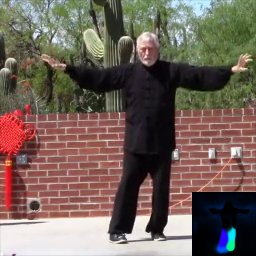}\includegraphics[width=0.15\columnwidth]{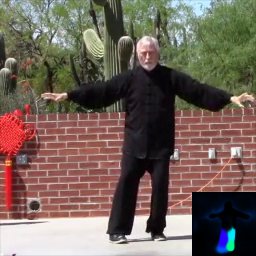}\includegraphics[width=0.15\columnwidth]{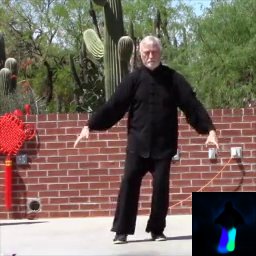}\includegraphics[width=0.15\columnwidth]{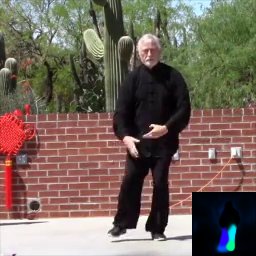}\includegraphics[width=0.15\columnwidth]{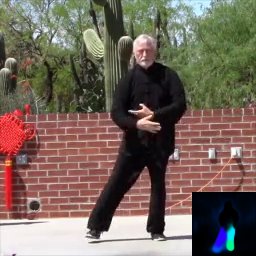} \\
         
         \includegraphics[width=0.15\columnwidth]{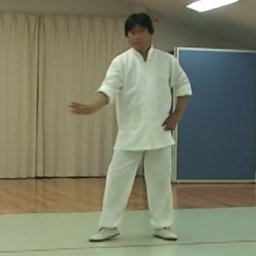}\includegraphics[width=0.15\columnwidth]{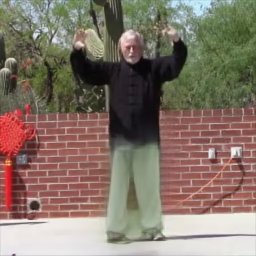}\includegraphics[width=0.15\columnwidth]{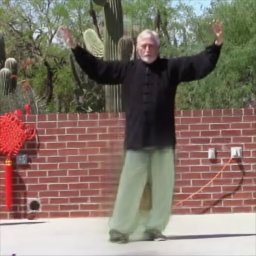}\includegraphics[width=0.15\columnwidth]{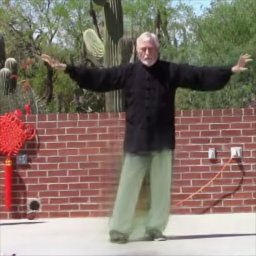}\includegraphics[width=0.15\columnwidth]{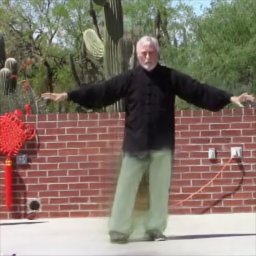}\includegraphics[width=0.15\columnwidth]{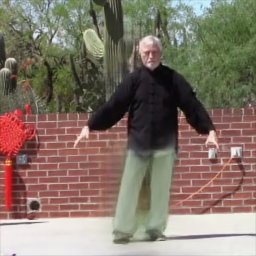}\includegraphics[width=0.15\columnwidth]{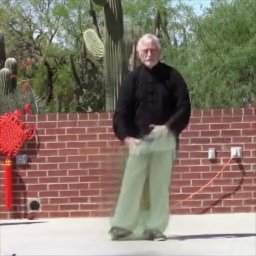}\includegraphics[width=0.15\columnwidth]{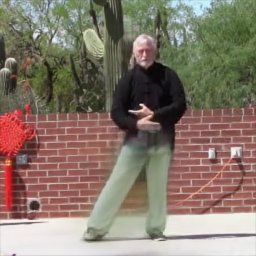} \\
         
         \includegraphics[width=0.15\columnwidth]{figures/vox/swaps/white}\includegraphics[width=0.15\columnwidth]{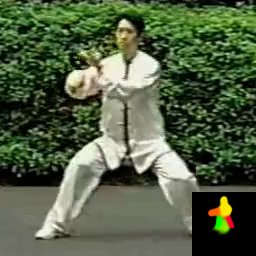}\includegraphics[width=0.15\columnwidth]{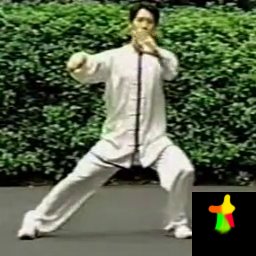}\includegraphics[width=0.15\columnwidth]{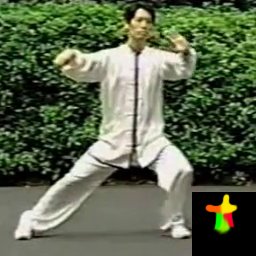}\includegraphics[width=0.15\columnwidth]{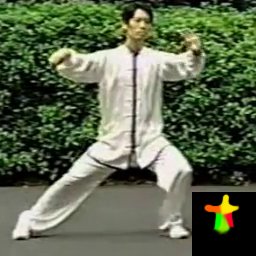}\includegraphics[width=0.15\columnwidth]{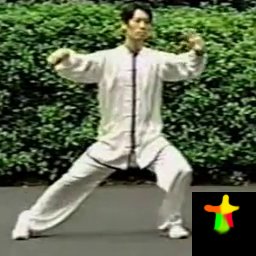}\includegraphics[width=0.15\columnwidth]{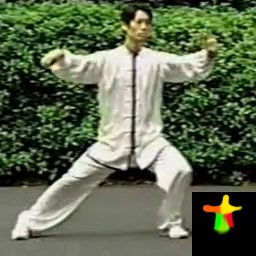}\includegraphics[width=0.15\columnwidth]{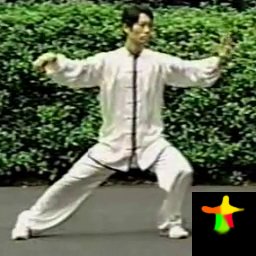} \\
         
         \includegraphics[width=0.15\columnwidth]{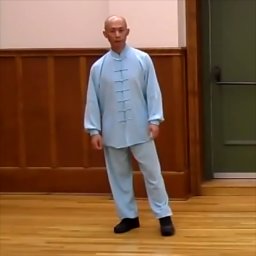}\includegraphics[width=0.15\columnwidth]{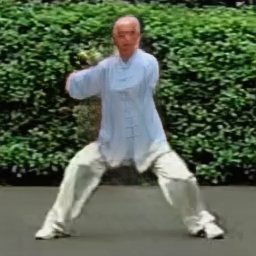}\includegraphics[width=0.15\columnwidth]{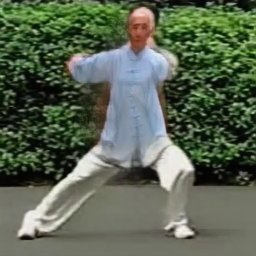}\includegraphics[width=0.15\columnwidth]{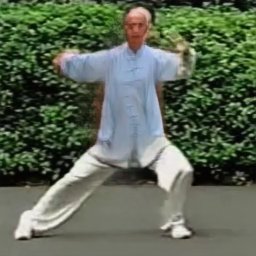}\includegraphics[width=0.15\columnwidth]{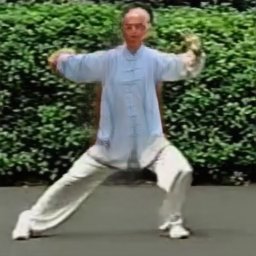}\includegraphics[width=0.15\columnwidth]{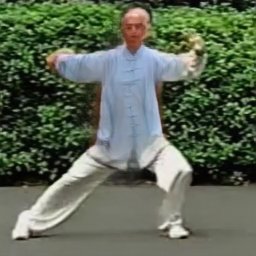}\includegraphics[width=0.15\columnwidth]{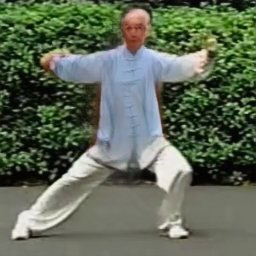}\includegraphics[width=0.15\columnwidth]{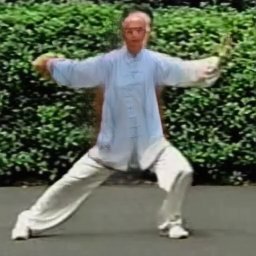} \\
         
         \includegraphics[width=0.15\columnwidth]{figures/vox/swaps/white}\includegraphics[width=0.15\columnwidth]{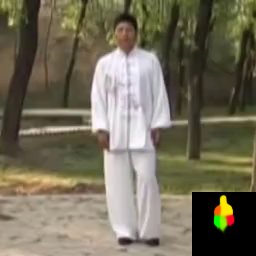}\includegraphics[width=0.15\columnwidth]{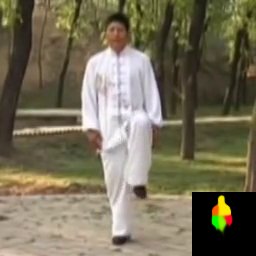}\includegraphics[width=0.15\columnwidth]{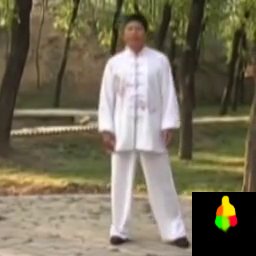}\includegraphics[width=0.15\columnwidth]{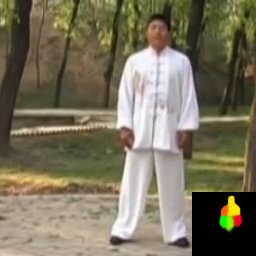}\includegraphics[width=0.15\columnwidth]{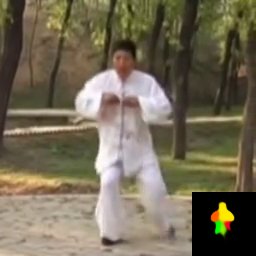}\includegraphics[width=0.15\columnwidth]{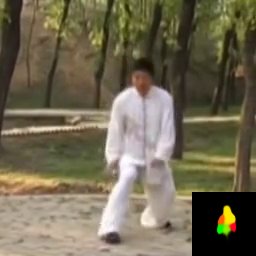}\includegraphics[width=0.15\columnwidth]{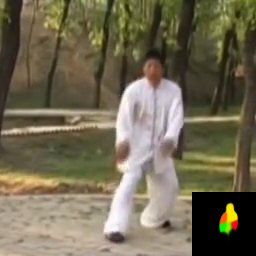} \\
         
         \includegraphics[width=0.15\columnwidth]{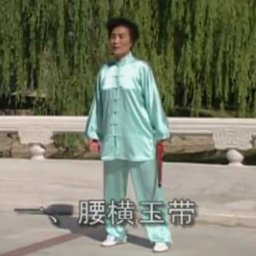}\includegraphics[width=0.15\columnwidth]{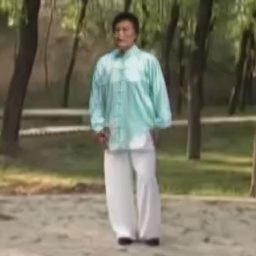}\includegraphics[width=0.15\columnwidth]{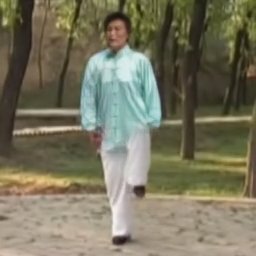}\includegraphics[width=0.15\columnwidth]{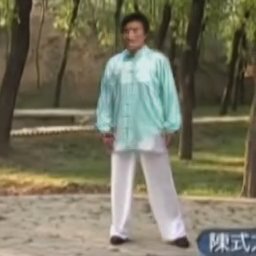}\includegraphics[width=0.15\columnwidth]{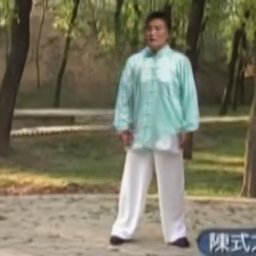}\includegraphics[width=0.15\columnwidth]{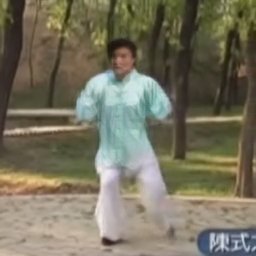}\includegraphics[width=0.15\columnwidth]{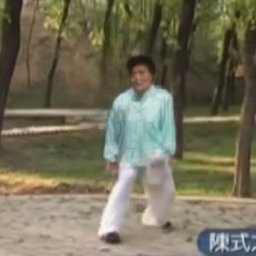}\includegraphics[width=0.15\columnwidth]{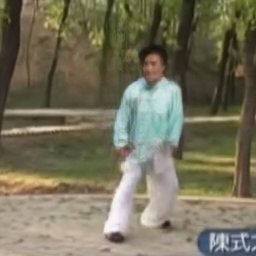} \\

    \end{tabular}}
    
    \captionof{figure}{\textbf{Visual results of video-editing for \textit{Tai-Chi-HD}}. In the odd rows the target frames from video sequences are depicted, alongside the masks of interest (in the right bottom corners) intended to be swapped. In the even rows the source image is shown, followed by the generated frames containing the indicated parts swapped from the source. Results achieved with model for \textit{K}=10 are depicted. Best viewed with digital zoom.}
    \label{fig:visual_taichi_swaps}

\end{table*}{}


\end{document}